\documentclass[11pt]{article}

\usepackage[final]{acl}

\usepackage{times}
\usepackage{latexsym}

\usepackage[T1]{fontenc}

\usepackage[utf8]{inputenc}

\usepackage{microtype}

\usepackage{inconsolata}

\usepackage{graphicx}
\usepackage{colortbl}
\definecolor{mygray}{gray}{.88}
\usepackage{amsthm} 
\usepackage{xcolor}
\usepackage{mathrsfs}
\usepackage{graphicx}
\usepackage{amssymb}
\usepackage{tablefootnote}
\usepackage{booktabs}
\usepackage{hyperref}
\usepackage{amsmath}
\usepackage{threeparttable}
\usepackage{wrapfig}
\usepackage{amssymb}
\usepackage{multirow}
\usepackage{graphicx}
\usepackage[section]{placeins}
\usepackage{booktabs}
\usepackage{arydshln}
\usepackage{algorithm}
\usepackage{newfloat}
\usepackage{listings}
\usepackage{tabularx}
\usepackage{caption}
\usepackage{algpseudocode}
\usepackage{subfigure}
\usepackage{subcaption}
\usepackage{wrapfig}
\usepackage{enumitem}
\usepackage{arydshln}
\usepackage{balance}
\usepackage{multicol}

\definecolor{color1}{HTML}{ffdebf}
\definecolor{color2}{HTML}{ffefe0}
\definecolor{color3}{HTML}{E6ECE3}
\definecolor{color4}{HTML}{feeafa}
\definecolor{color5}{HTML}{dee2ff}
\definecolor{color6}{HTML}{ffc0c3}
\definecolor{color7}{HTML}{00ffff}
\definecolor{dash1}{HTML}{db8a87}
\definecolor{dash2}{HTML}{ded6e6}
\newcolumntype{b}{>{\columncolor{gray!10}}c}

\newcommand{\xv}[0]{\ensuremath{\boldsymbol{x}} }

\title{PDR: A Plug-and-Play Positional Decay Framework for LLM Pre-training Data Detection}

\author{Jinhan Liu\\ \And
        Yibo Yang \\ \And
        Ruiying Lu \\ \And 
        Piotr Piękos \\ \And 
        Yimeng Chen \\ \And 
        Peng Wang\\ \And 
        Dandan Guo\\ \And
 }

\begin{document}
\maketitle

\begin{abstract}
\label{sec:abstract}

Detecting pre-training data in Large Language Models (LLMs) is crucial for auditing data privacy and copyright compliance, yet it remains challenging in black-box, zero-shot settings where computational resources and training data are scarce. While existing likelihood-based methods have shown promise, they typically aggregate token-level scores using uniform weights, thereby neglecting the inherent information-theoretic dynamics of autoregressive generation. In this paper, we hypothesize and empirically validate that memorization signals are heavily skewed towards the high-entropy initial tokens, where model uncertainty is highest, and decay as context accumulates. To leverage this linguistic property, we introduce Positional Decay Reweighting (PDR), a training-free and plug-and-play framework. PDR explicitly reweights token-level scores to amplify distinct signals from early positions while suppressing noise from later ones. Extensive experiments show that PDR acts as a robust prior and can usually enhance a wide range of advanced methods across multiple benchmarks.

\end{abstract}
\vspace{-2mm}

\section{Introduction}
\label{sec:introduction}
\vspace{-2mm}

As Large Language Models (LLMs)  are trained on vast and diverse corpora from the internet~\citep{achiam2023gpt, touvron2023llama}, there exists a non-negligible risk that sensitive or personally identifiable information may be memorized and unintentionally exposed through model outputs~\citep{grynbaum2023times, mozes2023use}.
Pre-training data detection as a special case of Membership Inference Attack (MIA) aims to determine whether a sample was part of a model's training set~\citep{hu2022membership, wu2025membership}, which has become increasingly critical in scenarios such as training data auditing, copyright infringement detection, and test set contamination analysis~\citep{bertran2023scalable,mia_mink_plus}, where identifying memorized content is essential for ensuring data integrity and compliance.

Existing detection methods for LLMs can be broadly categorized into likelihood-based and non-likelihood-based approaches. Among dominant likelihood-based methods, Loss~\citep{mia_loss} averages log-likelihoods across all tokens in the test sequence to serve as the detection score, while Min-k\%~\citep{mia_mink} and Min-k\%++~\citep{mia_mink_plus}  select some the tokens with  the lowest-probability from a sequence to compute its detection score. 
Methods like ReCaLL~\citep{xie-etal-2024-recall}, and Ref~\citep{mia_carlini} introduce a reference point to calibrate likelihood-based scores, either prefixing target data points with non-member context or using a smaller auxiliary LLM. FSD~\citep{zhang2025finetuning} fine-tunes the target LLM on some non-member samples before computing the likelihood-based score~\citep{zhang2025finetuning}. 
While varied in their specific strategies, these methods share a fundamental, unaddressed limitation: they typically aggregate token-level scores using uniform weights. 
Whether aggregating scores from all tokens or a selected subset in the sequence, 
they assign equal weight to each included token’s contribution to the final detection score, failing to explicitly account for the positional decay of memorization signals.

\begin{figure*}[t]
    \centering
    \begin{minipage}{0.45\textwidth}
        \centering
        \includegraphics[width=\linewidth]{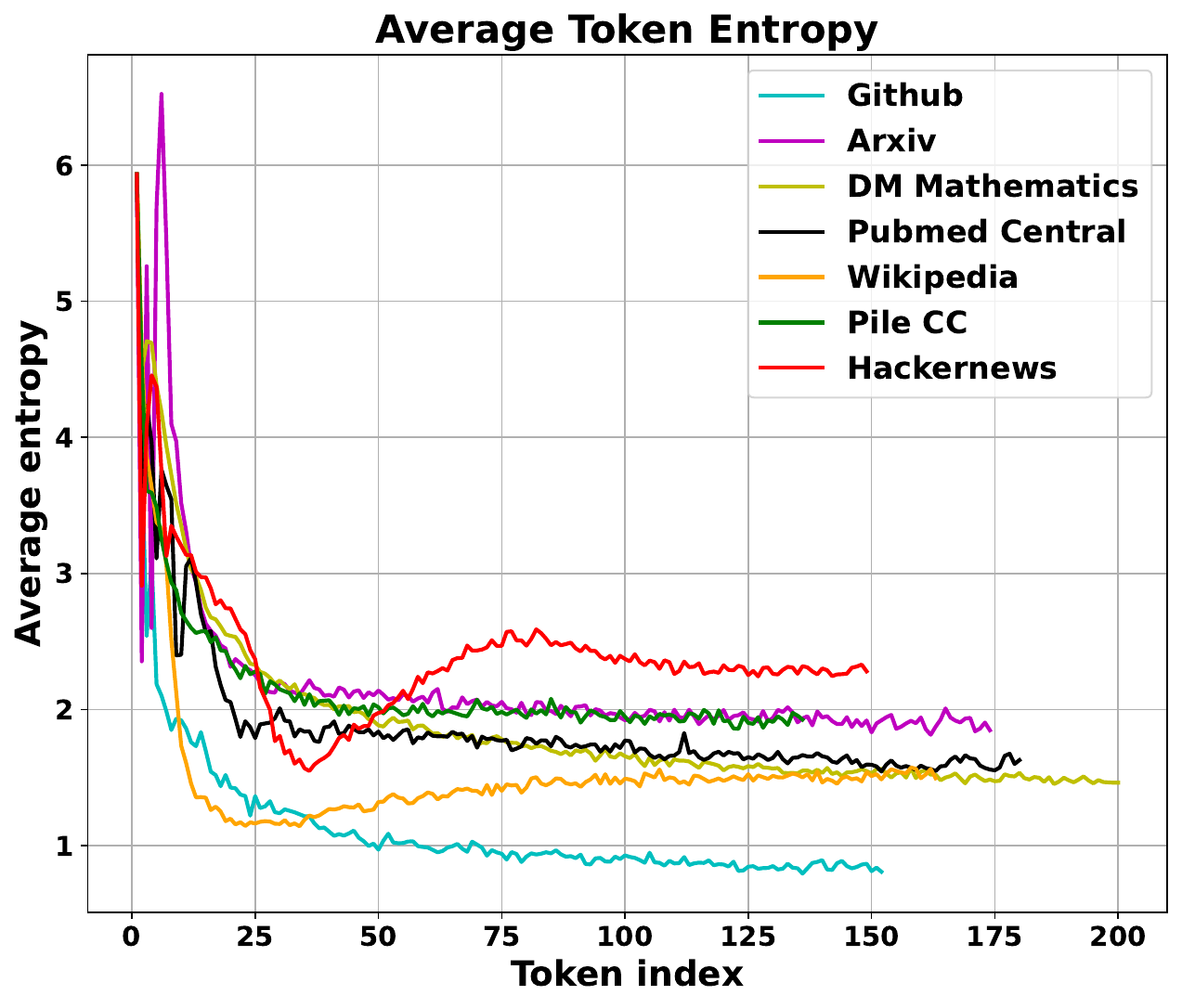}
        \centerline{(a) Entropy trends}
    \end{minipage}\hfill 
    \begin{minipage}{0.45\textwidth}
        \centering
        \includegraphics[width=\linewidth]{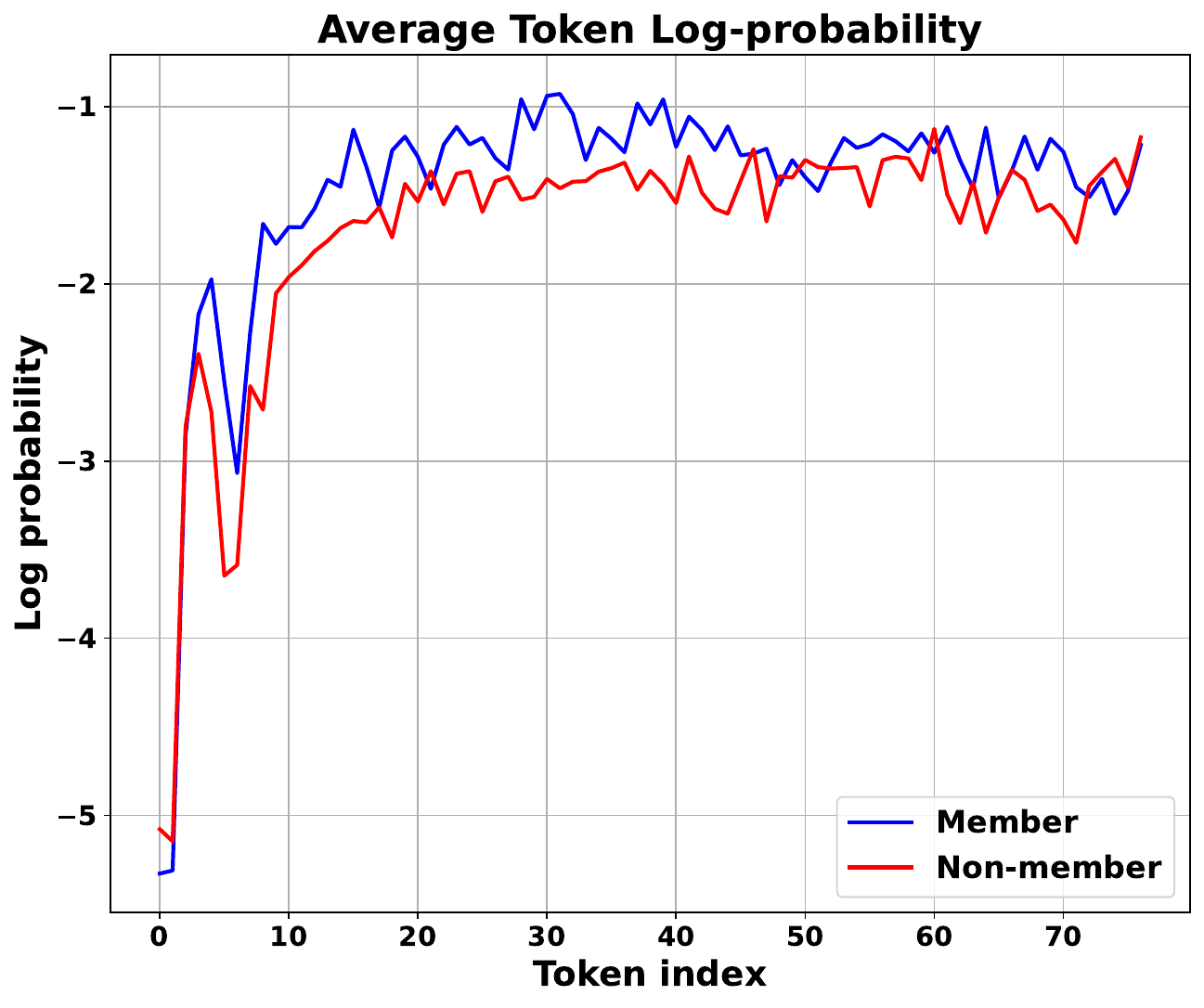}
        \centerline{(b) Log-probability trends}
    \end{minipage}
    \caption{\small{Visualization of (a) token-level entropy on subsets of the challenging Mimir dataset and (b)  the average token-level log-probability for members and non-members on  WikiMIA dataset   for LLaMA-13B model.}}
    \label{fig:motivation_plots}
    \vspace{-15pt}
\end{figure*}

Our work is motivated by a key insight from information theory: conditioning on more information cannot increase entropy \citep{shannon1948mathematical}. In autoregressive models, this implies that as more context accumulates, the model's predictive uncertainty for the same token should not increase. This motivated us to hypothesize an empirical trend: in typical language generation, token-level entropy usually tends to decrease as a sequence progresses. We empirically investigate this hypothesis in Fig.~\ref{fig:motivation_plots} (a). The results reveal a dominant, albeit sometimes noisy, downward trend across diverse datasets. While corpora with heterogeneous structures like ArXiv and HackerNews show volatility, all datasets share a crucial characteristic: a high-entropy initial region that drops sharply. Consequently, an unusually confident prediction (high probability) for an early, high-entropy token is far more surprising—and thus more indicative of memorization—than comparable confidence later in the sequence. This is because in later positions, the abundance of context makes predictions easier for both member and non-member samples, thus shrinking the discriminative gap between them. 

This leads to our core hypothesis: \textit{the memorization signal is not uniformly distributed, but is concentrated in the initial stages of a sequence, with its discriminative power generally decaying with token position}. However, existing likelihood-based methods, by utilizing uniform score aggregation, dilute this skewed and powerful signal with less informative signals from later positions. 

Capitalizing on this key observation, we introduce Positional Decay Reweighting (PDR), a simple, effective, and ``plug-and-play'' method designed to align the scoring process with this positional signal decay.
By applying monotonic decay functions (e.g., linear, exponential, polynomial), PDR systematically amplifies the high-value signals from early tokens while attenuating potential noise from later ones, thereby focusing the inference on the most informative parts of the sequence.
Its key advantage is versatility: PDR can be seamlessly integrated into existing likelihood-based scoring functions.
Extensive experiments validate that this straightforward modification yields substantial and consistent performance gains, improving upon advanced Min-$k$\%++ by up to 4.7 AUROC points on the WikiMIA benchmark of 128 length.

Our main contributions can be summarized as follows: \textit{(1)} We are the first to systematically demonstrate and analyze the positional decay of memorization signals from the view of token-level entropy, exposing the ``uniform score aggregation'' limitation inherent in prior methods.    \textit{(2)} We propose Positional Decay Reweighting (PDR), a training-free and plug-and-play framework that reweights token scores to amplify early signals. Crucially, PDR incurs zero  additional computational overhead, while requiring no access to training data. \textit{(3)} Our results across diverse LLMs and benchmarks establish PDR as an effective plug-and-play method, yielding notable performance gains especially for Min-$k$\%++.

\vspace{-2mm}
\section{Related Work}
\vspace{-2mm}
\subsection{Membership Inference Attacks.}
Membership Inference Attacks (MIA) have been a key topic in security and privacy~\citep{shokri2017membership, mia_loss}, aiming to determine if a data point was part of a model's training dataset. Research in both vision~\citep{Dubinski_2024_WACV} and language~\citep{mia_difficulty_calib,mia_neighbor} has led to advanced attack techniques, including \textbf{LiRA}~\citep{carlini2022membership} using shadow models to estimate logit distributions for likelihood-ratio tests,   \textbf{RMIA}~\citep{zarifzadeh2023low} constructs robust pairwise likelihood-ratio tests using a population of reference models, and \textbf{RaMIA}~\citep{tao2025range}, which  extends the scope by testing if the model was trained on any data within a specified semantic range, capturing privacy risks from similar or partially overlapping data. These methods offer insights into privacy risks~\citep{mireshghallah-etal-2022-quantifying}, test-set contamination~\citep{oren2023proving}, and copyright protection~\citep{meeus2023did,duarte2024decop}.

\subsection{Pre-training Data Detection.} 
Detecting training data in LLMs is challenging due to the scale of pre-training and the impracticality of shadow models \citep{mia_mink}. While some frameworks focus on prompt tuning or distribution-free attacks \citep{DF-MIA, MIA-Tuner}, \textbf{likelihood-based methods} remain dominant. 
Foundational approaches like \textbf{Loss} \citep{mia_loss} utilize average log-likelihood, which is often calibrated using reference models \textbf{Ref} \citep{mia_carlini} or synthetic neighbors \textbf{Neighbor} \citep{mia_neighbor}. To better capture memorization, outlier-based methods such as \textbf{Min-$k$\%} \citep{mia_mink} and its normalized version \textbf{Min-$k$\%++} \citep{mia_mink_plus} focus on the lowest-probability tokens. More recent variants incorporate conditional contexts \textbf{ReCaLL} \citep{xie-etal-2024-recall}, fine-tuning-based amplification \citep{zhang2025finetuning}, or multi dynamic signals like loss change rates \textbf{CAMIA} \citep{CAMIA}. Distinct from these scoring functions, our work identifies the \textbf{positional decay of memorization signals} through token entropy and provides a training-free and plug-and-play enhancement applicable to various likelihood-based baselines.


\section{Background}
\label{sec:background}

We first formalize the problem of pre-training data detection as defined in prior studies~\citep{shokri2017membership,mia_mink,mimir}, and then the likelihood-based scoring functions for Pre-training Data Detection methods in LLMs.

\subsection{Problem Statement}
Pre-training data detection is cast as a membership inference attack (MIA)~\citep{shokri2017membership}. 
Denote a pre-trained auto-regressive LLM as $M$ and its unknown training corpus as $\mathcal{D}$. For an arbitrary text sample $x$, MIA aims to infer whether $x\in\mathcal{D}$ (member sample) or $x\notin\mathcal{D}$ (non-member sample). Let $s(x;M)$ represent the scoring function that assigns a real-valued “membership” score to $x$ based on $M$'s outputs. We make a binary decision via
\begin{equation}
\hat{y} \;=\;\mathbb{I}\bigl(s(x;M)\ge\epsilon \bigr),
\end{equation}\label{auto}
where $\epsilon$ is a case-specific threshold and $\mathbb{I}(\cdot)$ is the indicator function. Consistent with the grey-box setting~\citep{mia_mink,mimir,mia_mink_plus}, we assume that only $M$'s output statistics (logits, token probabilities, loss values) are accessible; internal weights and gradients remain hidden. Designing an effective $s(x;M)$ to maximize the separation between member and non-member distributions is the core of detection task.

\subsection{Likelihood-based Score functions}
\label{sec:likehood-bassed score functions}
Modern LLMs are trained by maximizing the likelihood of training token sequences~\citep{GPT2,GPT3}. Concretely, given a sequence $\xv \!=\!(x_1,\dots,x_T)$, an auto-regressive LLM factorizes its joint probability using the chain rule: 
$p(\xv)
\!=\! \prod_{t=1}^T p\bigl(x_t \mid x_{<t}\bigr)$, 
where $x_{<t}=(x_1,\dots,x_{t-1})$ is the prefix context. At inference time, the model generates text token by sampling from the conditional distribution $p(\cdot\mid x_{<t})$. In light of this, researchers usually design likelihood-based scoring functions to detect pretraining data in LLMs \textcolor{blue}~\citep{mia_loss}. For example, based on the observation that members tend to have higher log-likelihood than non-members, the loss-based score~\citep{mia_loss} is defined as the (negative) log-likelihood of the input sequence,
\begin{equation}\label{loss}
s_{\text{loss}}(x) = \frac{1}{T}\sum_{t=1}^T \log p(x_t \mid x_{<t}),
\end{equation}
where we flip the sign of the conventional loss-based score so that, consistent with other methods, higher scores indicate stronger membership. Instead of using the likelihood of all tokens, Min-$k$\% ~\citep{mia_mink} selects the $k$\% tokens with the smallest log-probabilities and averages them:
\begin{equation}\label{mink}
s_{\text{Min-}k\%}(x) = \frac{1}{|\mathcal{S}_k|}\sum_{x_t \in \text{Min-}k\% (\xv)} \log p(x_t \mid x_{<t}),
\end{equation}
where $\mathcal{S}_k$ represents the set of token positions corresponding to the smallest $k$\% log-probabilities in the sequence. The intuition is that a non-member example is more likely to include a few outlier words with low likelihoods than members. Other methods are deferred to Appendix~\ref{apd:likelihood-based_score_function}.

\vspace{-2mm}
\section{Methodology}
\label{sec:methodology}
\vspace{-2mm}

We first present the core motivation based on an empirical observation about token entropy. We then introduce the general, plug-and-play method, Positional Decay Reweighting (PDR), and demonstrate how apply it to enhance existing likelihood-based Pre-training Data Detection scores.
\begin{figure*}[tbp]
  \centering
  \includegraphics[width=0.98\linewidth]{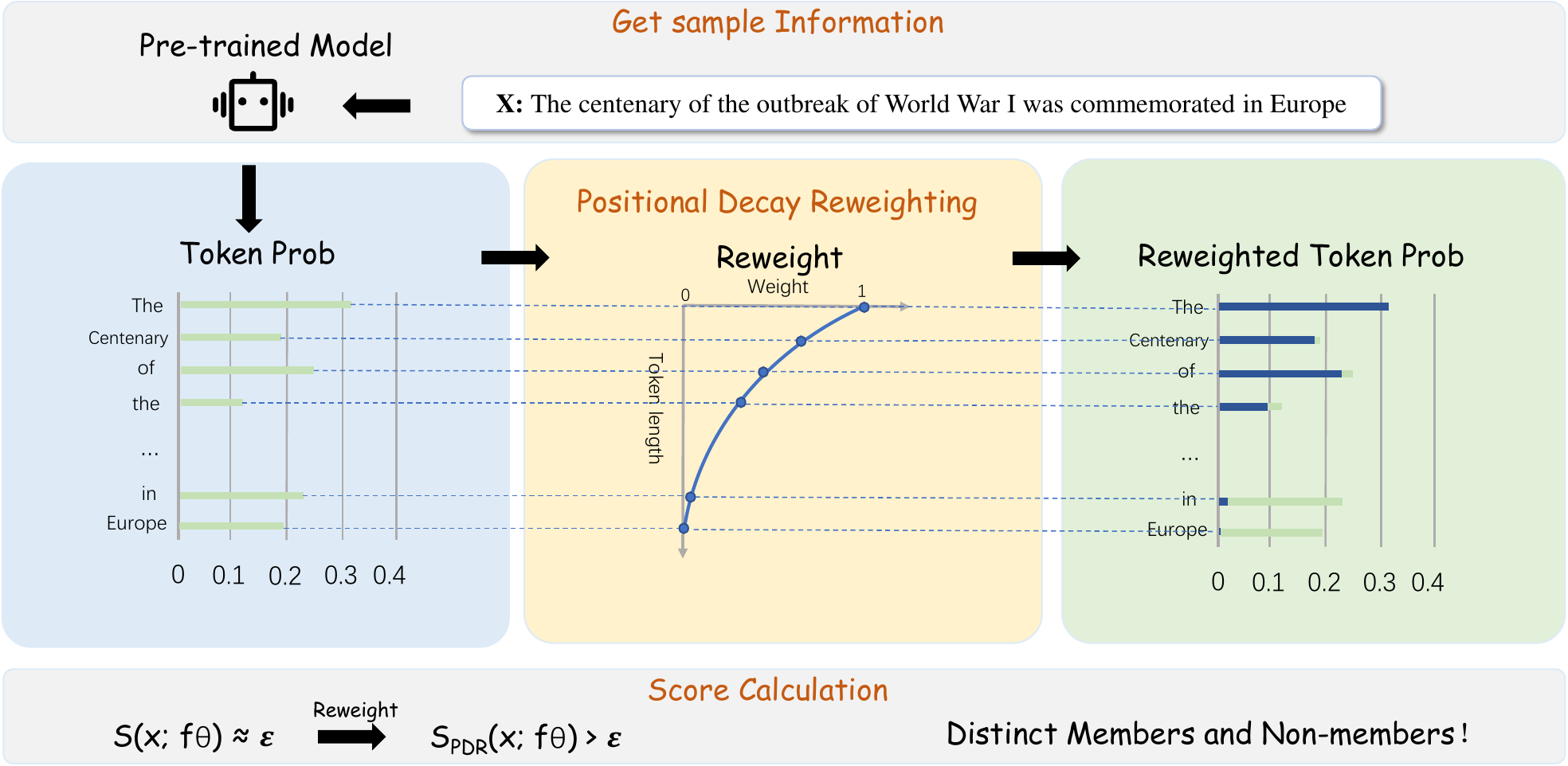}
   \caption{\textbf{Overview of Positional Decay Reweighting (PDR).} Our method reweights the predictive probabilities of input samples based on token positions, emphasizing early tokens with higher weights. This reweighting enhances the distinction between member and non-member samples by amplifying critical signals in the score $\mathcal{S}$, making it more effective for MIA. The framework is training-free, also plug-and-play for likelihood-based scoring methods.}
   \label{fig:method}
   \vspace{-10pt}
\end{figure*}
\subsection{Motivation}
Our methodology is built on a key insight into how autoregressive language models process information.
From an information-theoretic perspective, a fundamental principle is that conditioning on more information cannot increase entropy, i.e., $H(z|x,y) \leq H(z|y)$.
For autoregressive models, the uncertainty at each step can be quantified by the conditional entropy of the next token over the vocabulary $V$, given the prefix context $x_{<t}$:
\begin{equation}
    H(p(\cdot|x_{<t})) \!=\! -\sum_{v\in V}p(v|x_{<t})\log p(v|x_{<t}).
\end{equation}
Although the classic principle compares the entropy of the same random variable, whereas here we are comparing the entropy for different variables ($x_t$ and $x_{t+1}$), it is a widely observed empirical phenomenon that the entropy at position $t$ is frequently greater than at position $t+1$.

To empirically investigate this phenomenon, we visualized the average token-level entropy across multiple diverse datasets, as shown in Fig.~\ref{fig:motivation_plots} (a).
The visualization reveals two crucial findings. First, despite the varied nature of the corpora, they all exhibit a \textbf{dominant trend}: a high-entropy initial region followed by a general downward trend as the sequence progresses.
Second, it highlights key differences in these trends; while datasets like Github and Wikipedia show a relatively smooth decay, corpora with more heterogeneous structures---such as {ArXiv} (with section headings and equations) and {HackerNews} (mixing prose, code, and quotes)---display significantly more volatility.

These empirical findings have direct implications for membership inference, as the initial high-entropy region provides a unique setting to distinguish memorization from generalization.
An unusually confident (i.e., high-probability) prediction for a token in such a position strongly suggests that this confidence does not stem from contextual generalization, but rather from rote memorization of specific sequences in its training set. Conversely, in later positions (low-entropy regions), the abundance of context makes predictions easier for \textbf{both} member and non-member samples, thus shrinking the discriminative gap between them.
This is directly confirmed by Fig.~\ref{fig:motivation_plots} (b), which shows that this gap is largest at the beginning of the sequence and diminishes over time.

This analysis leads to our refined core hypothesis: \textit{the memorization signal is not uniformly distributed} but is heavily skewed towards the beginning of a sequence, with its strength generally decaying with token position. This crucial insight reveals a limitation in existing likelihood-based methods.
Whether they use scores from all tokens (like Loss) or from a subset of low-likelihood tokens (e.g., Min-$k$\%), {\textbf{they overwhelmingly rely on uniform weighting schemes
}.} By treating the scores from different positions as equally important, they dilute the potent, high-fidelity signals concentrated in the early positions with noisy, less informative signals from the end.
This oversight prevents them from fully exploiting the  powerful evidence of memorization.
Therefore, a principled, position-aware approach is not merely an incremental improvement, but a necessary step to enhance performance.

\subsection{Plug-and-Play Positional Decay Reweighting (PDR)}
Based on our core hypothesis established above---that memorization signals are heavily skewed towards the beginning of a sequence---we argue that the performance of likelihood-based Detection methods is fundamentally limited by their {uniform scoring mechanism}.
To rectify this, we propose Positional Decay Reweighting (PDR): a simple, effective, and ``plug-and-play'' framework designed to inject this crucial positional prior into existing methods. Our overview is illustrated in Fig. \ref{fig:method}.

 PDR operates by re-weighting token-level scores using monotonically decreasing functions based on a token's position $t$ in a sequence of length $T$.
This systematically assigns higher importance to earlier tokens, where the signal is strongest, and lower importance to later ones.
We explore three simple, standardized, and effective families of decay functions:

\begin{enumerate}[leftmargin=*,topsep=0em,itemsep=0em]
    \item \textbf{Linear Decay:} This function linearly decreases the weight from 1. The rate of decay is controlled by a single hyperparameter $\alpha \in [0, 1]$:
    \begin{equation}
    w_{\text{linear}}(t) = 1 - \alpha \left( \frac{t-1}{T-1} \right),
    \end{equation}
    where $T$ is the total sequence length. When $\alpha=0$, all tokens are weighted equally, reducing to the original unweighted score.

    \item \textbf{Exponential Decay:} This function applies a sharper, non-linear decay, placing a much stronger emphasis on the initial tokens:
    \begin{equation}
    w_{\text{exp}}(t) = \exp(-\alpha \cdot (t-1)).
    \end{equation}
    The hyperparameter $\alpha \ge 0$ controls the steepness of the decay.

    \item \textbf{Polynomial Decay:} This function provides a flexible decay curve whose shape is controlled by the exponent $\alpha$. The hyperparameter $\alpha > 0$ determines the curvature of the decay. Values of $\alpha > 1$ result in a slower initial decay, while values $0 < \alpha < 1$ lead to a faster initial decay:
    \begin{equation}
    w_{\text{poly}}(t) = \left(1 - \frac{t - 1}{T - 1} \right)^{\alpha}.
    \end{equation}
\end{enumerate}

We defer the visualization of three weight decay functions into Fig.\ref{fig:weight_visualization}
 of Appendix \ref{apd:weight_decay_sec}.

\subsection{Applying PDR to Detection Scoring Functions}
A key advantage of PDR is its ``plug-and-play'' nature. It operates as a lightweight wrapper designed to correct existing likelihood-based methods, requiring no modification to the target model's architecture or training process. This makes it a broadly applicable technique. We now show how PDR integrates with two representative scoring functions.

For methods that aggregate scores across the entire sequence, such as standard Loss score in Eq.\eqref{loss}, PDR injects the positional prior by applying weights to each token's log-probability before aggregation. The resulting PDR-Loss score is defined as:
\begin{equation}
s_{\text{PDR-Loss}}(x) = \frac{1}{T} \sum_{t=1}^{T} w(t) \cdot \log P(x_{t}|x_{<t}).
\end{equation}

The integration is more nuanced for outlier-based methods like Min-$k$\% in Eq.\eqref{mink}. Here, a crucial detail is the order of operations. To preserve the integrity of the outlier selection process, PDR is applied \textit{after} the tokens have been selected based on their original, unweighted scores. The re-weighting then uses the \textit{original position} of these selected tokens, ensuring that we are amplifying the most informative signals as identified by the baseline method. The PDR-Min-$k$\% score is thus:
\begin{equation}
s_{\text{PDR-Min-$k$\%}}(x) = \frac{1}{|\mathcal{S}_{k}|} \sum_{t \in \mathcal{S}_{k}} w(t) \cdot \log P(x_{t}|x_{<t}),
\end{equation}
where $\mathcal{S}_{k}$ is the set of token positions with the smallest $k$\% log-probabilities.
\begin{table*}[t]
\centering
\caption{\small{AUROC results on WikiMIA benchmark \citep{mia_mink}.  \textit{Ori.} and \textit{Para.} denote the original and paraphrased settings. {$^\dag$Neighbor results are from \citet{mia_mink_plus}.} For method pairs, the format is: baseline / \textbf{w/ LPDR} (our Linear PDR). The higher score in a pair is in \textbf{bold}, and the performance gains of our method on the average results are highlighted in \color{purple}{purple}.}}
\label{tab:wikimia_main}
\renewcommand{\arraystretch}{1.2}
\resizebox{\textwidth}{!}{
\begin{tabular}{llcccccccccccccc}
\toprule
& & \multicolumn{2}{c}{\textbf{Lowercase}} & \multicolumn{2}{c}{\textbf{Zlib}} & \multicolumn{2}{c}{\textbf{$^\dag$Neighbor}} & \multicolumn{2}{c}{\textbf{Loss}} & \multicolumn{2}{c}{\textbf{Ref}} & \multicolumn{2}{c}{\textbf{Min-$k$\%}} & \multicolumn{2}{c}{\textbf{Min-$k$\%++}} \\
\cmidrule(lr){3-4} \cmidrule(lr){5-6} \cmidrule(lr){7-8} \cmidrule(lr){9-10} \cmidrule(lr){11-12} \cmidrule(lr){13-14} \cmidrule(lr){15-16}
\textbf{Model} & \textbf{Len} & \textit{Ori.} & \textit{Para.} & \textit{Ori.} & \textit{Para.} & \textit{Ori.} & \textit{Para.} & \textit{Ori.} & \textit{Para.} & \textit{Ori.} & \textit{Para.} & \textit{Ori.} & \textit{Para.} & \textit{Ori.} & \textit{Para.} \\
\midrule
\multirow{3}{*}{\textbf{Mamba-1.4B}} 
& 32 & 60.9 & 60.6 & 61.9 & 62.3 & 64.1 & 63.6 & 61.0/\textbf{61.5} & 61.3/\textbf{61.8} & 62.2/62.2 & 62.3/62.3 & 63.3/\textbf{63.5} & 62.9/\textbf{63.1} & 66.4/\textbf{67.4} & 65.7/\textbf{66.3} \\
& 64 & 57.0 & 57.0 & 60.4 & 59.1 & 60.6 & 60.6 & 58.2/\textbf{59.7} & 56.4/\textbf{59.5} & 60.6/\textbf{61.1} & 59.6/\textbf{60.8} & 61.7/\textbf{62.9} & 58.0/\textbf{61.8} & 67.2/\textbf{68.2} & 62.2/\textbf{65.5} \\
& 128 & 58.5 & 57.7 & 65.6 & 65.3 & 64.8 & 62.6 & 63.3/\textbf{63.6} & 62.7/\textbf{64.1} & 62.0/\textbf{64.1} & 61.1/\textbf{64.6} & 66.8/65.5 & 64.4/\textbf{65.8} & 67.7/\textbf{70.2} & 63.3/\textbf{68.2} \\
\midrule
\multirow{3}{*}{\textbf{Pythia-6.9B}} 
& 32 & 62.2 & 61.7 & 64.4 & 64.2 & 65.8 & 65.5 & 63.8/\textbf{64.0} & 64.1/\textbf{64.2} & 63.6/63.5 & 63.5/63.5 & 66.3/66.3 & 65.1/65.1 & 70.3/\textbf{70.8} & 67.6/\textbf{67.7} \\
& 64 & 58.2 & 57.7 & 62.6 & 61.6 & 63.2 & 63.1 & 60.7/\textbf{62.6} & 59.3/\textbf{62.4} & 62.4/\textbf{63.3} & 62.9/\textbf{64.0} & 65.0/\textbf{66.7} & 61.1/\textbf{65.1} & 71.6/\textbf{72.1} & 64.2/\textbf{68.3} \\
& 128 & 60.5 & 59.9 & 67.6 & 67.4 & 67.5 & 64.3 & 65.1/\textbf{65.6} & 64.7/\textbf{66.6} & 63.3/\textbf{65.1} & 62.9/\textbf{65.9} & 69.5/67.8 & 67.0/\textbf{68.9} & 69.8/\textbf{72.4} & 65.9/\textbf{72.2} \\
\midrule
\multirow{3}{*}{\textbf{LLaMA-13B}} 
& 32 & 64.0 & 63.2 & 67.8 & 68.3 & 65.8 & 65.0 & 67.5/\textbf{67.7} & 68.0/\textbf{68.2} & 57.9/57.8 & 56.2/56.1 & 66.8/66.8 & 66.2/66.2 & 84.4/\textbf{85.9} & 82.7/\textbf{84.1} \\
& 64 & 62.0 & 61.0 & 65.3 & 65.3 & 64.1 & 64.7 & 63.6/\textbf{65.0} & 63.1/\textbf{66.2} & 63.4/59.8 & 60.9/57.9 & 66.0/\textbf{67.4} & 63.5/\textbf{67.2} & 84.3/\textbf{87.2} & 78.8/\textbf{84.3} \\
& 128 & 60.6 & 56.3 & 69.7 & 69.6 & 68.3 & 64.0 & 67.8/\textbf{68.7} & 67.2/\textbf{69.1} & 62.6/\textbf{64.9} & 59.7/\textbf{61.7} & 71.5/71.2 & 68.6/\textbf{71.0} & 83.8/\textbf{88.4} & 76.2/\textbf{84.3} \\
\midrule
\multirow{3}{*}{\textbf{GPT-NeoX-20B}} 
& 32 & 68.3 & 66.9 & 69.3 & 68.5 & 70.2 & 68.3 & 69.1/68.9 & 68.6/68.3 & 67.6/67.4 & 66.7/66.6 & 72.2/72.0 & 69.6/69.4 & 75.1/\textbf{75.2} & 69.7/69.5 \\
& 64 & 66.3 & 65.6 & 68.1 & 66.5 & 67.1 & 67.4 & 66.6/\textbf{67.6} & 64.4/\textbf{67.2} & 66.0/\textbf{66.8} & 66.0/\textbf{67.2} & 72.2/70.8 & 66.1/\textbf{68.7} & 76.5/76.4 & 66.2/\textbf{68.2} \\
& 128 & 68.0 & 67.6 & 72.3 & 72.0 & 71.6 & 69.6 & 70.7/70.7 & 69.7/\textbf{71.4} & 68.3/\textbf{69.5} & 68.4/\textbf{70.2} & 75.6/74.5 & 73.0/\textbf{75.2} & 75.4/\textbf{75.7} & 70.6/\textbf{72.6} \\
\midrule
\multirow{3}{*}{\textbf{OPT-66B}} 
& 32 & 62.8 & 62.3 & 65.8 & 65.3 & 68.2 & 66.7 & 65.6/65.6 & 65.3/65.0 & 68.6/68.6 & 67.9/\textbf{68.0} & 67.5/\textbf{67.7} & 65.8/65.8 & 69.7/\textbf{70.2} & 67.0/\textbf{67.1} \\
& 64 & 61.1 & 60.0 & 63.9 & 62.2 & 64.1 & 64.6 & 62.3/\textbf{64.2} & 60.3/\textbf{63.1} & 66.9/\textbf{68.2} & 67.8/\textbf{69.1} & 66.5/\textbf{68.1} & 62.5/\textbf{66.0} & 69.8/\textbf{70.1} & 63.3/\textbf{66.6} \\
& 128 & 58.9 & 57.6 & 67.3 & 66.9 & 67.7 & 63.4 & 65.5/\textbf{66.7} & 64.5/\textbf{66.9} & 66.9/\textbf{68.6} & 67.0/\textbf{69.5} & 70.6/70.6 & 67.2/\textbf{69.9} & 71.1/\textbf{72.9} & 67.0/\textbf{69.5} \\
\midrule
\multirow{3}{*}{\textbf{Average}} 
& 32 & 63.7 & 63.0 & 65.8 & 65.7 & 66.8 & 65.8 & 65.4/\textbf{65.5}$^{\color{purple}{+0.1}}$ & 65.5/65.5 & 64.0/63.9 & 63.3/63.3 & 67.2/\textbf{67.3}$^{\color{purple}{+0.1}}$ & 65.9/65.9 & 73.2/\textbf{73.9}$^{\color{purple}{+0.7}}$ & 70.5/\textbf{70.9}$^{\color{purple}{+0.4}}$ \\
& 64 & 60.9 & 60.3 & 64.1 & 62.9 & 63.8 & 64.1 & 62.3/\textbf{63.8}$^{\color{purple}{+1.5}}$ & 60.7/\textbf{63.7}$^{\color{purple}{+3.0}}$ & 63.9/63.9 & 63.5/\textbf{63.8}$^{\color{purple}{+0.3}}$ & 66.3/\textbf{67.2}$^{\color{purple}{+0.9}}$ & 62.2/\textbf{65.8}$^{\color{purple}{+3.6}}$ & 73.9/\textbf{74.8}$^{\color{purple}{+0.9}}$ & 66.9/\textbf{70.6}$^{\color{purple}{+3.7}}$ \\
& 128 & 61.3 & 59.8 & 68.5 & 68.2 & 68.0 & 64.8 & 66.5/\textbf{67.1}$^{\color{purple}{+0.6}}$ & 65.7/\textbf{67.6}$^{\color{purple}{+1.9}}$ & 64.6/\textbf{66.4}$^{\color{purple}{+1.8}}$ & 63.8/\textbf{66.4}$^{\color{purple}{+2.6}}$ & 70.8/69.9 & 68.0/\textbf{70.2}$^{\color{purple}{+2.2}}$ & 73.6/\textbf{75.9}$^{\color{purple}{+2.3}}$ & 68.6/\textbf{73.3}$^{\color{purple}{+4.7}}$ \\
\bottomrule
\end{tabular}
}
\vspace{-10pt}
\end{table*}

PDR can also be combined with other scoring functions, such as the reference-based method (Ref), the normalized outlier method (Min-$k$\%++), and finetuned-based FSD . The full set of PDR-enhanced scoring functions and  algorithm are detailed in Appendix~\ref{apd:likelihood-based_score_function_with_pdr} and Algorithm~\ref{alg:pdr_algorithm}. By systematically amplifying the signal from critical early tokens, PDR aims to widen the score distribution gap between member and non-member samples, thereby enhancing overall detection performance.

\section{Experiments}
\begin{figure*}[htbp]
 \vspace{-10pt}
    \begin{minipage}[c]{0.48\textwidth}
        \captionof{table}{\small{AUROC scores of various MIA methods over five Pythia models on the Mimir dataset. Pub and Wiki denote Pubmed Central and Wikipedia (en). Avg* scores are computed by excluding Arxiv and HackerNews.{ $^\dag$Neighbor results are from \citet{mia_mink_plus}, induces significant extra computational cost than others ($25\times$ in this case), for which reason we don't run on the 12B model.}}}
        \vspace{-5pt}
        \label{tab:mimir_main}

\centering
\resizebox{1.0\textwidth}{!}{%
    \begin{tabular}{lcccccc}
        \toprule
        \textbf{Method} & \textbf{Wiki} & \textbf{Pile-CC}& \textbf{Pub} & \textbf{Math} & \textbf{GitHub}& \textbf{Avg*}\\
        \midrule
        Lowercase & 52.2  & 49.3  & 51.1  & 48.9  & 71.1    & 54.5  \\ 
        Zlib & 52.7  & 50.4  & 50.4  & 48.1  & 71.9    & 54.7  \\ 
        {$^\dag$Neighbor} & 51.9 & 50.1  & 49.2  & 47.4 & 69.3  & 53.6  \\
        Loss & 51.9  & 50.3  & 50.3  & 48.5  & 70.8   & 54.4  \\ 
        \rowcolor{gray!15} \textbf{w/ LPDR}& \textbf{52.8}  & \textbf{50.7}  & 50.3  & \textbf{48.6}  & \textbf{70.9}  & \textbf{54.6}  \\ 
        Min-$k$\% & 51.8  & 50.7  & 50.9  & 49.2  & 70.9   & 54.7  \\ 
        \rowcolor{gray!15} \textbf{w/ LPDR} & \textbf{54.2}  & \textbf{51.2}  & \textbf{51.0}  & \textbf{49.5}  & \textbf{71.0}  & \textbf{55.4}  \\ 
        Min-$k$\%++ & 54.0  & 50.5  & 51.9  & 50.3  & 70.4 & 55.4  \\ 
        \rowcolor{gray!15} \textbf{w/ LPDR} & \textbf{55.5}  & \textbf{50.8}  & \textbf{52.4}  & 50.3  & 70.1  & \textbf{55.8} \\ 
        \bottomrule
    \end{tabular}
    }
    \end{minipage}\hfill
    \begin{minipage}[c]{0.48\textwidth}
        \centering
        \includegraphics[width=\linewidth]{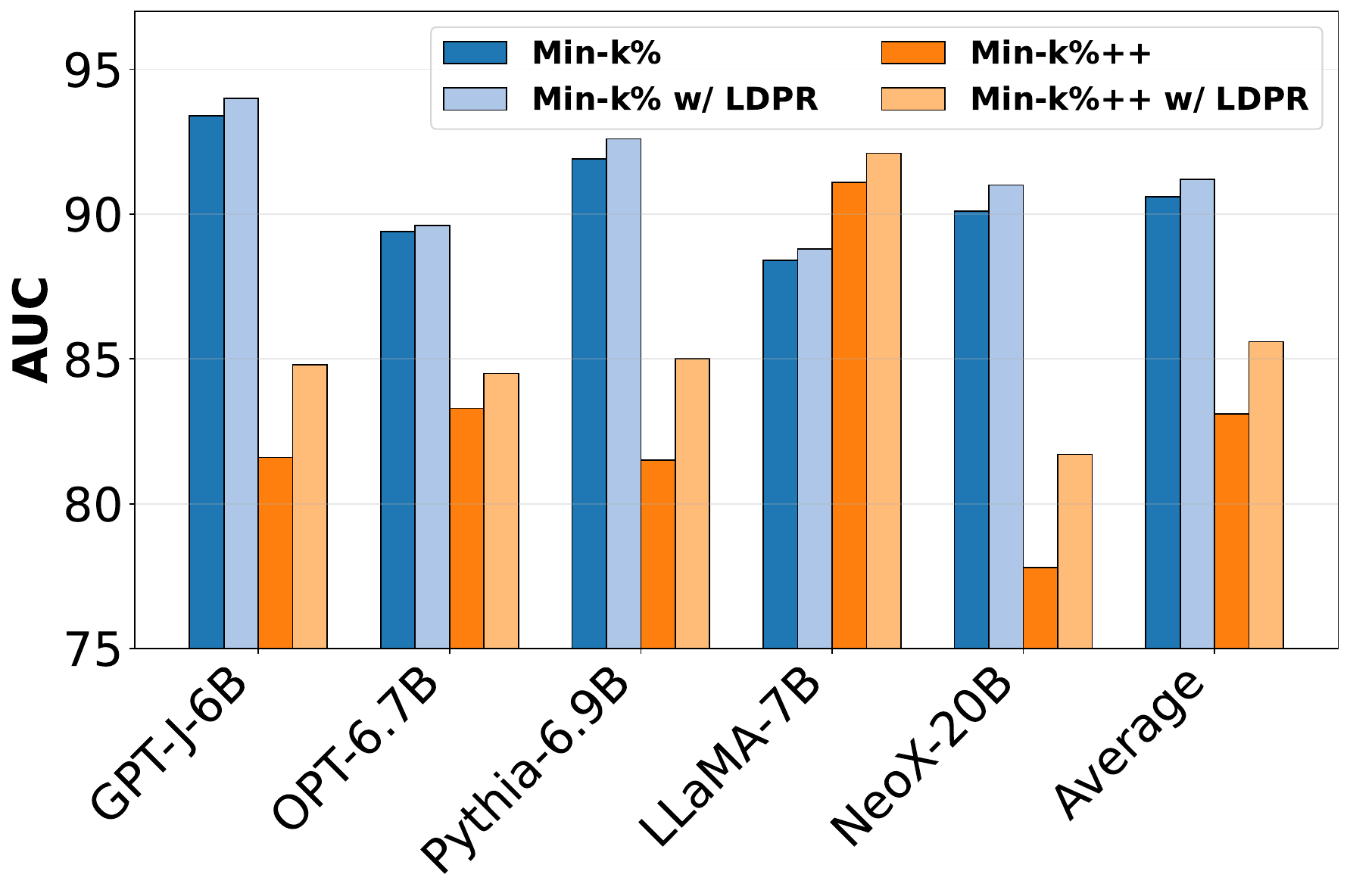}
        \captionof{figure}{\small{AUROC comparison of our LPDR method when integrated with Min-$k$\% and Min-$k$\%++ across various LLMs on WikiMIA dataset within the FSD framework.}}
        \label{fig:fsd_performance_comparison}
    \end{minipage}
    \vspace{-15pt}
\end{figure*}
\subsection{Setup}
\vspace{-2mm}
{\textbf{Benchmarks.} We evaluate our method on two commonly-used benchmarks for LLM pre-training data detection. (1)\textbf{WikiMIA}~\citep{mia_mink} uses Wikipedia texts, distinguishing members by timestamps, and includes different length text for both \textit{original} and \textit{paraphrased} settings. (2)\textbf{MIMIR}~\citep{mimir}, built on the Pile dataset~\citep{gao2020pile}, is more challenging as it minimizes distributional and temporal shifts between member and non-member data.}


{\textbf{Baselines.} We consider several representative and advanced methods as our baselines. A fundamental approach is \textbf{Loss}~\citep{mia_loss}. Reference-based methods include \textbf{Ref}~\citep{mia_carlini}, \textbf{Zlib}, and \textbf{Lowercase}~\citep{mia_carlini}. \textbf{Neighbor}~\citep{mia_neighbor} compares the sample's score against its neighbors' scores. \textbf{Min-$k$\%}~\citep{mia_mink} averages the lowest $k$\% of token scores, and \textbf{Min-$k$\%++}~\citep{mia_mink_plus} adds score normalization. We also include \textbf{FSD}~\citep{zhang2025finetuning}, which leverages score differences after fine-tuning the model on non-member data.}

{\textbf{Models.} For WikiMIA, we use Pythia~\citep{pythia} (2.8B, 6.9B, 12B), LLaMA~\citep{llama} (13B, 30B), GPT-NeoX~\citep{gptneox}(20B), OPT~\citep{opt} (66B), and Mamba~\citep{mamba} (1.4B, 2.8B). For MIMIR, we follow \cite{mimir} and use the Pythia model series (160M, 1.4B, 2.8B, 6.9B, 12B). For FSD, we follow \citet{zhang2025finetuning} and use GPT-J-6B, OPT-6.7B, Pythia-6.9B, LLaMA-7B, and GPT-NeoX-20B.}

\textbf{Metrics and Settings.}
Following standard practice~\citep{mia_carlini,mia_mink}, we use AUROC as the primary metric and also report True Positive Rate (TPR) at low False Positive Rates. For brevity, we use LPDR, EPDR, and PPDR to denote our PDR with Linear, Exponential, and Polynomial decay. We use a commonly-used $k=20$ for Min-$k$\% and Min-$k$\%++. In the main body, we primarily report results for LPDR. To demonstrate the general effectiveness of a simple and strong positional prior, we use a fixed $\alpha=1$ for all our  experiments with LPDR except for very short sequences (WikiMIA, $T=32$), where such a sharp is suboptimal. More details about datasets, baselines and settings are deferred to Appendix~\ref{apd:exp_settings}.
\subsection{Main Results.} 

\textbf{Results on WikiMIA.} As shown in Tab.~\ref{tab:wikimia_main}, we report the AUROC results of different methods on WikiMIA  with varying sequence lengths of \{32, 64, 128\} on different backbones; please see Appendix~\ref{apd:wikimia results}  for overall results on more methods (including our EPDR, PPDR), backbones and TPR numbers. {Besides, we also plot the ROC curves in Appendix~\ref{apd:auroc_curve_visualization} to demonstrate the consistent superiority of our method across various False Positive Rate (FPR) thresholds.}

We observe that introducing our proposed linear positional decay reweighting strategy generally enhances the performance of existing likelihood-based MIA methods. This improvement is especially evident on the advanced Min-$k$\%++. 
For instance, when combined with the Min-$k$\%++ method, the performance gains from our LPDR become more pronounced as sequence length increases.  Our LPDR improves the average AUROC by 0.7 (\textit{Ori.}) and 0.4 (\textit{Para.}) for length 32, by 0.9 (\textit{Ori.}) and 3.7 (\textit{Para.}) for length 64, and achieves the most significant gains of 2.3 (\textit{Ori.}) and 4.7 (\textit{Para.}) for length 128. These results prove the effectiveness of the our designed weight decay method in re-weighting token-level scores.

\begin{table*}[ht]
    \centering
    \caption{\small{Ablation study  about weighting schemes. We evaluate slope(CAMIA sample loss slope), different weight orderings (Random, Reverse), entropy-based weights (Sample, Dataset), and the reweighting for K-select in Min-$k$\% and Min-$k$\%++. Results are reported on the WikiMIA dataset using the Pythia-6.9B model with a sequence length of $T=128$. }}
    \vspace{-5pt}
    \label{tab:ab_ablation_weights}
    \renewcommand{\arraystretch}{1.2}
    \resizebox{0.90\textwidth}{!}{%
    \begin{tabular}{lcccccccccc}
        \toprule
        & & &\multicolumn{2}{c}{\textbf{Weights Order}} & \multicolumn{2}{c}{\textbf{Entropy}} & \multicolumn{1}{c}{\textbf{K select}} & \multicolumn{3}{c}{\textbf{w/ PDR(Ours)}} \\
        \cmidrule(lr){4-5} \cmidrule(lr){6-7} \cmidrule(lr){8-8} \cmidrule(lr){9-11}
        \textbf{Method} & \textbf{Base} & \textbf{Slope} & \textbf{Random} & \textbf{Reverse} & \textbf{Sample} & \textbf{Dataset} & \textbf{Before}  & \textbf{LPDR} & \textbf{EPDR} & \textbf{PPDR}\\
        \midrule
        Loss        & 65.1 & 65.1 & 64.5 & 63.4 & 63.2 & \textbf{66.4} &- &  \underline{65.6}&64.8&\underline{66.1} \\
        Ref         & 63.3 & 63.5 & 61.8 & 58.0 & 62.1 & \underline{63.8} &- & \textbf{65.1}&\textbf{67.5} & \textbf{65.7} \\
        Min-$k$\%   & 69.5 & 69.5 & 64.3 & 59.8 & 64.1 & \textbf{70.6} & 68.1 & 67.8 &69.2 &66.7\\
        Min-$k$\%++ & 69.8 & 69.8 & 66.5 & 61.0 & 68.7 & 70.5 & 70.3& \textbf{72.4} & \textbf{71.2} & \textbf{72.7}\\
        \bottomrule
    \end{tabular}
    }
    \vspace{-10pt}
\end{table*}


\textbf{Combination with FSD on WikiMIA.}
Since ours is a plug-and-play reweighting method, it can also be used to enhance the finetune-based FSD. We perform the experiments on  WikiMIA dataset with different LLMs, where we first finetune LLMs with non-member samples following its official code, then use Min-$k$\% and Min-$k$\%++ as its score functions. To combine ours with FSD, we apply our LPDR  to reweight the score functions. We show the results in Fig.~\ref{fig:fsd_performance_comparison} and defer details into Appendix~\ref{apd:fsd results}. We can find that our LPDR provides consistent improvements on FSD no matter with Min-$k$\%++ or Min-$k$\%++ as the score functions. It demonstrates that ours is also beneficial for finetune-based methods by reweighting its score functions.

\textbf{MIMIR Results.} As noted by prior work, MIMIR is particularly difficult because its training and non-training texts are sourced from the same sources, minimizing distributional shifts. Furthermore, we identify that the sub-datasets within MIMIR exhibit notable differences in structural composition. As visually confirmed by their volatile entropy profiles in Fig.~\ref{fig:motivation_plots} (a), corpora like \textit{ArXiv} and \textit{HackerNews} are structurally heterogeneous. This distinguishes them from more homogeneous corpora like Wikipedia or GitHub. We treat the heterogeneous datasets as stress tests and compute an \texttt{Avg*} score on the five sub-datasets that align with our method's positional prior. As listed in Tab.~\ref{tab:mimir_main}, on this benchmark, baselines themselves perform close to random guess, underscoring its difficulty. We can find that introducing ours can improve the Loss, Min-$k$\%, and Min-$k$\%++. It validates the efficacy of PDR's positional prior in structurally homogeneous text datasets. The complete averaged results for all sub-datasets and the detailed results are deferred to Appendix~\ref{apd:mimir results}.

\begin{figure}[htpb]
    \centering
    \includegraphics[width=\linewidth]{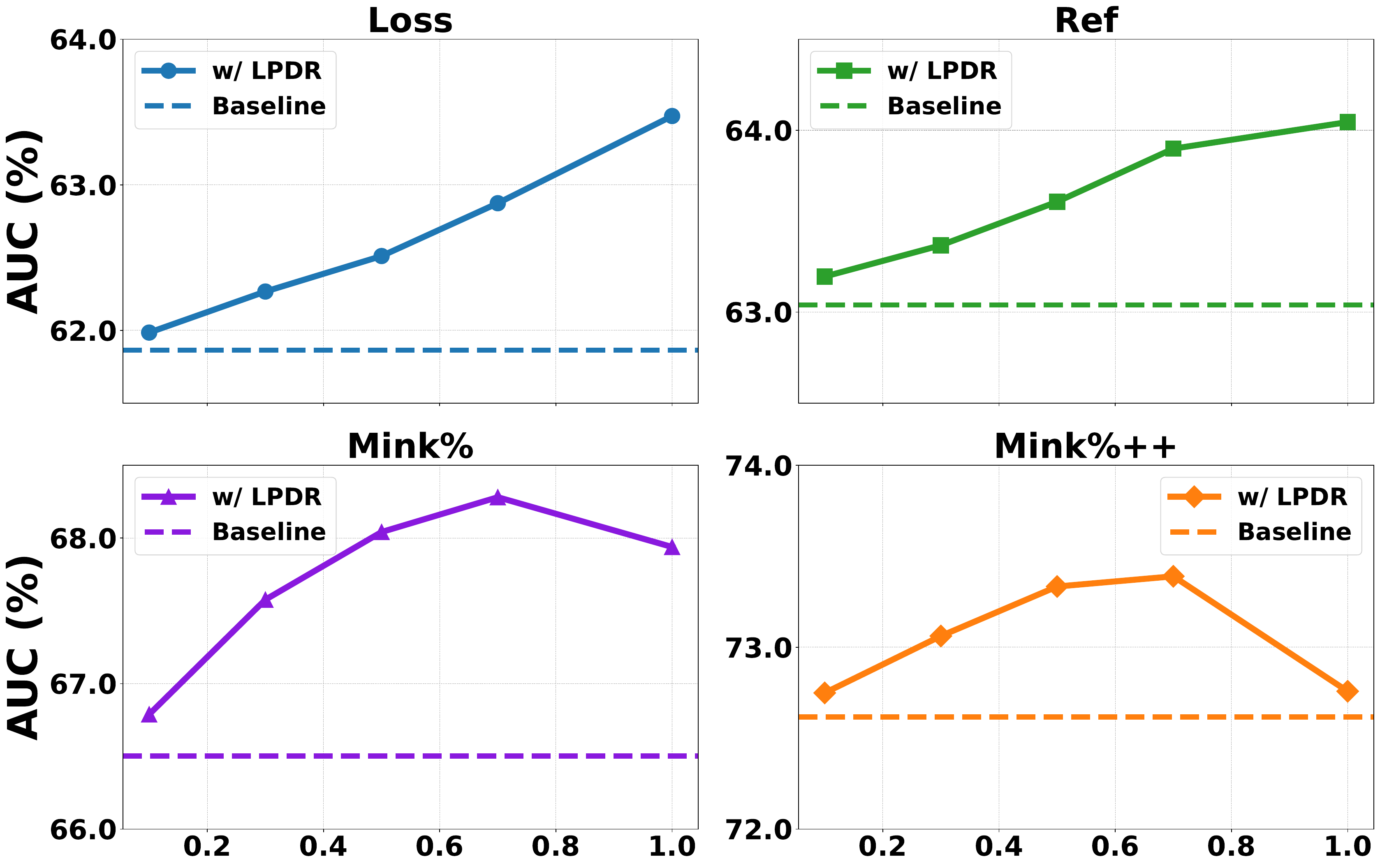}
    \caption{\small{AUC comparison of different $\mathbf{\alpha}$ for LPDR on log-likelihood methods (Loss, Ref, Min-$k$\%, Min-$k$\%++), results from Pythia-12B model at WikiMIA \textit{ori.} of 64 length.}}
    \label{fig:weights_ablation}
    \vspace{-20pt}
\end{figure}

\subsection{Further Analysis}
\textbf{Ablation Study about Weight Design.} 
We conduct an ablation study with several alternative weighting schemes. We consider applying PDR weights in different orders: \textit{Random} (a random shuffled sequence) and \textit{Reverse} (a monotonically increasing sequence). We also evaluate token-level entropy weights derived from a single \textit{Sample} or the entire \textit{Dataset}. Additionally, following CAMIA~\citep{CAMIA}, we explore a dynamic strategy using a \textit{Fitted Slope} from each sample's loss sequence as the decay parameter $\alpha$ (details in Appendix~\ref{apd:baselines}). For Min-$k$\% methods, we compare our standard reweighting \textit{after} selection against reweighting the full sequence \textit{Before} selection.


As shown in Tab.~\ref{tab:ab_ablation_weights}, \textit{Random} and \textit{Reverse} orders degrade performance, confirming the necessity of monotonically decreasing weights. For Min-$k$\%, reweighting \textit{after} selection proves superior by amplifying isolated signals. Regarding alternatives, \textit{Dataset}-level entropy is effective but impractical, while \textit{Sample}-level entropy and the \textit{Fitted Slope} suffer from high volatility, yielding marginal gains. Thus, PDR strikes a critical balance as a simple, robust, and effective prior.

\begin{figure}[t]
    \centering
    \begin{minipage}{0.48\textwidth}
        \centering
        \includegraphics[width=\linewidth]{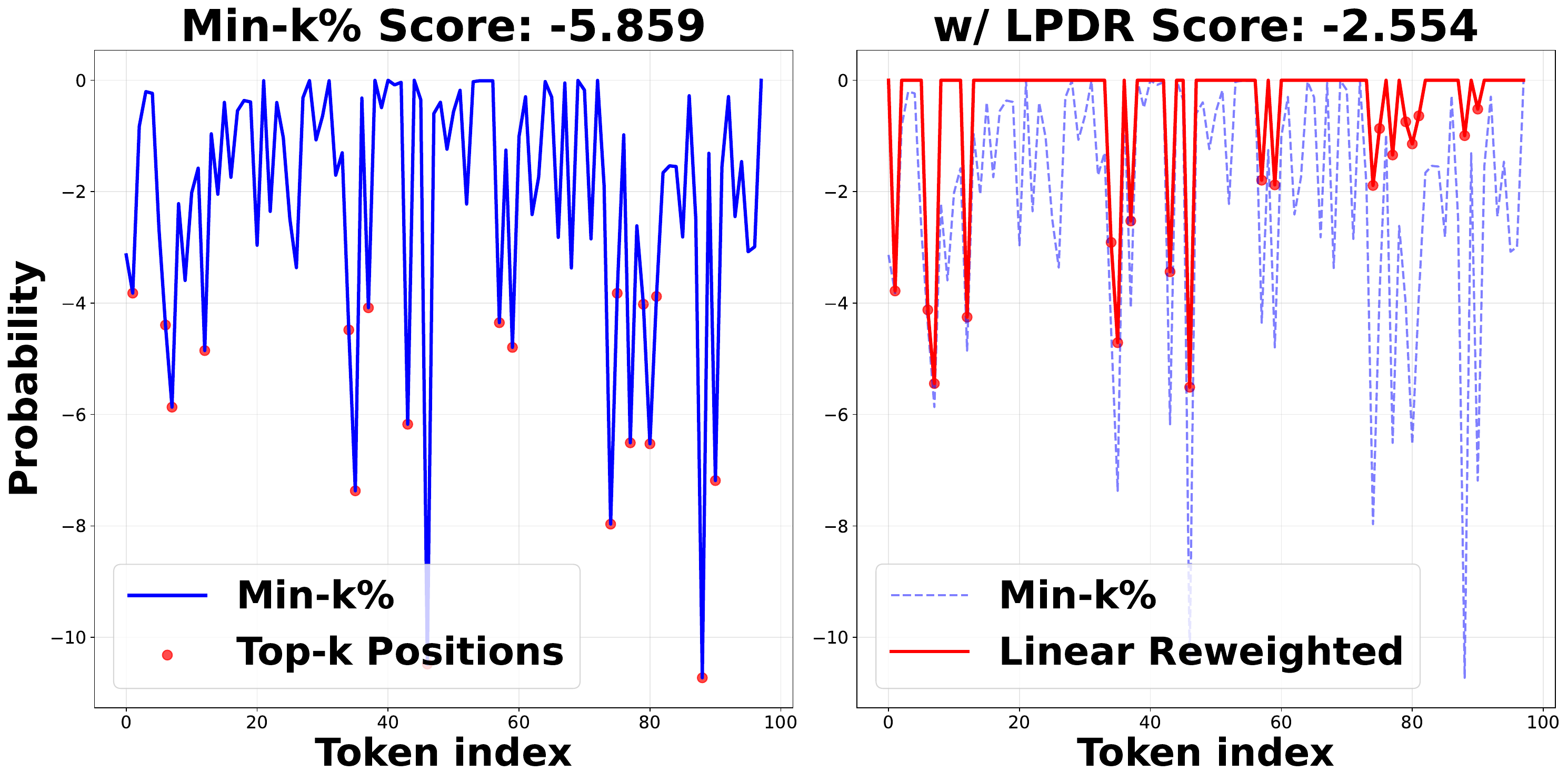}
        \centerline{\small{(a) Member sample}}
    \end{minipage}
    \begin{minipage}{0.48\textwidth}
        \centering
        \includegraphics[width=\linewidth]{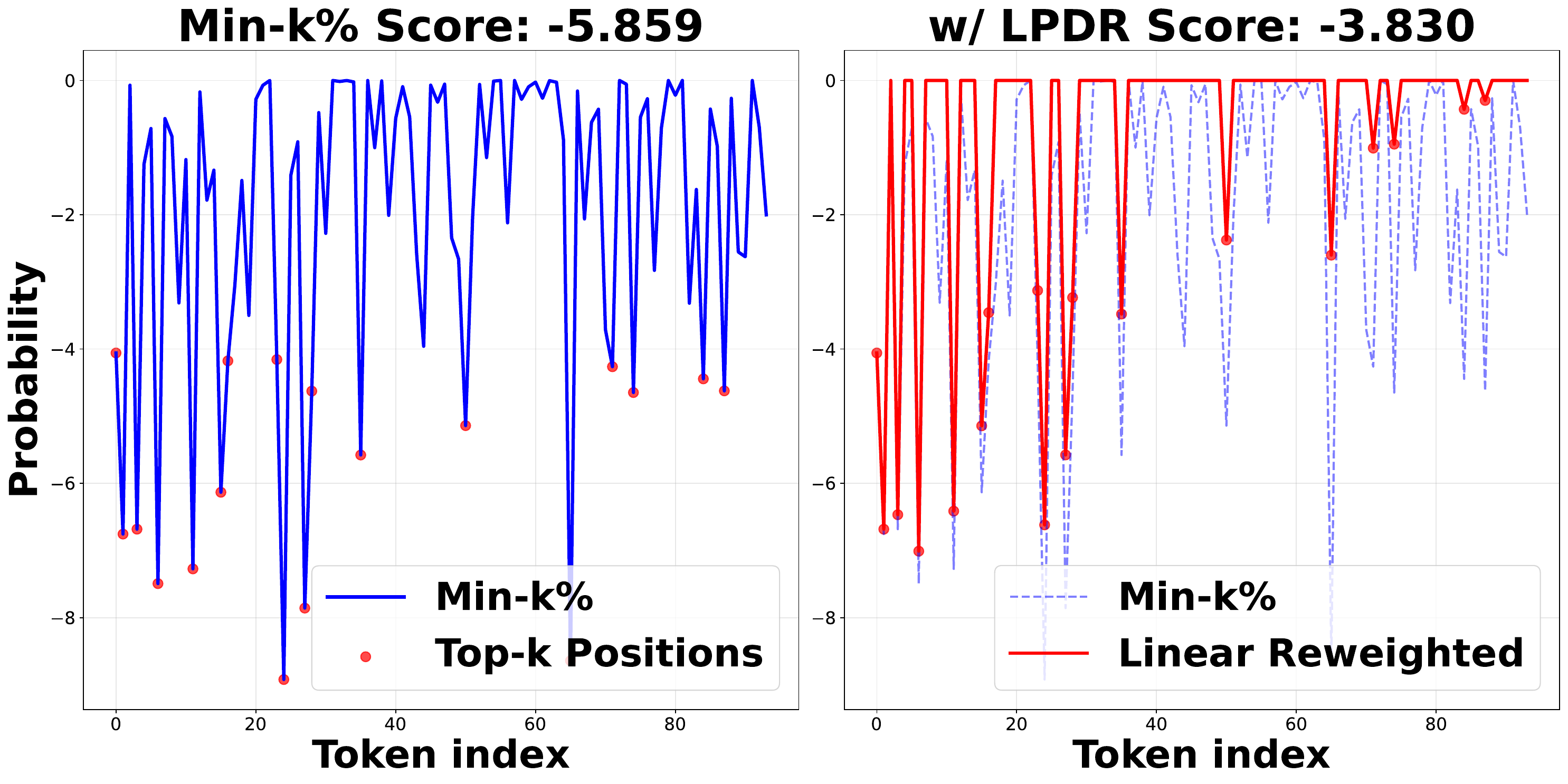}
        \centerline{\small{(b) Non-member sample}}
    \end{minipage}
    \vspace{-10pt}
    \caption{\small{Token-level score changes for (a)  member sample and (b) non-member sample after applying LPDR to  Min-$k$\%. The red dot means the selected token by Min-k\%, blue and red line denote the original and reweighted token-level score.} }  
    \label{fig:reweighting_effect_comparison}
    \vspace{-20pt}
\end{figure}

\textbf{The Effect of Sharpness in Weight Decay.} 
As shown in Fig.~\ref{fig:weights_ablation}, LPDR consistently outperforms baselines across various decay rates $\alpha$. 
While full-sequence methods (Loss, Ref) peak with sharp decay ($\alpha=1$) and outlier-based methods (Min-$k$\%, Min-$k$\%++) favor smoother decay, $\alpha=1$ remains a robust default delivering substantial gains without method-specific tuning. 
Analyses of weight functions and truncation ratios in Appendix~\ref{apd:weight_decay_sec} and \ref{apd:truncation_analysis} further confirm PDR's robustness across settings.

\textbf{Score Changes when Using Our Method.}
To further illustrate the mechanism of PDR, we selected a pair of member and non-member samples that are indistinguishable within Min-$k$\%  for sharing the same score. Fig.~\ref{fig:reweighting_effect_comparison} visualizes how our LPDR method resolves this ambiguity. 
By applying a monotonically decreasing weight, PDR enhances the importance of tokens at earlier positions and decreases the importance of tokens at later positions. 
For the member sample (a), whose memorization signals (high-probability tokens) are concentrated at the beginning, this reweighting process significantly amplifies its final score. In contrast, the non-member sample (b) is less affected. Thus, PDR effectively breaks the tie, creating a clear distinction between the two samples and enhancing overall detection accuracy. {For additional examples, detailed sample analysis, and score distribution visualizations, refer to Appendices~\ref{apd:sample_reweighted_analysis}, \ref{apd:analysis_of_selected_token}, and \ref{apd:score_distribution}.}


\section{Conclusion}

\label{sec:conclusion}
In this paper, we address a critical oversight in likelihood-based MIA methods: the failure to account for the positional nature of memorization signals. Based on the observation that a model's predictive uncertainty decreases as a sequence progresses, we argue that membership signals are strongest in early tokens. Existing methods dilute these signals by treating all positions equally. We introduce Positional Decay Reweighting (PDR), a simple, plug-and-play framework that applies a monotonically decreasing weight to token scores, enhancing existing MIA methods without requiring model changes. Extensive experiments validate that PDR significantly improves the performance of strong baselines like Min-$k$\% and Min-$k$\%++, both standalone and within advanced frameworks like FSD. Our work provides a more robust approach to membership inference by systematically prioritizing more reliable positional signals, contributing to a deeper understanding of privacy risks in LLMs.
\clearpage
\section{Limitations}
\label{sec:limitations}

We acknowledge several limitations in our study that could influence the generalization and applicability of our findings.



\textbf{Evaluation Language Limitation.} Our evaluations primarily focus on English-language data, following established benchmarks like WikiMIA and MIMIR. While the principle of positional entropy decay is likely universal across auto regressive models regardless of language, its practical effectiveness may vary. For instance, tokenization granularity differs significantly across languages, which could alter positional information distribution and the ideal decay shape. Furthermore, unique linguistic structures or cultural contexts might create different memorization patterns that are less susceptible to a uniform decay model. Assessing how PDR generalizes to multilingual settings, code-switched data, or low-resource languages remains an important direction for future research to confirm its universality.



\textbf{Black-box Constraints and Dataset Complexity.} We primarily operate in a black-box scenario where only the logits corresponding to the input text are available~\citep{mia_mink,mia_mink_plus}. This setting, while realistic for external auditors, is highly restrictive compared to white-box or gray-box access where gradients or model parameters could be used for more powerful attacks. This restriction limits the information leverageable for attacks, making it difficult to train effective auxiliary models without access to additional in-distribution samples. Furthermore, on complex benchmarks like MIMIR~\citep{mimir}, the high degree of data homogeneity between members and non-members means the statistical signals for membership are extremely subtle. For black-box methods that rely on summary statistics like loss, these signals are often too weak to achieve strong discrimination, posing a significant challenge for detection.


\section{Ethical Considerations}
\label{sec:ethics}

We undertake this study with a strong commitment to ethical research practices and responsible disclosure. By transparently communicating our methodology, findings, and limitations, we aim to raise awareness about privacy vulnerabilities associated with Large Language Models (LLMs).

\textbf{Privacy Research Purpose.} Our goal is to contribute constructively to the broader community by highlighting how easily memorized data can be exposed, even by simple reweighting strategies. Understanding these vulnerabilities is a prerequisite for developing stronger defenses, such as differential privacy training or advanced unlearning techniques, aligning our efforts with ongoing global initiatives for AI safety and privacy.

\textbf{Data and Model Usage.} Our study exclusively utilizes publicly available datasets (WikiMIA, MIMIR) and open-weights models (Pythia, LLaMA, OPT, etc.). These resources were employed strictly within their intended research purposes and license terms. Our research did not involve the collection of private user data or human subjects. To the best of our knowledge, the artifacts employed do not include personally identifiable information (PII) beyond what is already publicly present in the training corpora of these open models.

\textbf{Potential for Misuse.} We acknowledge that Membership Inference Attacks can potentially be misused to compromise privacy. However, the methods discussed herein are post-hoc auditing tools designed to evaluate model safety. We believe that the benefits of exposing these risks to the research community—thereby accelerating the development of mitigation strategies—outweigh the risks of disclosure.

\textbf{Tools and AI Assistance.} For data preprocessing, modeling, and evaluation tasks, we employed widely-used, open-source software packages including Hugging Face Transformers and standard Python libraries such as NumPy and SciPy. Model-specific tokenizers and default parameter configurations were used unless explicitly stated otherwise. Lastly, AI assistants (e.g., ChatGPT) were utilized to revise this manuscript. All generated content was rigorously reviewed and finalized by the authors.

\bibliography{custom}
\clearpage
\appendix
\section{Likelihood-Based Score funcions}
\label{apd:likelihood-based_score_function}

This section provides the formal definitions for the baseline likelihood-based MIA score functions discussed in the main paper. For a given input sequence $\xv = \{x_1, \dots, x_T\}$ of length $T$, these methods compute a score based on the token-level log-probabilities produced by the target model $P$.

\textbf{Loss.}~\citep{mia_loss} The standard Loss-based method, also known as Negative Log-Likelihood (NLL), uses the average log-probability of a sequence as its score. A higher score (lower loss) is indicative of membership. The score is defined as:
\begin{equation}
S_{\text{Loss}}(\xv) = \frac{1}{T} \sum_{t=1}^{T} \log P(x_t|x_{<t})
\end{equation}

\textbf{Ref.}~\citep{mia_carlini} The Reference-based method (Ref) calibrates the target model's likelihood by subtracting the log-likelihood from a smaller reference model ($P_{\text{ref}}$). This helps to normalize for tokens that are inherently common or easy to predict. The score is:
\begin{equation}
S_{\text{Ref}}(\xv) = \frac{1}{T} \sum_{t=1}^{T} \left( \log P(x_t|x_{<t}) - \log P_{\text{ref}}(x_t|x_{<t}) \right)
\end{equation}

\textbf{Min-$k$\%.}~\citep{mia_mink} The Min-$k$\% method operates on the assumption that member samples have fewer "outlier" tokens with very low probabilities. It computes the score by averaging only the lowest $k$\% log-probabilities in the sequence. Let $\mathcal{S}_{k}$ be the set of token positions corresponding to the lowest $k$\% log-probabilities. The score is:
\begin{equation}
S_{\text{Min-}k\text{\%}}(\xv) = \frac{1}{|\mathcal{S}_{k}|} \sum_{t \in \mathcal{S}_{k}} \log P(x_t|x_{<t})
\end{equation}

\textbf{Min-$k$\%++.}~\citep{mia_mink_plus} The Min-$k$\%++ method enhances Min-$k$\% by first normalizing the log-probability at each position $t$ using pre-computed mean ($\mu_t$) and standard deviation ($\sigma_t$) statistics for that position. This accounts for positional biases in the model's predictions. Let the normalized score be $z_t = (\log P(x_t|x_{<t}) - \mu_t) / \sigma_t$. Let $\mathcal{S}_{k}$ be the set of positions corresponding to the lowest $k$\% normalized scores $z_t$. The final score is:
\begin{equation}
S_{\text{Min-}k\text{\%++}}(\xv) = \frac{1}{|\mathcal{S}_{k}|} \sum_{t \in \mathcal{S}_{k}} z_t
\end{equation}

\textbf{FSD.}~\citep{zhang2025finetuning} Finetuning-based Score Difference (FSD) is a framework that enhances any base scoring function $S(\cdot)$. It computes the difference between the score from the original model ($M$) and the score from a model fine-tuned on non-member data ($M'$). A larger difference suggests membership.
\begin{equation}
S_{\text{FSD}}(\xv) = S(\xv; M) - S(\xv; M')
\end{equation}

\section{Likelihood-Based Score Functions with PDR}
\label{apd:likelihood-based_score_function_with_pdr}
This section details how our Position Difference Reweighting (PDR) method is integrated with various likelihood-based MIA score functions. For a given input sequence $\xv = \{x_1, \dots, x_T\}$ of length $T$, PDR introduces a positional weight $w(t)$ for each token $x_t$ at position $t$. The final score is then computed based on the weighted combination of token-level scores.

\textbf{PDR-Loss.} The standard Loss-based method, often conceptualized as Negative Log-Likelihood (NLL), uses the average log-probability of a sequence as its score. With PDR, we apply positional weights to the log-probabilities of each token before averaging. A lower weighted loss (which corresponds to a higher score) suggests the sequence is a member. The PDR-enhanced score is:
\begin{equation}
S_{\text{PDR-Loss}}(\xv) = \frac{1}{T} \sum_{t=1}^{T} w(t) \cdot \log P(x_t|x_{<t})
\end{equation}

\textbf{PDR-Ref.} The Reference-based method (Ref) calibrates the target model's likelihood by subtracting the log-likelihood from a smaller reference model ($P_{\text{ref}}$). PDR is applied to the resulting difference at each position. The score is defined as:
\begin{equation}
\begin{split}
S_{\text{PDR-Ref}}(\mathbf{x}) = \frac{1}{T} \sum_{t=1}^{T} w(t) \cdot \Big( & \log P(x_t|x_{<t}) \\
& - \log P_{\text{ref}}(x_t|x_{<t}) \Big)
\end{split}
\end{equation}

\begin{algorithm*}[htp]
\caption{Overall algorithm for applying PDR to different logit-based MIA methods}
\label{alg:pdr_algorithm}
\begin{algorithmic}[1]
\State \textbf{Input:} 
\Statex \quad Test dataset $\mathcal{D} = \{\xv^1, \dots, \xv^N\}$; 
\Statex \quad Target model's predictive distribution $P(x_{t}|x_{<t})$;
\Statex \quad A chosen decay function $f_{\text{decay}} \in \{\text{Linear}, \text{Exponential}, \text{Polynomial}\}$; 
\Statex \quad Decay hyperparameter $\alpha$ or $p$; 
\Statex \quad A base MIA scoring method $\mathcal{M} \in \{\text{Loss}, \text{Ref}, \text{Min-}K\%, \text{Min-}K\%\text{++}\}$;
\Statex \quad \textit{(Optional)} Reference model $P_{\text{ref}}(x_{t}|x_{<t})$;
\Statex \quad \textit{(Optional)} Positional normalization stats $\{\mu_t, \sigma_t\}_{t=1}^T$.
\State \textbf{Output:} A list of PDR-enhanced scores $\mathcal{S}_{\text{PDR}} = \{s_1, \dots, s_N\}$.

\State Initialize an empty list $\mathcal{S}_{\text{PDR}}$.

\For{$i = 1$ to $N$} \Comment{Iterate over all sequences in the dataset}
    \State $\xv^i = \{\xv^i_1, \dots, \xv^i_T\}$
    \Statex \textit{// Step 1: Compute Positional Weights for the current sequence}
    \For{$t = 1$ to $T$} \Comment{Iterate over all positions in the sequence}
        \If{$f_{\text{decay}}$ is Linear}
            \State $w(t) \gets 1 - \alpha \cdot \frac{t-1}{T-1}$
        \ElsIf{$f_{\text{decay}}$ is Exponential}
            \State $w(t) \gets \exp(-\alpha \cdot (t-1))$
        \ElsIf{$f_{\text{decay}}$ is Polynomial}
            \State $w(t) \gets \left(1 - \frac{t-1}{T-1} \right)^p$
        \EndIf
    \EndFor

    \Statex \textit{// Step 2: Apply PDR to the chosen base MIA method}
    \State Initialize current score $s^{i} \gets 0$.
    \If{$\mathcal{M}$ is Loss}
        \State $s^{i} \gets -\frac{1}{T} \sum_{t=1}^{T} w(t) \cdot \log P(x^i_{t}|x^i_{<t})$
    \ElsIf{$\mathcal{M}$ is Ref}
        \State $s^{i} \gets \frac{1}{T} \sum_{t=1}^{T} w(t) \cdot \left( \log P(x^i_{t}|x^i_{<t}) - \log P_{\text{ref}}(x^i_{t}|x^i_{<t}) \right)$
    \ElsIf{$\mathcal{M}$ is Min-$k$\%}
        \State Let $\mathcal{S}_{k}$ be the set of token positions with the smallest $k$\% log-probabilities.
        \State $s^{i} \gets \frac{1}{|\mathcal{S}_{k}|} \sum_{t \in \mathcal{S}_{k}} w(t) \cdot \log P(x^i_{t}|x^i_{<t})$
    \ElsIf{$\mathcal{M}$ is Min-$k$\%\text{++}}
        \State For each $t$, compute normalized score $z_t = \frac{\log P(x^i_t|x^i_{<t}) - \mu_t}{\sigma_t}$.
        \State Let $\mathcal{S}_{k}$ be the set of token positions with the smallest $k$\% normalized scores $z_t$.
        \State $s^{i} \gets \frac{1}{|\mathcal{S}_{k}|} \sum_{t \in \mathcal{S}_{k}} w(t) \cdot z_t$
    \EndIf
    \State Append $s^i$ to $\mathcal{S}_{\text{PDR}}$.
\EndFor

\State \textbf{return} $\mathcal{S}_{\text{PDR}}$
\end{algorithmic}
\end{algorithm*}

\textbf{PDR-Min-$k$\%.} Following the standard Min-$k$\% procedure, we first identifies the token positions corresponding to the lowest $k$\% log-probabilities. Then computes the final score by taking a weighted average of the log-probabilities at only these selected positions, where each score is multiplied by its corresponding positional weight $w(t)$. Let $\mathcal{S}_{k}$ be the set of token positions corresponding to the lowest $k$\% values of $\{\log P(x_t|x_{<t})\}_{t=1}^T$. The score is:
\begin{equation}
S_{\text{PDR-Min-}k\text{\%}}(\xv) = \frac{1}{|\mathcal{S}_{k}|} \sum_{t \in \mathcal{S}_{k}} w(t) \cdot \log P(x_t|x_{<t})
\end{equation}

\textbf{PDR-Min-$k$\%++.} Similarly, for Min-$k$\%++, we first identify the positions of the lowest $k$\% normalized z-scores. The PDR-enhanced score is then the weighted average of these selected z-scores, with positional weights applied before averaging. Let $z_t = (\log P(x_t|x_{<t}) - \mu_t) / \sigma_t$, and let $\mathcal{S}_{k}$ be the set of positions for the lowest $k$\% values of $\{z_t\}_{t=1}^T$. The score is:
\begin{equation}
S_{\text{PDR-Min-}k\text{\%++}}(\xv) = \frac{1}{|\mathcal{S}_{k}|} \sum_{t \in \mathcal{S}_{k}} w(t) \cdot z_t
\end{equation}

\textbf{PDR-FSD.} Finetuning-based Score Difference (FSD) calculates the difference between a score function $S(\cdot)$ computed before and after fine-tuning the model on non-member data. Our PDR method can be applied to the base score function $S(\cdot)$ used within the FSD framework. If we denote the fine-tuned model as $M'$, the FSD score using a PDR-enhanced base method $S_{\text{+PDR}}$ is:
\begin{equation}
S_{\text{PDR-FSD}}(\xv) = S_{\text{+PDR}}(\xv; M) - S_{\text{+PDR}}(\xv; M')
\end{equation}
where $S_{\text{+PDR}}(\xv; M)$ and $S_{\text{+PDR}}(\xv; M')$ are the PDR-enhanced scores computed using the original model $M$ and the fine-tuned model $M'$, respectively.

The detailed process of applying our PDR method to various likelihood-based MIA methods is outlined in Algorithm~\ref{alg:pdr_algorithm}. This algorithm specifically illustrates the computation for combining PDR with Loss, Ref, Min-$k$\%, and Min-$k$\%++.

\textbf{PDR-Min-$k$\%.} Following the standard Min-$k$\% procedure, we first identifies the token positions corresponding to the lowest $k$\% log-probabilities. Then computes the final score by taking a weighted average of the log-probabilities at only these selected positions, where each score is multiplied by its corresponding positional weight $w(t)$. Let $\mathcal{S}_{k}$ be the set of token positions corresponding to the lowest $k$\% values of $\{\log P(x_t|x_{<t})\}_{t=1}^T$. The score is:
\begin{equation}
S_{\text{PDR-Min-}k\text{\%}}(\xv) = \frac{1}{|\mathcal{S}_{k}|} \sum_{t \in \mathcal{S}_{k}} w(t) \cdot \log P(x_t|x_{<t})
\end{equation}

\textbf{PDR-Min-$k$\%++.} Similarly, for Min-$k$\%++, we first identify the positions of the lowest $k$\% normalized z-scores. The PDR-enhanced score is then the weighted average of these selected z-scores, with positional weights applied before averaging. Let $z_t = (\log P(x_t|x_{<t}) - \mu_t) / \sigma_t$, and let $\mathcal{S}_{k}$ be the set of positions for the lowest $k$\% values of $\{z_t\}_{t=1}^T$. The score is:
\begin{equation}
S_{\text{PDR-Min-}k\text{\%++}}(\xv) = \frac{1}{|\mathcal{S}_{k}|} \sum_{t \in \mathcal{S}_{k}} w(t) \cdot z_t
\end{equation}

\textbf{PDR-FSD.} Finetuning-based Score Difference (FSD) calculates the difference between a score function $S(\cdot)$ computed before and after fine-tuning the model on non-member data. Our PDR method can be applied to the base score function $S(\cdot)$ used within the FSD framework. If we denote the fine-tuned model as $M'$, the FSD score using a PDR-enhanced base method $S_{\text{+PDR}}$ is:
\begin{equation}
S_{\text{PDR-FSD}}(\xv) = S_{\text{+PDR}}(\xv; M) - S_{\text{+PDR}}(\xv; M')
\end{equation}
where $S_{\text{+PDR}}(\xv; M)$ and $S_{\text{+PDR}}(\xv; M')$ are the PDR-enhanced scores computed using the original model $M$ and the fine-tuned model $M'$, respectively.

The detailed process of applying our PDR method to various likelihood-based MIA methods is outlined in Algorithm~\ref{alg:pdr_algorithm}. This algorithm specifically illustrates the computation for combining PDR with Loss, Ref, Min-$k$\%, and Min-$k$\%++.

\section{Experiments Setting details}
\label{apd:exp_settings}
\subsection{Benchmarks}
\label{apd:benchmarks}
We focus on two commonly-used benchmarks for pre-training data detection:
(1) \textbf{WikiMIA}~\citep{mia_mink} is the first benchmark for pre-training data detection, comprising texts from Wikipedia events. The distinction between training and non-training data is established based on temporal timestamps. To enable fine-grained evaluation, WikiMIA organizes data into splits according to sentence length. It also includes two evaluation settings: the \textit{original} setting evaluates the detection of verbatim training texts, while the \textit{paraphrased} setting uses ChatGPT to paraphrase training texts and evaluates on paraphrased inputs.

(2) \textbf{MIMIR}~\citep{mimir} is built upon the Pile dataset~\citep{gao2020pile}. This benchmark poses greater challenges compared to WikiMIA, as the shared dataset origin between training and non-training texts eliminates substantial distribution shifts and temporal discrepancies~\citep{mimir}.

\subsection{Baselines}
\label{apd:baselines}
We consider several representative methods as our baselines:
\begin{itemize}
    \item \textbf{Loss}~\citep{mia_loss} is a general technique that directly uses the loss of the model as the detection score. 
    \item \textbf{Ref}~\citep{mia_carlini} employs an additional, typically smaller, language model as a reference to calibrate the likelihood of the input text.
    \item \textbf{zlib} and \textbf{lowercase}~\citep{mia_carlini} use the compression entropy of zlib and the likelihood of the lowercase text as references to calibrate the likelihood.
    \item \textbf{Min-$k$\%}~\citep{mia_mink} examines the exact probabilities of the token and averages a subset of the lowest token scores from the input sequence.
    \item \textbf{Min-$k$\%++}~\citep{mia_mink_plus} extends Min-$k$\% by standardizing the log-probability of each token using the mean and standard deviation of log-probabilities at that specific position, making scores more comparable across different positions before applying the Min-$k$\% selection. 
    \item \textbf{FSD}~\citep{zhang2025finetuning} involves fine-tuning the model on non-member samples and using the difference in logit-based scores before and after fine-tuning for detection. 
    \item \textbf{CAMIA}~\citep{CAMIA} esigns multiple dynamic signals for detection, we explore a variant based on one of its key signals: the loss decreasing rate. Specifically, for each sample $\xv$, we compute a slope by performing a linear regression of its token-level losses $L_t(x_t)$ against their positions $t$. The slope is calculated as:
\begin{equation}
f_{\text{Slope}}(\xv) = \frac{\sum_{t=1}^{T} (t - \bar{t})(L_t(x_t) - \bar{L})}{\sum_{t=1}^{T} (t - \bar{t})^2}
\end{equation}
with $\bar{t} = \frac{T+1}{2}$ and $\bar{L} = \frac{1}{T} \sum_{t=1}^{T} L_t(x_t)$. We then use this dynamically computed current sample's slope sample as the decay parameter $\alpha$ in our reweighting scheme, referring to this method a w/ LDPR (CAMIA slope).
\end{itemize}

\subsection{Environment}
All experiments were conducted on the Ubuntu 20.04.4 LTS operating system, Intel(R) Xeon(R) Gold 5220 CPU @ 2.20GHz with a single NVIDIA A40 48GB GPU and 512GB of RAM. The framework is implemented with Python 3.9.0 and PyTorch 2.6.0. Other key packages include transformer 4.40.1, numpy 1.24 and accelerate 0.26.0.

\subsection{Models}
This section details the specific models used in our experiments. For the Ref method, the choice of reference model depends on the dataset following~\citep{mia_carlini, mia_mink, mia_mink_plus}.On the \textbf{WikiMIA} dataset, we used the following reference models for different model families:
\begin{itemize}
    \item For the \textbf{Pythia} family, we used \texttt{Pythia-70M}.
    \item For the \textbf{Llama} family, we used \texttt{Llama-7B}.
    \item For the \textbf{GPT} family, we used \texttt{GPT-Neo-125M}.
    \item For the \textbf{Mamba} family, we used \texttt{Mamba-130M}.
    \item For the \textbf{OPT} family, we used \texttt{OPT-350M}.
\end{itemize}

\subsection{Hyperparameter}
\label{apd:hyperparameter}
For Min-$k$\% and Min-$k$\%++, we consistently use $k=20\%$ following common practice. In our experiments, settings are as follows:
\begin{itemize}
    \item  \textbf{LPDR}: on the WikiMIA dataset, we generally set $\alpha=1.0$ for sequence lengths of 64 and 128. For the shorter length of 32, a smaller weight was preferred like 0.1 or 0.5. On the Mimir dataset, we set $\alpha=1.0$.
    \item \textbf{EPDR}: on the WikiMIA dataset, Ref and Min-$k$\%++ set $\alpha=0.02$, where as Loss and Min-$k$\% set $\alpha=0.002$. On Mimir, $\alpha=0.002$ was used.
    \item \textbf{PPDR}: on the WikiMIA dataset, we used $p=0.1$ for length 32, but a much steeper decay of $p=2.0$ for lengths 64 and 128. On Mimir, we set a gentler $p=0.1$.
\end{itemize}
A clear trend emerges from these results. On the WikiMIA dataset, longer sequences tend to benefit from more aggressive, steeper weight functions, while shorter sequences and the more challenging Mimir dataset favor gentler, more gradual decay.
\subsection{FSD Settings}
For the Finetuning-based Score Difference (FSD) experiments, we follow the implementation details from the original paper. To construct the non-member dataset for fine-tuning, we first randomly sample 30\% of the entire dataset. All non-member samples within this subset are then used as the fine-tuning dataset. The remaining 70\% of the data is reserved for testing. We use LoRA \citep{hu2022lora} to fine-tune the base model for 3 epochs with a batch size of 8. The initial learning rate is set to 0.001 and is adjusted using a cosine scheduling strategy. 

\begin{table}[H]
\centering
\caption{\small{Performance comparison on Pythia-6.9B across different sequence lengths. We report AUROC and TPR@0.5\%FPR (mean $\pm$ std). \textbf{w/ LPDR} denotes our method using linear weights. The p-value indicates the statistical significance of the improvement of LPDR over the corresponding baseline.}}
\vspace{-5pt}
\label{tab:wiki_pythia6.9}
\renewcommand{\arraystretch}{1.2}
\resizebox{0.98\linewidth}{!}{
\begin{tabular}{clccccc}
\toprule
\textbf{Length} & \textbf{Method} & \textbf{AUROC} & \textbf{AUROC Std} & \textbf{TPR@0.5\%FPR} & \textbf{TPR Std} & \textbf{p-value} \\
\midrule

\multirow{10}{*}{32} 
& Zlib & 64.2 & 2.0 & 12.7 & 2.6 & - \\
& Lowercase & 61.7 & 2.1 & 11.9 & 3.2 & - \\
\cline{2-7}
& Loss & 64.1 & 2.0 & 15.0 & 2.8 & - \\
\rowcolor{gray!15} \cellcolor{white} & \textbf{w/ LPDR} & 64.2 & 2.0 & 15.0 & 2.8 & 0.275 \\
\cline{2-7}
& Ref & 63.5 & 2.0 & 6.2 & 3.0 & - \\
\rowcolor{gray!15} \cellcolor{white} & \textbf{w/ LPDR} & 63.5 & 2.0 & 5.7 & 3.2 & 0.785 \\
\cline{2-7}
& Min-$k$\% & 65.1 & 2.0 & 21.7 & 3.5 & - \\
\rowcolor{gray!15} \cellcolor{white} & \textbf{w/ LPDR} & 65.1 & 2.0 & \textbf{22.0} & 4.0 & 0.342 \\
\cline{2-7}
& Min-$k$\%++ & 67.6 & 2.0 & 14.5 & 3.2 & - \\
\rowcolor{gray!15} \cellcolor{white} & \textbf{w/ LPDR} & \textbf{67.8} & 2.0 & 14.5 & 2.6 & 0.218 \\
\midrule

\multirow{10}{*}{64} 
& Zlib & 61.6 & 2.5 & 15.8 & 3.5 & - \\
& Lowercase & 57.7 & 2.5 & 11.3 & 2.2 & - \\
\cline{2-7}
& Loss & 59.3 & 2.5 & 10.6 & 3.7 & - \\
\rowcolor{gray!15} \cellcolor{white} & \textbf{w/ LPDR} & \textbf{62.4} & 2.5 & \textbf{13.0} & 5.4 & \textbf{0.001} \\
\cline{2-7}
& Ref & 62.9 & 2.4 & 16.2 & 2.8 & - \\
\rowcolor{gray!15} \cellcolor{white} & \textbf{w/ LPDR} & \textbf{64.0} & 2.3 & 7.7 & 3.6 & 0.152 \\
\cline{2-7}
& Min-$k$\% & 61.1 & 2.5 & 12.7 & 3.0 & - \\
\rowcolor{gray!15} \cellcolor{white} & \textbf{w/ LPDR} & \textbf{65.1} & 2.4 & \textbf{18.3} & 5.2 & \textbf{0.019} \\
\cline{2-7}
& Min-$k$\%++ & 64.2 & 2.4 & 10.2 & 3.2 & - \\
\rowcolor{gray!15} \cellcolor{white} & \textbf{w/ LPDR} & \textbf{68.3} & 2.4 & \textbf{12.7} & 4.6 & \textbf{0.006} \\
\midrule

\multirow{10}{*}{128} 
& Zlib & 67.4 & 3.4 & 20.9 & 5.8 & - \\
& Lowercase & 59.9 & 3.6 & 11.5 & 4.2 & - \\
\cline{2-7}
& Loss & 64.7 & 3.5 & 16.5 & 6.8 & - \\
\rowcolor{gray!15} \cellcolor{white} & \textbf{w/ LPDR} & \textbf{66.6} & 3.4 & 15.1 & 4.9 & 0.061 \\
\cline{2-7}
& Ref & 62.9 & 3.5 & 8.6 & 5.7 & - \\
\rowcolor{gray!15} \cellcolor{white} & \textbf{w/ LPDR} & \textbf{65.9} & 3.4 & \textbf{10.8} & 5.0 & \textbf{0.016} \\
\cline{2-7}
& Min-$k$\% & 67.0 & 3.4 & 16.5 & 6.1 & - \\
\rowcolor{gray!15} \cellcolor{white} & \textbf{w/ LPDR} & \textbf{68.9} & 3.3 & \textbf{23.0} & 9.4 & 0.270 \\
\cline{2-7}
& Min-$k$\%++ & 65.9 & 3.5 & 18.0 & 7.1 & - \\
\rowcolor{gray!15} \cellcolor{white} & \textbf{w/ LPDR} & \textbf{72.2} & 3.3 & \textbf{18.7} & 6.9 & \textbf{0.001} \\
\bottomrule

\end{tabular}
}
\end{table}

\section{WikiMIA Results}
\subsection{WikiMIA Results on fixed $\alpha$}
\label{apd:wikimia results}
This section presents the comprehensive results on the WikiMIA benchmark. We report detailed AUROC and TPR scores across all models and sequence lengths under both the original and paraphrased settings. The AUROC results are shown in Tab.~\ref{tab:wikimia ori. auc} and Tab.~\ref{tab:wikimia para. auc}, respectively, while the corresponding TPR results are reported in Tab.~\ref {tab:wikimia ori. tpr} and Tab.~\ref{tab:wikimia para. tpr}.



\subsection{ROC curve Visualization}
\label{apd:auroc_curve_visualization}
\begin{figure*}[htbp]
     \centering
    \begin{minipage}{0.5\textwidth}
        \centering
        \includegraphics[width=\linewidth]{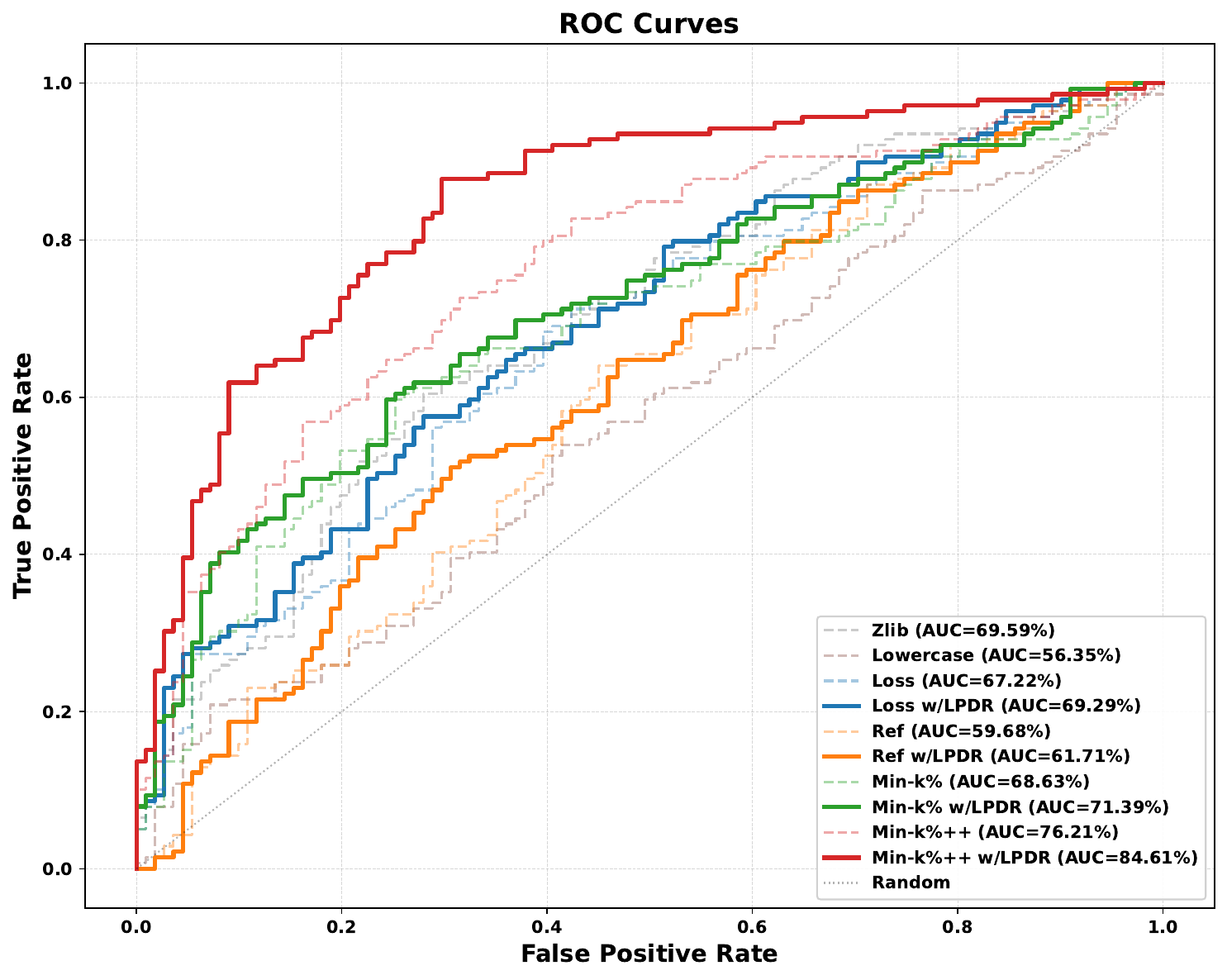}
        \centerline{(a) Llama-13B (WikiMIA-128 Para.)}
    \end{minipage}\hfill 
    \begin{minipage}{0.5\textwidth}
        \centering
        \includegraphics[width=\linewidth]{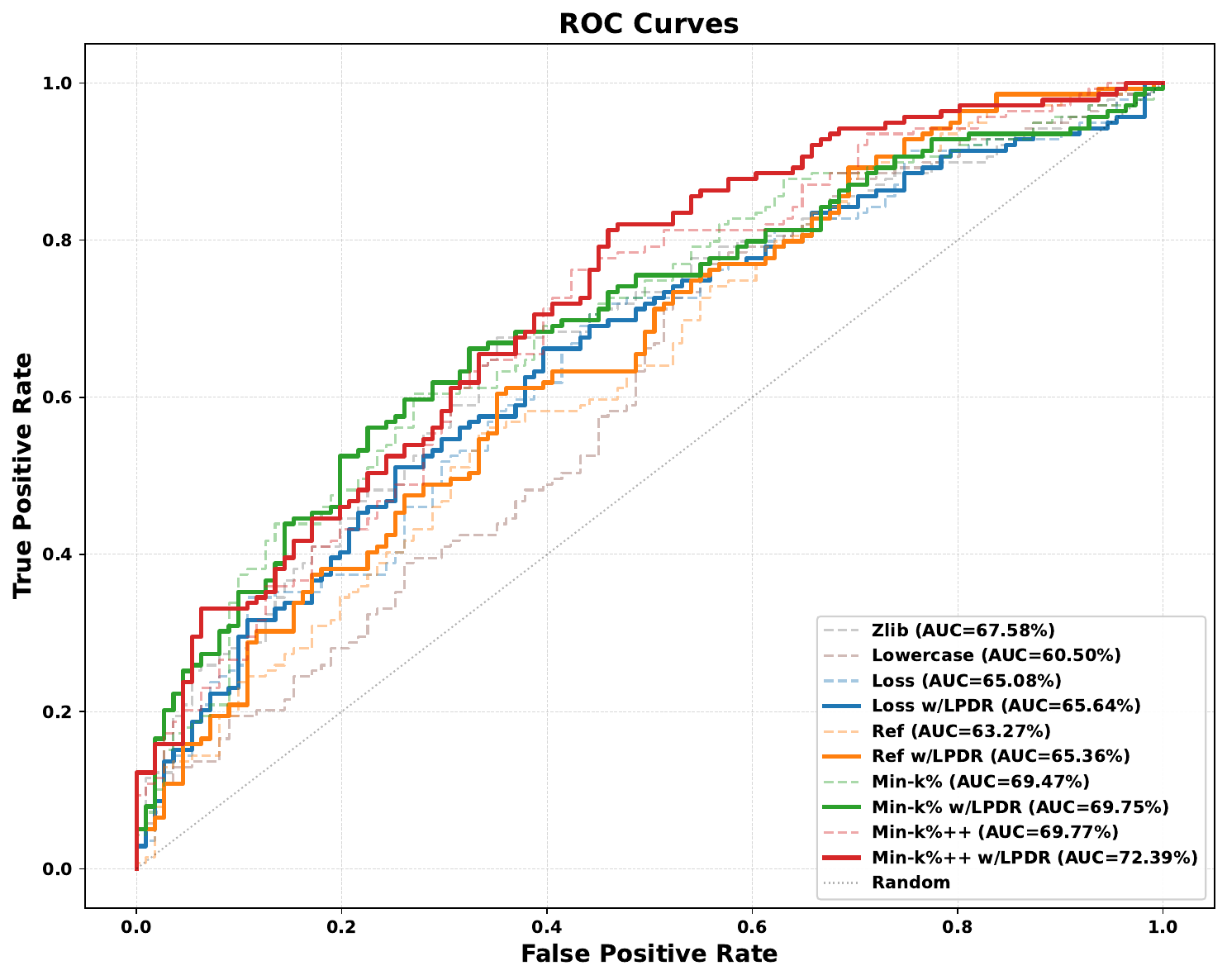}
        \centerline{(b) Pythia-6.9B (WikiMIA-128 Ori.)}
    \end{minipage}
    \caption{{ROC curve comparison for various baseline methods and their PDR-enhanced versions on the WikiMIA-128 benchmark.} }
    \label{fig:roc_visualization}
\end{figure*}

{This section provides ROC curve visualizations to offer a more detailed view of our method's performance. Figure~\ref{fig:roc_visualization} plots the ROC curves for several baseline methods and their LPDR-enhanced counterparts on the WikiMIA benchmark (length 128). 
Specifically, we show results for (a) Llama-13B on the paraphrased setting and (b) Pythia-6.9B on the original setting.
As illustrated, the PDR-enhanced methods consistently offer a more favorable trade-off, achieving a higher True Positive Rate (TPR) for any given False Positive Rate (FPR). This enhanced discriminative capability helps to explain the AUROC gains reported throughout the paper.}

\begin{table*}[htbp]
    \centering
    \vspace{-5pt}
    \caption{AUC-ROC on WikiMIA benchmark under original setting. w/ LPDR utilizes linear weights for reweighting, w/ EPDR utilizes exponential weights for reweighting, w/ PPDR utilizes polynomial weights for reweighting.{$^\dag$Neighbor results are from \citet{mia_mink_plus}.}}
\label{tab:wikimia ori. auc}
\renewcommand{\arraystretch}{1.3}
\resizebox{0.95\textwidth}{!}{%
    \begin{tabular}{c| c |c c c |c c c c c| c c c c c| c c c c c |c c c c c }
      
    \toprule
        Length & Models & Lowercase & Zlib & {$^\dag$Neighbor} & Loss & \small{w/ LDPR (CAMIA slope)} & w/ LPDR & w/ EPDR & w/ PPDR & Ref & \small{w/ LDPR (CAMIA slope)} & w/ LPDR &  w/ EPDR & w/ PPDR & Min-$k$\% &  \small{w/ LDPR (CAMIA slope)} & w/ LPDR &  w/ EPDR &  w/ PPDR &  Min-$k$\% ++ & \small{w/ LDPR (CAMIA slope)} & w/ LPDR & w/ EPDR & w/ PPDR \\   \midrule
    \multirow{10}{*}{32} 
        &Mamba-1.4B & 60.9  & 61.9 & 64.1 & 61.0  & 61.1 & 61.5  & 60.9  & 61.2  & 62.2  & 62.0 & 62.2  & 62.7  & 62.2  & 63.3  & 63.2 & 63.5  & 63.2  & 63.8  & 66.4  & 66.6 & 67.4  & 66.6  & 67.1  \\ 
        & Mamba-2.8B & 63.6  & 64.7 & 67.0 & 64.1  & 64.2 & 64.5  & 64.0  & 64.2  & 67.0  & 66.6 & 66.9  & 67.1  & 66.7  & 66.1  & 66.1 & 66.2  & 66.0  & 66.1  & 69.0  & 69.1 & 69.4  & 68.9  & 69.1  \\ 
        & Pythia-2.8B & 60.9  & 62.1 & 64.2 & 61.4  & 61.4 & 61.7  & 61.2  & 61.4  & 61.3  & 61.2 & 61.3  & 62.4  & 61.0  & 61.7  & 61.6 & 61.9  & 61.6  & 61.9  & 64.0  & 64.1 & 64.7  & 64.2  & 64.4  \\ 
        & Pythia-6.9B & 62.2  & 64.4 & 65.8 & 63.8  & 63.8 & 64.0  & 63.7  & 63.8  & 63.6  & 63.4 & 63.5  & 64.6  & 63.2  & 66.3  & 66.2 & 66.3  & 66.1  & 65.4  & 70.3  & 70.4 & 70.8  & 70.1  & 70.5  \\ 
        & Pythia-12B & 64.8  & 65.8 & 66.6 & 65.4  & 65.3 & 65.4  & 65.2  & 65.4  & 65.1  & 65.0 & 65.1  & 66.1  & 64.7  & 68.1  & 67.9 & 68.0  & 67.8  & 67.4  & 72.2  & 72.3 & 72.3  & 71.3  & 72.0  \\ 
        & Llama-13B & 64.0  & 67.8 & 65.8 & 67.5  & 67.5 & 67.7  & 67.5  & 67.5  & 57.9  & 57.8 & 57.8  & 57.2  & 57.6  & 66.8  & 66.8 & 66.8  & 66.8  & 66.6  & 84.4  & 84.4 & 85.9  & 86.2  & 85.0  \\ 
        & Llama-30B & 64.1  & 69.8 & 67.6 & 69.4  & 69.4 & 69.6  & 69.4  & 69.5  & 63.5  & 63.5 & 63.5  & 62.8  & 63.2  & 69.3  & 69.3 & 69.4  & 69.3  & 69.2  & 84.4  & 84.3 & 85.4  & 85.5  & 84.6  \\ 
        & OPT-66B & 62.8  & 65.8 & 68.2 & 65.6  & 65.5 & 65.6  & 65.5  & 65.7  & 68.6  & 68.5 & 68.6  & 69.3  & 68.4  & 67.5  & 67.4 & 67.7  & 67.4  & 67.5  & 69.7  & 69.8 & 70.2  & 69.4  & 70.0  \\ 
        & GPT-NeoX-20B & 68.3  & 69.3 & 70.2 & 69.1  & 68.9 & 68.9  & 68.9  & 69.0  & 67.6  & 67.0 & 67.4  & 66.7  & 67.0  & 72.2  & 71.9 & 72.0  & 71.8  & 71.1  & 75.1  & 75.2 & 75.2  & 74.4  & 75.0  \\
        \cdashline{2-25}
        & Average & 63.5  & 65.7 & 66.6 & 65.3  & 65.2 & \textbf{65.4}  & 65.1  & 65.3  & 64.1  & 63.9 & 64.0  & \textbf{64.3}  & 63.8  & 66.8  & 66.7 & \textbf{66.9}  & 66.7  & 66.5  & 72.8  & 72.9 & \textbf{73.5}  & \textbf{72.9}  & \textbf{73.1}  \\    
        \midrule
    \multirow{10}{*}{64} 
        & Mamba-1.4B & 57.0  & 60.4 & 60.6 & 58.2  & 58.4 & 59.7  & 58.2  & 60.7  & 60.6  & 60.7 & 61.1  & 61.9  & 60.3  & 61.7  & 62.0 & 62.9  & 62.0  & 62.8  & 67.2  & 67.4 & 68.2  & 67.9  & 68.2  \\ 
        & Mamba-2.8B & 61.7  & 63.0 & 63.6 & 61.2  & 61.3 & 63.0  & 61.2  & 63.6  & 64.3  & 64.5 & 66.1  & 66.6  & 65.5  & 65.1  & 65.4 & 66.2  & 65.4  & 65.3  & 70.6  & 70.8 & 70.4  & 70.0  & 69.6  \\ 
        & Pythia-2.8B & 57.8  & 60.6 & 61.3 & 58.4  & 58.6 & 60.1  & 58.4  & 60.8  & 59.6  & 59.8 & 60.5  & 62.2  & 60.0  & 61.2  & 61.4 & 63.3  & 61.4  & 62.8  & 64.8  & 64.9 & 65.9  & 65.1  & 65.6  \\ 
        & Pythia-6.9B & 58.2  & 62.6 & 63.2 & 60.7  & 60.9 & 62.6  & 60.6  & 63.0  & 62.4  & 62.5 & 63.3  & 65.0  & 62.7  & 65.0  & 65.2 & 66.7  & 65.2  & 65.3  & 71.6  & 71.8 & 72.1  & 71.2  & 71.2  \\ 
        & Pythia-12B & 59.6  & 63.5 & 62.6 & 61.9  & 62.0 & 63.5  & 61.8  & 63.9  & 63.0  & 63.2 & 64.0  & 65.9  & 63.4  & 66.5  & 66.7 & 67.9  & 66.7  & 67.2  & 72.6  & 72.8 & 72.8  & 71.8  & 71.5  \\ 
        & Llama-13B & 62.0  & 65.3 & 64.1 & 63.6  & 63.7 & 65.0  & 63.7  & 65.7  & 63.4  & 63.4 & 59.8  & 60.1  & 57.0  & 66.0  & 66.1 & 67.4  & 66.1  & 66.7  & 84.3  & 84.4 & 87.2  & 87.0  & 87.4  \\ 
        & Llama-30B & 61.9  & 67.5 & 67.1 & 66.1  & 66.2 & 67.5  & 66.2  & 67.9  & 68.9  & 68.8 & 65.5  & 65.4  & 63.2  & 68.4  & 68.5 & 69.6  & 68.6  & 68.9  & 84.3  & 84.3 & 87.7  & 87.0  & 87.4  \\ 
        & OPT-66B & 61.1  & 63.9 & 64.1 & 62.3  & 62.4 & 64.2  & 62.2  & 64.4  & 66.9  & 67.1 & 68.2  & 69.0  & 68.0  & 66.5  & 66.7 & 68.1  & 66.7  & 66.8  & 69.8  & 69.9 & 70.1  & 69.5  & 69.0  \\ 
        & GPT-NeoX-20B & 66.3  & 68.1 & 67.1 & 66.6  & 66.7 & 67.6  & 66.5  & 67.6  & 66.0  & 66.1 & 66.8  & 67.1  & 65.4  & 72.2  & 72.3 & 70.8  & 72.2  & 68.6  & 76.5  & 76.6 & 76.4  & 76.0  & 75.0  \\
        \cdashline{2-25}
        & Average & 60.6  & 63.9 & 63.7 & 62.1  & 62.2 & \textbf{63.7}  & 62.1  & \textbf{64.2}  & 63.9  & 64.0 & 63.9  & \textbf{64.8}  & 62.8  & 65.8  & 66.0 & \textbf{67.0}  & \textbf{66.0}  & \textbf{66.0}  & 73.5  & 73.7 & \textbf{74.5}  & \textbf{73.9}  & \textbf{73.9}  \\     
        \midrule
    \multirow{10}{*}{128} 
        & Mamba-1.4B & 58.5  & 65.6 & 64.8 & 63.3  & 63.3 & 63.6  & 62.9  & 64.2  & 62.0  & 62.1 & 64.1  & 66.7  & 65.4  & 66.8  & 66.9 & 65.5  & 66.9  & 64.8  & 67.7  & 67.8 & 70.2  & 69.8  & 70.5  \\ 
        & Mamba-2.8B & 62.4  & 68.5 & 67.7 & 66.3  & 66.2 & 66.7  & 65.9  & 66.9  & 66.9  & 67.1 & 69.4  & 71.7  & 70.2  & 70.3  & 70.5 & 69.3  & 70.4  & 68.0  & 71.9  & 72.0 & 73.4  & 71.1  & 72.3  \\ 
        & Pythia-2.8B & 59.5  & 65.0 & 65.2 & 62.8  & 62.9 & 63.1  & 62.6  & 63.3  & 59.6  & 59.8 & 61.4  & 63.5  & 62.2  & 66.9  & 66.9 & 64.3  & 66.6  & 63.2  & 66.3  & 66.4 & 66.8  & 66.4  & 66.9  \\ 
        & Pythia-6.9B & 60.5  & 67.6 & 67.5 & 65.1  & 65.1 & 65.6  & 64.8  & 66.1  & 63.3  & 63.5 & 65.1  & 67.5  & 65.7  & 69.5  & 69.5 & 67.8  & 69.2  & 66.7  & 69.8  & 69.8 & 72.4  & 71.2  & 72.7  \\ 
        & Pythia-12B & 61.4  & 67.8 & 67.1 & 65.8  & 65.8 & 66.2  & 65.6  & 66.7  & 63.9  & 64.0 & 65.1  & 67.2  & 66.0  & 70.7  & 70.7 & 70.0  & 70.5  & 68.9  & 71.8  & 72.0 & 73.5  & 71.9  & 73.5  \\ 
        & Llama-13B & 60.6  & 69.7 & 68.3 & 67.8  & 67.7 & 68.7  & 67.8  & 69.1  & 62.6  & 62.6 & 64.9  & 62.2  & 64.1  & 71.5  & 71.6 & 71.2  & 72.1  & 70.9  & 83.8  & 83.9 & 88.4  & 89.1  & 89.1  \\ 
        & Llama-30B & 59.0  & 71.8 & 72.2 & 70.3  & 70.3 & 71.0  & 70.3  & 71.2  & 71.9  & 72.0 & 70.4  & 66.0  & 68.0  & 73.7  & 73.7 & 72.5  & 73.8  & 71.6  & 82.7  & 82.8 & 85.9  & 87.4  & 87.0  \\ 
        & OPT-66B & 58.9  & 67.3 & 67.7 & 65.5  & 65.5 & 66.7  & 65.6  & 67.6  & 66.9  & 67.1 & 68.6  & 70.3  & 69.0  & 70.6  & 70.7 & 70.6  & 71.3  & 70.3  & 71.1  & 71.3 & 72.9  & 71.9  & 72.4  \\ 
        & GPT-NeoX-20B & 68.0  & 72.3 & 71.6 & 70.7  & 70.8 & 70.7  & 70.4  & 71.0  & 68.3  & 68.6 & 69.5  & 69.3  & 69.6  & 75.6  & 75.8 & 74.5  & 75.7  & 72.6  & 75.4  & 75.5 & 75.7  & 73.9  & 75.1  \\
        \cdashline{2-25}
        & Average & 61.0  & 68.4 & 68.0 & 66.4  & 66.4 & \textbf{66.9}  & 66.2  & \textbf{67.4}  & 65.0  & 65.2 & \textbf{66.5}  & \textbf{67.2}  & \textbf{66.7}  & 70.6  & 70.7 & 69.5  & \textbf{70.7}  & 68.5  & 73.4  & 73.5 & \textbf{75.5}  & \textbf{74.7}  & \textbf{75.5} \\
        \bottomrule
        \end{tabular}
    }
\end{table*}

\begin{table*}[htbp]
    \centering
    \vspace{-5pt}
    \caption{AUC-ROC on WikiMIA benchmark under paraphrased setting. w/ LPDR utilizes linear weights for reweighting, w/ EPDR utilizes exponential weights for reweighting, w/ PPDR utilizes polynomial weights for reweighting. {$^\dag$Neighbor results are from \citet{mia_mink_plus}.}}
\label{tab:wikimia para. auc}
\renewcommand{\arraystretch}{1.3}
\resizebox{0.95\textwidth}{!}{%
    \begin{tabular}{c| c |c c c |c c c c c| c c c c c| c c c c c |c c c c c }
      
    \toprule
        Length & Models & Lowercase & Zlib & {$^\dag$Neighbor} & Loss & \small{w/ LDPR (CAMIA slope)} & w/ LPDR & w/ EPDR & w/ PPDR & Ref & \small{w/ LDPR (CAMIA slope)} & w/ LPDR &  w/ EPDR & w/ PPDR & Min-$k$\% &  \small{w/ LDPR (CAMIA slope)} & w/ LPDR &  w/ EPDR &  w/ PPDR &  Min-$k$\% ++ & \small{w/ LDPR (CAMIA slope)} & w/ LPDR & w/ EPDR & w/ PPDR \\   \midrule
    \multirow{10}{*}{32} 
        & Mamba-1.4B & 60.6  & 62.3 & 63.6 & 61.3  & 61.5 & 61.8  & 61.3  & 61.5  & 62.3  & 62.2 & 62.3  & 62.7  & 62.4  & 62.9  & 62.9 & 63.1  & 62.9  & 63.4  & 65.7  & 65.9 & 66.3  & 65.3  & 66.2  \\ 
        & Mamba-2.8B & 63.5  & 64.8 & 66.3 & 64.5  & 64.5 & 64.8  & 64.3  & 64.6  & 66.6  & 66.2 & 66.5  & 66.7  & 66.2  & 65.3  & 65.2 & 65.4  & 65.2  & 64.9  & 67.3  & 67.5 & 67.5  & 66.6  & 67.2  \\ 
        & Pythia-2.8B & 60.3  & 62.3 & 64.5 & 61.6  & 61.6 & 61.8  & 61.4  & 61.5  & 61.2  & 61.1 & 61.2  & 62.3  & 60.9  & 60.9  & 60.8 & 61.1  & 60.8  & 60.6  & 61.3  & 61.5 & 61.7  & 61.0  & 61.4  \\ 
        & Pythia-6.9B & 61.7  & 64.2 & 65.5 & 64.1  & 64.1 & 64.2  & 63.9  & 64.1  & 63.5  & 63.3 & 63.5  & 64.4  & 63.1  & 65.1  & 65.0 & 65.1  & 64.9  & 64.4  & 67.6  & 67.7 & 67.7  & 66.6  & 67.7  \\ 
        & Pythia-12B & 64.4  & 65.9 & 66.8 & 65.6  & 65.6 & 65.7  & 65.4  & 65.7  & 64.9  & 64.7 & 64.9  & 66.0  & 64.5  & 67.2  & 67.0 & 67.2  & 66.9  & 66.2  & 69.4  & 69.6 & 69.4  & 68.1  & 69.1  \\ 
        & Llama-13B & 63.2  & 68.3 & 65.0 & 68.0  & 68.0 & 68.2  & 68.0  & 68.2  & 56.2  & 56.0 & 56.1  & 54.9  & 55.9  & 66.2  & 66.2 & 66.2  & 66.3  & 65.6  & 82.7  & 82.8 & 84.1  & 84.3  & 83.3  \\ 
        & Llama-30B & 61.3  & 70.4 & 66.3 & 70.2  & 70.2 & 70.3  & 70.1  & 70.2  & 62.4  & 62.5 & 62.4  & 61.3  & 62.2  & 68.5  & 68.5 & 68.4  & 68.3  & 67.7  & 81.2  & 81.2 & 82.4  & 82.6  & 81.6  \\ 
        & OPT-66B & 62.3  & 65.3 & 66.7 & 65.3  & 65.1 & 65.0  & 65.1  & 65.3  & 68.0  & 68.0 & 68.0  & 68.9  & 68.1  & 65.8  & 65.6 & 65.8  & 65.5  & 64.9  & 67.0  & 67.0 & 67.1  & 66.0  & 66.9  \\ 
        & GPT-NeoX-20B & 66.9  & 68.5 & 68.3 & 68.3  & 68.4 & 68.4  & 68.5  & 66.7  & 66.6  & 66.1 & 66.0  & 66.1  & 69.6  & 69.4  & 69.5 & 69.3  & 68.0  & 69.7  & 69.5  & 69.7 & 68.4  & 69.2  \\ 
        \cdashline{2-25}
        & Average & 62.7  & 65.8  & 65.9 & 65.5  & 65.4 & \textbf{65.6}  & 65.3  & 65.5  & 63.5  & 63.3 & 63.5  & \textbf{63.7}  & 63.3  & 65.7  & 65.6 & \textbf{65.8}  & 65.6  & 65.1  & 70.2  & 70.3 & \textbf{70.6}  & 69.9  & \textbf{70.3}  \\ 
        \midrule
    \multirow{10}{*}{64} 
        & Mamba-1.4B & 57.0  & 59.1 & 60.6 & 56.4  & 56.7 & 59.5  & 56.4  & 60.5  & 59.6  & 59.7 & 60.8  & 61.1  & 60.2  & 58.0  & 58.3 & 61.8  & 58.4  & 61.6  & 62.2  & 62.4 & 65.5  & 63.5  & 65.8  \\ 
        & Mamba-2.8B & 62.0  & 61.9 & 63.7 & 59.8  & 60.1 & 62.9  & 59.7  & 63.6  & 64.5  & 64.7 & 66.3  & 66.6  & 65.2  & 62.4  & 62.7 & 65.1  & 62.7  & 64.0  & 64.9  & 65.1 & 67.9  & 66.0  & 67.5  \\ 
        & Pythia-2.8B & 56.1  & 59.0 & 59.6 & 56.5  & 56.8 & 59.3  & 56.5  & 60.3  & 59.2  & 59.3 & 60.8  & 62.0  & 60.1  & 56.7  & 57.0 & 61.6  & 57.0  & 61.3  & 57.7  & 57.9 & 62.0  & 59.8  & 62.1  \\ 
        & Pythia-6.9B & 57.7  & 61.6 & 63.1 & 59.3  & 59.6 & 62.4  & 59.3  & 62.9  & 62.9  & 63.1 & 64.0  & 65.6  & 63.1  & 61.1  & 61.4 & 65.1  & 61.5  & 63.7  & 64.2  & 64.5 & 68.3  & 65.6  & 68.0  \\ 
        & Pythia-12B & 59.2  & 62.1 & 62.8 & 60.0  & 60.2 & 63.0  & 60.0  & 63.6  & 63.2  & 63.3 & 64.5  & 66.1  & 63.7  & 62.5  & 62.8 & 66.1  & 62.7  & 64.8  & 65.1  & 65.3 & 68.2  & 65.5  & 67.5  \\ 
        & Llama-13B & 61.0  & 65.3 & 64.7 & 63.1  & 63.3 & 66.2  & 63.4  & 66.9  & 60.9  & 60.8 & 57.9  & 57.7  & 55.2  & 63.5  & 63.7 & 67.2  & 64.1  & 66.2  & 78.8  & 78.9 & 84.3  & 83.5  & 85.0  \\ 
        & Llama-30B & 59.8  & 67.4 & 66.7 & 65.5  & 65.7 & 68.4  & 65.7  & 69.0  & 65.3  & 65.2 & 62.8  & 62.9  & 60.8  & 64.9  & 65.1 & 67.9  & 65.3  & 66.7  & 74.7  & 74.9 & 81.7  & 80.5  & 82.4  \\ 
        & OPT-66B & 60.0  & 62.2 & 64.6 & 60.3  & 60.5 & 63.1  & 60.3  & 63.5  & 67.8  & 68.0 & 69.1  & 69.8  & 68.9  & 62.5  & 62.8 & 66.0  & 62.8  & 64.9  & 63.3  & 63.5 & 66.6  & 64.3  & 66.4  \\ 
        & GPT-NeoX-20B & 65.6  & 66.5 & 67.4 & 64.4  & 64.7 & 67.2  & 64.4  & 67.2  & 66.0  & 66.1 & 67.2  & 66.9  & 65.4  & 66.1  & 66.4 & 68.7  & 66.4  & 66.5  & 66.2  & 66.4 & 68.2  & 66.2  & 67.5  \\ 
        \cdashline{2-25}
        & Average & 59.8  & 62.8 & 63.7 & 60.6  & 60.8 & \textbf{63.5}  & 60.6  & \textbf{64.2}  & 63.3  & 63.4 & \textbf{63.7}  & \textbf{64.3}  & 62.5  & 61.9  & 62.3 & \textbf{65.5}  & \textbf{62.3}  & \textbf{64.4}  & 66.3  & 66.6 & \textbf{70.3}  & \textbf{68.3}  & \textbf{70.3}  \\       
        \midrule
    \multirow{10}{*}{128} 
        & Mamba-1.4B & 57.7  & 65.3 & 62.6 & 62.7  & 62.8 & 64.1  & 62.7  & 64.6  & 61.1  & 61.2 & 64.6  & 67.9  & 66.3  & 64.4  & 64.5 & 65.8  & 65.1  & 66.1  & 63.3  & 63.5 & 68.2  & 69.5  & 69.8  \\ 
        & Mamba-2.8B & 61.2  & 68.4 & 64.6 & 65.7  & 65.8 & 67.3  & 65.7  & 67.9  & 66.6  & 66.8 & 69.6  & 72.1  & 70.5  & 68.0  & 68.2 & 69.9  & 69.1  & 68.7  & 68.9  & 69.0 & 71.5  & 70.7  & 71.6  \\ 
        & Pythia-2.8B & 59.6  & 65.0 & 61.9 & 62.3  & 62.3 & 64.0  & 62.5  & 64.1  & 59.5  & 59.8 & 62.6  & 64.4  & 63.8  & 64.7  & 64.9 & 63.5  & 64.9  & 62.7  & 62.7  & 62.8 & 65.8  & 66.0  & 66.8  \\ 
        & Pythia-6.9B & 59.9  & 67.4 & 64.3 & 64.7  & 64.8 & 66.6  & 64.7  & 67.3  & 62.9  & 63.3 & 65.9  & 68.5  & 67.2  & 67.0  & 67.2 & 68.9  & 67.8  & 67.8  & 65.9  & 66.1 & 72.2  & 72.0  & 73.3  \\ 
        & Pythia-12B & 60.4  & 67.9 & 64.3 & 65.4  & 65.5 & 67.0  & 65.4  & 67.6  & 63.9  & 64.1 & 66.2  & 68.5  & 66.9  & 68.5  & 68.7 & 69.1  & 69.3  & 68.0  & 67.7  & 67.8 & 72.2  & 71.9  & 73.0  \\ 
        & Llama-13B & 56.3  & 69.6 & 64.0 & 67.2  & 67.2 & 69.1  & 67.6  & 69.9  & 59.7  & 59.7 & 61.7  & 58.8  & 60.6  & 68.6  & 68.8 & 71.0  & 70.3  & 70.7  & 76.2  & 76.4 & 84.3  & 87.2  & 86.1  \\ 
        & Llama-30B & 55.3  & 71.5 & 67.2 & 69.3  & 69.3 & 71.2  & 69.7  & 72.0  & 69.8  & 69.9 & 68.6  & 64.4  & 66.0  & 70.3  & 70.4 & 72.5  & 71.6  & 71.9  & 73.4  & 73.5 & 80.5  & 83.9  & 82.9  \\ 
        & OPT-66B & 57.6  & 66.9 & 63.4 & 64.5  & 64.5 & 66.9  & 64.8  & 67.9  & 67.0  & 67.2 & 69.5  & 70.9  & 70.0  & 67.2  & 67.4 & 69.9  & 68.4  & 69.5  & 67.0  & 67.2 & 69.5  & 70.0  & 70.4  \\ 
        & GPT-NeoX-20B & 67.6  & 72.0 & 69.6 & 69.7  & 69.8 & 71.4  & 69.6  & 71.9  & 68.4  & 68.5 & 70.2  & 70.0  & 70.6  & 73.0  & 73.1 & 75.2  & 73.8  & 73.5  & 70.6  & 70.7 & 72.6  & 72.3  & 72.9  \\
        \cdashline{2-25}
        & Average & 59.5  & 68.2 & 64.7 & 65.7  & 65.8 & \textbf{67.5}  & \textbf{65.9}  & \textbf{68.1}  & 64.3  & 64.5 & \textbf{66.5}  & \textbf{67.3}  & \textbf{66.9}  & 68.0  & 68.1 & \textbf{69.5}  & \textbf{68.9}  & \textbf{68.8}  & 68.4  & 68.5 & \textbf{73.0}  & \textbf{73.7}  & \textbf{74.1} \\ 
        \bottomrule
    \end{tabular}
    }
\end{table*}

\begin{table*}[htbp]
    \centering
    \vspace{-5pt}
    \caption{TPR on WikiMIA benchmark under original setting. w/ LPDR utilizes linear weights for reweighting, w/ EPDR utilizes exponential weights for reweighting, w/ PPDR utilizes polynomial weights for reweighting. {$^\dag$Neighbor results are from \citet{mia_mink_plus}.}}
\label{tab:wikimia ori. tpr}
\renewcommand{\arraystretch}{1.3}
\resizebox{0.95\textwidth}{!}{%
    \begin{tabular}{c| c |c c c |c c c c c| c c c c c| c c c c c |c c c c c }
      
    \toprule
        Length & Models & Lowercase & Zlib & {$^\dag$Neighbor} & Loss & \small{w/ LDPR (CAMIA slope)} & w/ LPDR & w/ EPDR & w/ PPDR & Ref & \small{w/ LDPR (CAMIA slope)} & w/ LPDR &  w/ EPDR & w/ PPDR & Min-$k$\% & \small{w/ LDPR (CAMIA slope)} &  w/ LPDR &  w/ EPDR &  w/ PPDR &  Min-$k$\% ++ & \small{w/ LDPR (CAMIA slope)} & w/ LPDR & w/ EPDR & w/ PPDR \\   \midrule
    \multirow{10}{*}{32} 
        & Mamba-1.4B & 11.1  & 15.5 & 11.9 & 14.2  & 14.5  & 15.2  & 14.0  & 14.0  & 7.8  & 8.3  & 7.8  & 8.8  & 7.2  & 14.2  & 14.2  & 14.7  & 15.0  & 13.4  & 11.4  & 12.7  & 11.6  & 14.0  & 11.6  \\ 
        & Mamba-2.8B & 16.8  & 16.3 & 16.0 & 14.7  & 14.2  & 17.6  & 15.8  & 15.2  & 9.8  & 9.3  & 10.1  & 10.9  & 10.1  & 17.3  & 16.0  & 16.3  & 15.0  & 16.5  & 11.4  & 11.6  & 11.4  & 14.0  & 11.4  \\ 
        & Pythia-2.8B & 11.1  & 15.8 & 15.0 & 14.7  & 15.5  & 17.6  & 15.0  & 15.5  & 6.2  & 8.0  & 6.2  & 11.4  & 5.4  & 16.5  & 17.3  & 17.3  & 16.5  & 17.6  & 10.6  & 10.6  & 10.9  & 14.2  & 10.9  \\ 
        & Pythia-6.9B & 10.6  & 16.3 & 16.5 & 14.2  & 14.5  & 15.2  & 14.5  & 13.4  & 6.7  & 6.2  & 6.5  & 12.1  & 5.7  & 17.8  & 17.8  & 18.1  & 18.1  & 18.1  & 14.5  & 14.5  & 15.2  & 17.3  & 15.2  \\ 
        & Pythia-12B & 16.3  & 17.1 & 19.4 & 17.1  & 17.1  & 17.8  & 17.6  & 15.5  & 9.0  & 9.3  & 8.8  & 11.1  & 9.8  & 23.0  & 22.5  & 22.7  & 23.3  & 23.8  & 16.5  & 16.0  & 17.3  & 19.9  & 15.8  \\ 
        & Llama-13B & 9.6  & 11.6 & 11.6 & 14.0  & 14.2  & 14.2  & 14.0  & 14.7  & 4.7  & 5.4  & 4.9  & 5.2  & 4.9  & 18.9  & 19.6  & 19.9  & 20.2  & 21.2  & 33.1  & 33.1  & 40.1  & 43.4  & 38.2  \\ 
        & Llama-30B & 11.4  & 14.5 & 9.3 & 18.3  & 18.3  & 17.8  & 18.3  & 18.9  & 10.1  & 10.1  & 10.9  & 7.2  & 9.3  & 22.0  & 22.7  & 22.7  & 23.0  & 21.2  & 31.8  & 31.3  & 35.7  & 37.0  & 29.5  \\ 
        & OPT-66B & 11.4  & 16.5 & 21.7 & 14.2  & 15.0  & 15.2  & 15.2  & 16.0  & 11.1  & 9.6  & 11.1  & 10.6  & 10.1  & 21.7  & 20.9  & 21.7  & 20.9  & 20.2  & 11.9  & 12.7  & 14.0  & 15.2  & 14.7  \\ 
        & GPT-NeoX-20B & 16.8  & 20.4 & 22.2 & 20.4  & 20.2  & 22.7  & 20.2  & 21.2  & 15.5  & 17.1  & 16.5  & 17.6  & 15.2  & 28.9  & 29.5  & 30.0  & 28.2  & 26.6  & 19.1  & 19.1  & 19.9  & 20.4  & 21.2  \\ 
        \cdashline{2-25}
        & Average & 12.8  & 16.0 & 16.0 & 15.8  & 15.9  & \textbf{17.1}  & \textbf{16.0}  & \textbf{16.0}  & 9.0  & 9.2  & \textbf{9.2}  & \textbf{10.5}  & 8.6  & 20.0  & 20.1  & \textbf{20.4}  & 20.0  & 19.8  & 17.8  & 17.9  & \textbf{19.6}  & \textbf{21.7}  & \textbf{18.7}  \\   
        \midrule
    \multirow{10}{*}{64} 
        & Mamba-1.4B & 8.8  & 14.1 & 8.8 & 9.5  & 9.5  & 13.0  & 11.3  & 17.3  & 4.6  & 4.6  & 6.7  & 7.0  & 4.2  & 15.8  & 15.8  & 15.5  & 17.3  & 17.3  & 13.7  & 13.4  & 12.7  & 12.3  & 9.9  \\ 
        & Mamba-2.8B & 16.5  & 14.8 & 10.6 & 10.2  & 10.6  & 15.8  & 12.7  & 16.2  & 9.2  & 9.2  & 9.9  & 10.9  & 9.9  & 19.0  & 18.3  & 19.7  & 19.0  & 21.5  & 18.7  & 19.0  & 14.4  & 16.5  & 12.7  \\ 
        & Pythia-2.8B & 10.2  & 14.4 & 10.2 & 10.2  & 10.6  & 14.1  & 10.9  & 16.2  & 10.6  & 9.5  & 5.6  & 10.9  & 3.5  & 18.3  & 18.0  & 22.2  & 16.9  & 21.8  & 13.4  & 13.4  & 14.1  & 14.4  & 12.3  \\ 
        & Pythia-6.9B & 11.6  & 16.2 & 10.9 & 13.4  & 13.4  & 13.0  & 12.3  & 15.5  & 12.0  & 11.3  & 6.3  & 12.0  & 4.9  & 19.0  & 18.0  & 18.7  & 17.3  & 16.9  & 16.2  & 16.5  & 20.1  & 19.7  & 18.7  \\ 
        & Pythia-12B & 12.3  & 11.3 & 11.3 & 9.2  & 9.2  & 13.7  & 8.8  & 16.2  & 13.0  & 12.3  & 7.4  & 10.9  & 8.1  & 21.5  & 21.5  & 19.0  & 21.1  & 23.2  & 16.9  & 17.6  & 22.2  & 23.2  & 19.4  \\ 
        & Llama-13B & 11.6  & 12.7 & 10.2 & 11.3  & 11.3  & 16.5  & 10.6  & 14.4  & 4.2  & 3.5  & 4.9  & 6.0  & 6.7  & 17.3  & 17.3  & 21.1  & 15.8  & 21.5  & 31.3  & 31.3  & 47.9  & 52.5  & 44.7  \\ 
        & Llama-30B & 9.9  & 15.5 & 9.9 & 13.7  & 13.7  & 16.9  & 14.1  & 18.7  & 10.6  & 9.9  & 8.8  & 7.0  & 8.1  & 16.5  & 16.5  & 20.8  & 18.7  & 18.7  & 33.8  & 33.8  & 43.7  & 46.1  & 45.1  \\ 
        & OPT-66B & 10.9  & 13.7 & 12.0 & 13.4  & 13.4  & 13.7  & 13.4  & 16.9  & 13.4  & 13.0  & 9.5  & 12.3  & 7.4  & 21.8  & 21.5  & 22.2  & 22.9  & 18.3  & 19.4  & 19.4  & 16.5  & 18.7  & 18.3  \\ 
        & GPT-NeoX-20B & 16.2  & 19.4 & 13.0 & 12.3  & 12.7  & 15.1  & 13.0  & 21.5  & 14.8  & 15.5  & 13.7  & 16.2  & 12.7  & 19.0  & 19.7  & 26.4  & 19.7  & 18.7  & 22.5  & 22.2  & 20.8  & 21.5  & 21.1  \\
        \cdashline{2-25}
        & Average & 12.0  & 14.7 & 10.8 & 11.5  & 11.6  & \textbf{14.7}  & \textbf{11.9}  & \textbf{17.0}  & 10.3  & 9.9  & 8.1  & \textbf{10.4}  & 7.3  & 18.7  & 18.5  & \textbf{20.6}  & 18.7  & \textbf{19.8}  & 20.7  & 20.7  & \textbf{23.6}  & \textbf{25.0}  & \textbf{22.5}  \\ 
        \midrule
    \multirow{10}{*}{128} 
        & Mamba-1.4B & 13.0  & 19.4 & 15.8 & 11.5  & 11.5  & 18.0  & 13.0  & 11.5  & 10.1  & 10.1  & 9.4  & 15.8  & 12.2  & 9.4  & 9.4  & 19.4  & 11.5  & 21.6  & 10.1  & 10.1  & 13.7  & 12.2  & 11.5  \\ 
        & Mamba-2.8B & 13.7  & 23.7 & 15.1 & 19.4  & 20.1  & 18.0  & 21.6  & 16.5  & 10.1  & 9.4  & 15.8  & 14.4  & 16.5  & 20.1  & 20.1  & 33.8  & 20.1  & 23.0  & 19.4  & 20.1  & 26.6  & 18.7  & 18.0  \\ 
        & Pythia-2.8B & 10.8  & 18.7 & 8.6 & 9.4  & 9.4  & 13.7  & 11.5  & 14.4  & 10.1  & 10.1  & 10.1  & 15.8  & 7.9  & 13.7  & 13.7  & 18.0  & 18.0  & 15.8  & 14.4  & 14.4  & 17.3  & 13.7  & 16.5  \\ 
        & Pythia-6.9B & 13.0  & 20.9 & 10.8 & 14.4  & 15.1  & 15.1  & 15.8  & 15.8  & 13.7  & 13.7  & 15.1  & 17.3  & 13.7  & 18.0  & 18.0  & 28.8  & 21.6  & 21.6  & 20.1  & 20.1  & 23.7  & 18.0  & 20.1  \\ 
        & Pythia-12B & 13.0  & 23.7 & 10.1 & 18.0  & 18.0  & 15.1  & 18.0  & 13.0  & 12.2  & 13.7  & 11.5  & 12.2  & 11.5  & 20.1  & 20.1  & 25.9  & 22.3  & 23.0  & 18.0  & 18.0  & 26.6  & 25.9  & 25.9  \\ 
        & Llama-13B & 15.8  & 18.7 & 12.9 & 21.6  & 21.6  & 19.4  & 20.9  & 15.8  & 10.8  & 10.1  & 5.8  & 5.0  & 5.8  & 20.1  & 20.9  & 28.8  & 22.3  & 28.1  & 38.1  & 38.1  & 56.8  & 57.6  & 54.7  \\ 
        & Llama-30B & 10.1  & 18.0 & 15.1 & 23.7  & 24.5  & 18.0  & 21.6  & 18.7  & 10.8  & 10.8  & 12.2  & 7.9  & 13.7  & 22.3  & 22.3  & 26.6  & 23.0  & 28.1  & 22.3  & 22.3  & 36.0  & 57.6  & 46.8  \\ 
        & OPT-66B & 15.1  & 21.6 & 12.9 & 20.9  & 20.9  & 16.5  & 22.3  & 20.1  & 15.8  & 15.8  & 20.1  & 15.8  & 13.0  & 21.6  & 22.3  & 28.1  & 23.7  & 29.5  & 16.5  & 16.5  & 18.0  & 19.4  & 18.0  \\ 
        & GPT-NeoX-20B & 13.0  & 22.3 & 15.8 & 18.0  & 18.0  & 15.1  & 16.5  & 15.8  & 16.5  & 16.5  & 17.3  & 15.8  & 17.3  & 21.6  & 21.6  & 30.2  & 22.3  & 28.8  & 24.5  & 25.2  & 21.6  & 26.6  & 18.7  \\ 
        \cdashline{2-25}
        & Average & 13.0  & 20.8 & 13.0 & 17.4  & 17.7  & 16.5  & \textbf{17.9}  & 15.7  & 12.2  & 12.2  & \textbf{13.0}  & \textbf{13.3}  & \textbf{12.4}  & 18.5  & 18.7  & \textbf{26.6}  & \textbf{20.5}  & \textbf{24.4}  & 20.4  & 20.5  & \textbf{26.7}  & \textbf{27.7}  & \textbf{25.6} \\     
        \bottomrule
    \end{tabular}
    }
\end{table*}

\begin{table*}[htbp]
    \centering
    \vspace{-5pt}
    \caption{TPR on WikiMIA benchmark under paraphrased setting. w/ LPDR utilizes linear weights for reweighting, w/ EPDR utilizes exponential weights for reweighting, w/ PPDR utilizes polynomial weights for reweighting. {$^\dag$Neighbor results are from \citet{mia_mink_plus}.}}
\label{tab:wikimia para. tpr}
\renewcommand{\arraystretch}{1.3}
\resizebox{0.95\textwidth}{!}{%
    \begin{tabular}{c |c |c c c |c c c c c| c c c c c| c c c c c |c c c c c }
      
    \toprule
        Length & Models & Lowercase & Zlib & {$^\dag$Neighbor} & Loss & \small{w/ LDPR (CAMIA slope)} & w/ LPDR & w/ EPDR & w/ PPDR & Ref & \small{w/ LDPR (CAMIA slope)} & w/ LPDR &  w/ EPDR & w/ PPDR & Min-$k$\% & \small{w/ LDPR (CAMIA slope)} &  w/ LPDR &  w/ EPDR &  w/ PPDR &  Min-$k$\% ++ & \small{w/ LDPR (CAMIA slope)} & w/ LPDR & w/ EPDR & w/ PPDR \\   \midrule
    \multirow{10}{*}{32} 
        & Mamba-1.4B & 13.2  & 13.2 & 7.2 & 14.2  & 15.0  & 16.0  & 12.9  & 15.5  & 5.9  & 6.5  & 5.9  & 9.6  & 6.2  & 11.9  & 13.7  & 14.0  & 12.4  & 15.0  & 7.8  & 7.5  & 11.4  & 11.6  & 9.8  \\ 
        & Mamba-2.8B & 15.0  & 12.7 & 9.3 & 16.5  & 16.3  & 17.8  & 16.3  & 17.3  & 10.1  & 9.3  & 9.3  & 11.9  & 10.6  & 19.9  & 20.4  & 20.2  & 20.4  & 15.2  & 11.1  & 11.6  & 12.9  & 11.1  & 9.8  \\ 
        & Pythia-2.8B & 11.6  & 14.5 & 8.5 & 14.2  & 15.0  & 15.5  & 15.0  & 14.2  & 7.2  & 6.7  & 7.2  & 12.9  & 7.5  & 16.3  & 16.8  & 16.8  & 15.2  & 15.8  & 10.9  & 10.6  & 10.6  & 10.9  & 11.4  \\ 
        & Pythia-6.9B & 11.9  & 12.7 & 9.6 & 15.0  & 15.0  & 14.0  & 14.2  & 14.5  & 6.2  & 5.9  & 5.7  & 12.9  & 5.7  & 21.7  & 23.0  & 22.0  & 22.7  & 18.1  & 14.5  & 15.0  & 16.5  & 15.5  & 15.0  \\ 
        & Pythia-12B & 16.5  & 15.5 & 9.8 & 17.3  & 17.6  & 16.8  & 17.8  & 18.9  & 8.0  & 9.3  & 8.3  & 10.3  & 8.3  & 19.9  & 20.7  & 20.4  & 20.7  & 18.3  & 15.5  & 15.0  & 13.4  & 14.0  & 13.4  \\ 
        & Llama-13B & 9.6  & 15.0 & 8.5 & 16.3  & 16.3  & 16.5  & 16.0  & 16.8  & 5.4  & 5.9  & 5.9  & 4.4  & 6.2  & 14.2  & 14.2  & 15.2  & 14.5  & 15.5  & 33.9  & 34.4  & 37.7  & 33.9  & 35.9  \\ 
        & Llama-30B & 13.2  & 15.2 & 9.3 & 14.7  & 15.2  & 17.6  & 15.5  & 16.8  & 8.3  & 8.0  & 7.5  & 5.9  & 7.8  & 18.1  & 18.3  & 17.8  & 17.8  & 19.4  & 25.8  & 26.6  & 29.2  & 31.8  & 24.8  \\ 
        & OPT-66B & 12.9  & 16.8 & 12.1 & 15.2  & 15.2  & 17.6  & 15.5  & 15.5  & 9.3  & 9.3  & 9.6  & 10.9  & 8.5  & 16.8  & 16.0  & 18.3  & 18.1  & 14.7  & 16.3  & 16.0  & 16.3  & 17.1  & 13.4  \\ 
        & GPT-NeoX-20B & 14.5  & 19.6 & 15.2 & 17.6  & 19.4  & 19.4  & 18.9  & 19.9  & 15.2  & 14.2  & 15.0  & 14.2  & 14.0  & 19.1  & 18.6  & 19.4  & 19.4  & 19.6  & 13.4  & 13.2  & 13.2  & 11.9  & 11.1  \\
        \cdashline{2-25}
        & Average & 13.1  & 15.0 & 9.9 & 15.7  & \textbf{16.8}  & \textbf{15.8}  & \textbf{16.6}  & 8.4  & 8.3  & \textbf{10.3}  & 8.3  & 17.5  & \textbf{18.2}  & \textbf{17.9}  & 16.9  & 16.6  & \textbf{17.9}  & \textbf{17.5}  & 16.1  \\     
        \midrule
    \multirow{10}{*}{64} 
        & Mamba-1.4B & 9.5  & 15.1 & 9.5 & 8.1  & 8.1  & 12.0  & 9.9  & 12.7  & 8.1  & 7.4  & 9.2  & 10.9  & 6.3  & 7.7  & 8.1  & 14.8  & 9.2  & 16.2  & 7.0  & 7.0  & 8.5  & 6.3  & 10.2  \\ 
        & Mamba-2.8B & 14.8  & 14.8 & 18.3 & 12.3  & 12.3  & 13.0  & 13.0  & 15.8  & 11.3  & 10.9  & 12.3  & 14.8  & 9.5  & 11.6  & 11.6  & 19.7  & 12.3  & 18.3  & 9.5  & 9.5  & 14.1  & 12.7  & 16.5  \\ 
        & Pythia-2.8B & 11.3  & 16.5 & 11.3 & 9.5  & 11.3  & 13.4  & 12.3  & 15.1  & 13.0  & 12.7  & 9.9  & 12.7  & 7.4  & 9.9  & 9.9  & 18.3  & 11.3  & 18.7  & 8.1  & 8.1  & 9.9  & 7.4  & 13.4  \\ 
        & Pythia-6.9B & 11.3  & 15.8 & 12.7 & 10.6  & 10.9  & 13.0  & 12.7  & 14.4  & 16.2  & 14.4  & 7.7  & 15.5  & 5.6  & 12.7  & 12.0  & 18.3  & 13.0  & 17.3  & 10.2  & 10.6  & 12.7  & 10.9  & 16.5  \\ 
        & Pythia-12B & 13.4  & 16.2 & 10.6 & 11.6  & 11.3  & 14.1  & 11.3  & 12.0  & 14.4  & 13.0  & 10.2  & 14.4  & 9.2  & 12.7  & 13.7  & 20.8  & 16.9  & 17.6  & 10.9  & 11.6  & 18.7  & 15.8  & 16.9  \\ 
        & Llama-13B & 13.7  & 13.4 & 14.4 & 12.0  & 12.0  & 14.1  & 12.3  & 14.8  & 4.6  & 4.6  & 6.0  & 6.0  & 5.6  & 13.4  & 13.0  & 21.1  & 12.3  & 14.4  & 23.2  & 23.6  & 31.3  & 31.3  & 29.2  \\ 
        & Llama-30B & 7.7  & 16.9 & 11.6 & 13.4  & 13.4  & 13.0  & 13.7  & 15.8  & 8.5  & 7.7  & 7.4  & 6.7  & 6.7  & 14.8  & 14.1  & 17.3  & 10.6  & 16.5  & 20.8  & 21.5  & 35.6  & 30.6  & 35.6  \\ 
        & OPT-66B & 12.3  & 14.8 & 13.7 & 13.4  & 13.4  & 14.8  & 15.1  & 14.1  & 13.7  & 13.4  & 8.5  & 12.7  & 6.7  & 13.0  & 13.7  & 22.2  & 14.8  & 19.7  & 11.6  & 12.3  & 15.5  & 11.3  & 13.4  \\ 
        & GPT-NeoX-20B & 13.4  & 19.0 & 18.3 & 14.8  & 15.8  & 19.4  & 15.5  & 18.7  & 15.5  & 15.5  & 14.4  & 15.5  & 12.0  & 15.1  & 15.1  & 20.8  & 14.4  & 17.6  & 10.6  & 10.6  & 13.4  & 15.1  & 13.0  \\ 
        \cdashline{2-25}
        & Average & 11.9  & 15.8 & 13.4 & 11.7  & 12.1  & \textbf{14.1}  & \textbf{12.9}  & \textbf{14.8}  & 11.7  & 11.1  & 9.5  & \textbf{12.1}  & 7.7  & 12.3  & 12.4  & \textbf{19.2}  & \textbf{12.8}  & \textbf{17.4}  & 12.4  & 12.8  & \textbf{17.7}  & \textbf{15.7}  & \textbf{18.3}  \\     
        \midrule
    \multirow{10}{*}{128} 
        & Mamba-1.4B & 11.5  & 17.3 & 13.7 & 13.7  & 13.7  & 18.7  & 10.1  & 14.4  & 11.5  & 11.5  & 12.2  & 16.5  & 12.2  & 5.0  & 5.8  & 20.9  & 10.8  & 23.7  & 6.5  & 6.5  & 9.4  & 13.7  & 8.6  \\ 
        & Mamba-2.8B & 15.1  & 20.1 & 17.3 & 16.5  & 15.1  & 22.3  & 16.5  & 15.8  & 10.8  & 10.1  & 15.1  & 18.0  & 20.9  & 12.2  & 11.5  & 30.2  & 18.7  & 20.1  & 13.0  & 12.2  & 16.5  & 15.1  & 15.1  \\ 
        & Pythia-2.8B & 8.6  & 16.5 & 12.2 & 14.4  & 14.4  & 11.5  & 7.2  & 13.0  & 7.2  & 7.2  & 10.1  & 11.5  & 5.8  & 12.2  & 12.2  & 21.6  & 15.8  & 18.0  & 7.9  & 8.6  & 10.1  & 13.0  & 8.6  \\ 
        & Pythia-6.9B & 11.5  & 20.9 & 17.3 & 16.5  & 16.5  & 15.1  & 18.0  & 13.0  & 8.6  & 9.4  & 10.8  & 18.0  & 8.6  & 16.5  & 16.5  & 23.0  & 16.5  & 21.6  & 18.0  & 18.0  & 18.7  & 14.4  & 17.3  \\ 
        & Pythia-12B & 12.2  & 19.4 & 10.1 & 19.4  & 19.4  & 14.4  & 16.5  & 12.2  & 8.6  & 9.4  & 13.0  & 10.1  & 13.0  & 18.7  & 18.0  & 33.1  & 10.8  & 29.5  & 9.4  & 10.1  & 15.1  & 20.1  & 18.7  \\ 
        & Llama-13B & 15.8  & 21.6 & 13.7 & 18.0  & 18.0  & 23.7  & 21.6  & 25.9  & 4.3  & 4.3  & 7.9  & 6.5  & 7.9  & 15.1  & 14.4  & 30.9  & 20.9  & 33.1  & 35.3  & 35.3  & 46.8  & 45.3  & 49.6  \\ 
        & Llama-30B & 11.5  & 19.4 & 14.4 & 18.0  & 17.3  & 25.9  & 15.1  & 23.0  & 17.3  & 17.3  & 13.0  & 11.5  & 10.1  & 17.3  & 17.3  & 33.8  & 16.5  & 37.4  & 21.6  & 22.3  & 34.5  & 49.6  & 46.8  \\ 
        & OPT-66B & 11.5  & 18.7 & 12.9 & 18.7  & 18.7  & 15.8  & 18.0  & 15.8  & 16.5  & 17.3  & 17.3  & 13.0  & 12.2  & 15.1  & 15.8  & 31.7  & 21.6  & 23.7  & 13.7  & 13.7  & 15.8  & 22.3  & 18.0  \\ 
        & GPT-NeoX-20B & 15.1  & 22.3 & 18.7 & 20.1  & 20.9  & 25.9  & 20.1  & 24.5  & 18.0  & 18.0  & 12.2  & 11.5  & 13.0  & 23.0  & 23.0  & 32.4  & 23.7  & 31.7  & 20.9  & 20.9  & 9.4  & 20.9  & 12.2  \\ 
        \cdashline{2-25}
        & Average & 12.6  & 19.6 & 14.5 & 17.3  & 17.1  & \textbf{19.3}  & 15.9  & \textbf{17.5}  & 11.4  & 11.6  & \textbf{12.4}  & \textbf{12.9}  & \textbf{11.5}  & 15.0  & 14.9  & \textbf{28.6}  & \textbf{17.3}  & \textbf{26.5}  & 16.2  & 16.4  & \textbf{19.6}  & \textbf{23.8}  & \textbf{21.7} \\  
        \bottomrule
    \end{tabular}
    }
\end{table*}

\begin{table*}[tp]
\centering
\caption{AUROC results on WikiMIA benchmark, compare with FSD}
\vspace{-5pt}
\label{tab:acl_fsd_auc}
\resizebox{0.8\textwidth}{!}{%
\begin{tabular}{llcccccccccccc}\\
\toprule
\multirow{2}{*}{\textbf{Dataset}} & \multirow{2}{*}{\textbf{Method}} 
& \multicolumn{2}{c}{\textbf{GPT-J-6B}} 
& \multicolumn{2}{c}{\textbf{OPT-6.7B}} 
& \multicolumn{2}{c}{\textbf{Pythia-6.9B}} 
& \multicolumn{2}{c}{\textbf{LLaMA-7B}} 
& \multicolumn{2}{c}{\textbf{NeoX-20B}}  
& \multicolumn{2}{c}{\textbf{Average}}  \\
\cmidrule(lr){3-4} \cmidrule(lr){5-6} \cmidrule(lr){7-8} \cmidrule(lr){9-10} \cmidrule(lr){11-12} \cmidrule(lr){13-14}
& & \textit{Base} & \textit{w/ FSD} 
  & \textit{Base} & \textit{w/ FSD} 
  & \textit{Base} & \textit{w/ FSD} 
  & \textit{Base} & \textit{w/ FSD} 
  & \textit{Base} & \textit{w/ FSD}  
  & \textit{Base} & \textit{w/ FSD}  \\
\midrule

\multirow{8}{*}{WikiMIA} 
        &Min-$k$\% & 67.9  & 93.4  & 62.5  & 89.4  & 66.7  & 91.9  & 65.4  & 88.4  & 73.4  & 90.1  & 67.2  & 90.6  \\ 
        &w/ LPDR & 68.1  & 94.0  & 62.9  & 89.6  & 67.0  & 92.6  & 66.7  & 88.8  & 73.5  & 91.0  & \textbf{67.6}  & \textbf{91.2}  \\ 
        &w/ EPDR & 67.1  & 94.1  & 62.1  & 90.0  & 66.0  & 92.8  & 65.8  & 89.1  & 72.1  & 91.3  & 66.6  & \textbf{91.4}  \\ 
        &w/ PPDR & 67.9  & 93.8  & 62.9  & 89.5  & 66.9  & 92.2  & 66.6  & 88.9  & 73.3  & 90.7  & \textbf{67.5}  & \textbf{91.0}  \\ 
        &Min-$k$\%++  & 67.6  & 81.6  & 63.0  & 83.3  & 68.1  & 81.5  & 79.9  & 91.1  & 74.4  & 77.8  & 70.6  & 83.1  \\ 
        &w/ LPDR & 68.3  & 84.8  & 63.6  & 84.5  & 69.4  & 85.0  & 80.8  & 92.1  & 74.8  & 81.7  & \textbf{71.4}  & \textbf{85.6}  \\ 
        &w/ EPDR & 68.2  & 83.7  & 63.3  & 84.5  & 69.2  & 83.6  & 80.8  & 91.8  & 74.8  & 81.2  & \textbf{71.2}  & \textbf{85.0}  \\ 
        &w/ PPDR & 68.3  & 85.5  & 63.7  & 84.6  & 69.5  & 85.8  & 80.7  & 92.1  & 74.6  & 82.4  & \textbf{71.4}  & \textbf{86.1}  \\ 
\bottomrule

\end{tabular}
}

\end{table*}
\begin{table*}[tp]
\centering
\caption{TPR results on WikiMIA benchmark, compare with FSD}
\vspace{-5pt}
\label{tab:acl_fsd_tpr}
\resizebox{0.8\linewidth}{!}{%
\begin{tabular}{llcccccccccccc}\\
\toprule
\multirow{2}{*}{\textbf{Dataset}} & \multirow{2}{*}{\textbf{Method}} 
& \multicolumn{2}{c}{\textbf{GPT-J-6B}} 
& \multicolumn{2}{c}{\textbf{OPT-6.7B}} 
& \multicolumn{2}{c}{\textbf{Pythia-6.9B}} 
& \multicolumn{2}{c}{\textbf{LLaMA-7B}} 
& \multicolumn{2}{c}{\textbf{NeoX-20B}}  
& \multicolumn{2}{c}{\textbf{Average}}  \\
\cmidrule(lr){3-4} \cmidrule(lr){5-6} \cmidrule(lr){7-8} \cmidrule(lr){9-10} \cmidrule(lr){11-12} \cmidrule(lr){13-14}
& & \textit{Base} & \textit{w/ FSD} 
  & \textit{Base} & \textit{w/ FSD} 
  & \textit{Base} & \textit{w/ FSD} 
  & \textit{Base} & \textit{w/ FSD} 
  & \textit{Base} & \textit{w/ FSD}  
  & \textit{Base} & \textit{w/ FSD}  \\
\midrule

\multirow{8}{*}{WikiMIA} 
        &Min-$k$\% & 17.2  & 55.6  & 13.9  & 40.6  & 17.2  & 57.6  & 14.7  & 32.6  & 24.7  & 36.2  & 17.5  & 44.5  \\ 
        &w/ LPDR & 18.0  & 61.9  & 16.4  & 41.9  & 18.2  & 61.1  & 16.4  & 33.7  & 27.7  & 52.9  & \textbf{19.3}  & \textbf{50.3}  \\ 
        &w/ EPDR & 16.5  & 60.1  & 13.5  & 45.4  & 16.5  & 59.8  & 17.0  & 33.6  & 21.5  & 50.8  & 17.0  & \textbf{49.9}  \\ 
        &w/ PPDR & 17.9  & 59.9  & 15.4  & 39.9  & 19.9  & 61.9  & 18.5  & 31.6  & 27.0  & 50.3  & \textbf{19.7}  & \textbf{48.7}  \\ 
        &Min-$k$\%++  & 15.9  & 24.2  & 11.7  & 29.0  & 19.0  & 25.0  & 20.4  & 39.1  & 17.5  & 10.5  & 16.9  & 25.6  \\ 
        &w/ LPDR & 18.9  & 32.2  & 12.9  & 35.7  & 19.0  & 37.4  & 22.2  & 48.4  & 19.7  & 15.5  & \textbf{18.5}  & \textbf{33.9}  \\ 
        &w/ EPDR & 19.0  & 39.6  & 12.2  & 32.4  & 19.4  & 34.9  & 20.2  & 46.4  & 19.0  & 18.7  & \textbf{18.0}  & \textbf{34.4}  \\ 
        &w/ PPDR & 18.9  & 34.9  & 12.9  & 36.7  & 17.0  & 40.4  & 24.0  & 51.6  & 19.7  & 17.4  & \textbf{18.5}  & \textbf{36.2}  \\ 
\bottomrule

\end{tabular}
}

\end{table*}

\begin{figure*}[t]
    \centering
    \includegraphics[width=0.98\textwidth]{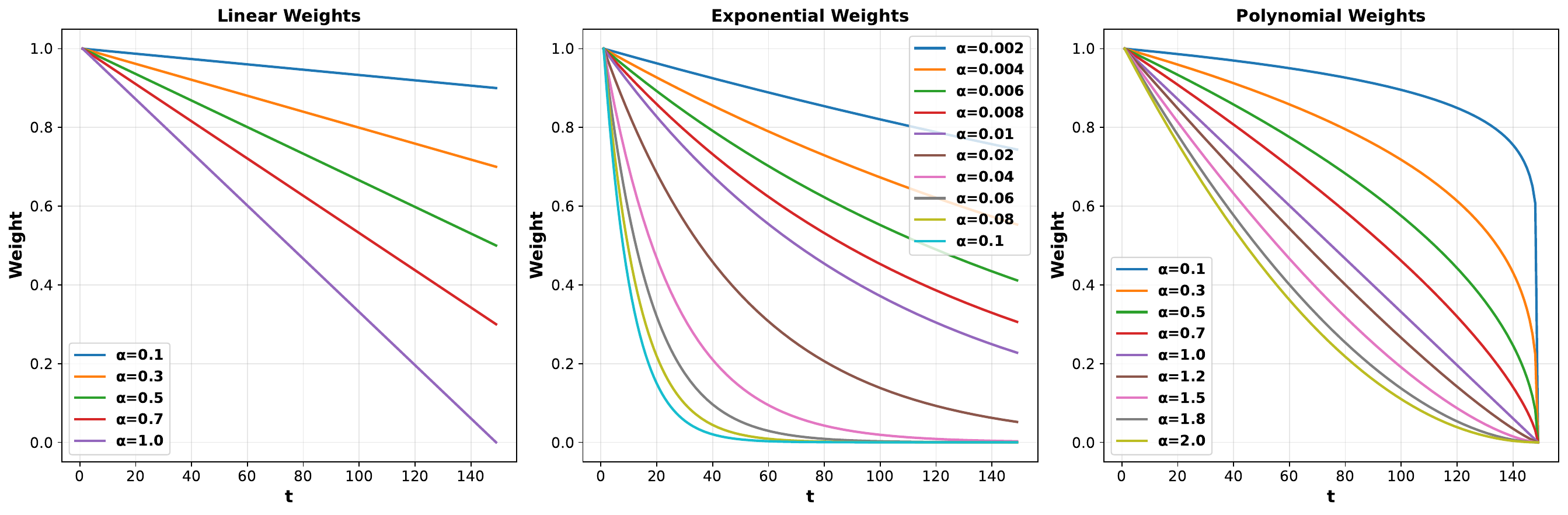}
    \caption{Visualization of different positional weight functions (Linear, Exponential, Polynomial) with various hyperparameters. The x-axis represents the token position in a sequence of length 150, while the y-axis shows the corresponding weight applied to that position.}
    \label{fig:weight_visualization}
    \vspace{-5pt}
\end{figure*}

\subsection{Static Analysis}
\label{apd:static_analysis}

To rigorously assess the statistical significance of our method's improvements, we performed a non-parametric bootstrap analysis based on the model prediction scores and ground-truth labels. Our procedure is as follows:

\begin{itemize}
    \item \textbf{Bootstrap Resampling:} For each method, we generated $N=1000$ bootstrap replicates. Each replicate was created by sampling indices from the original test set with replacement, forming a new dataset of the same size. A random seed was fixed to ensure reproducibility. If a replicate happened to contain samples from only one class, it was discarded for that specific calculation.
    \item \textbf{Metrics and Confidence Intervals:} For each of the $N$ bootstrap replicates, we calculated the AUROC and TPR@0.5\%FPR. After generating all replicates, we computed the mean and standard deviation of these metrics. The 95\% confidence intervals were then derived empirically from the 2.5th and 97.5th percentiles of the resulting distribution of 1000 metric values.
    \item \textbf{Paired Significance Testing:} To evaluate if the improvements of our PDR-enhanced methods over their respective baselines are statistically significant, we conducted a paired bootstrap test. This approach is crucial for reducing variance caused by the sampling process. For each of the $N=1000$ replicates, we used the \textit{exact same set of resampled indices} to evaluate both the baseline method and the PDR-enhanced method. We then calculated the performance difference for that replicate: $\delta = \text{AUROC}_{\text{PDR}} - \text{AUROC}_{\text{Baseline}}$. A one-sided P-value was subsequently derived by calculating the proportion of replicates where this difference was not positive ($\delta \le 0$). This P-value directly tests the null hypothesis $H_0: \text{AUROC}_{\text{PDR}} \le \text{AUROC}_{\text{Baseline}}$. A P-value less than 0.05 is considered to indicate a statistically significant improvement.
\end{itemize}

The analysis reveals a clear trend: the effectiveness of PDR is strongly correlated with sequence length. For short sequences (e.g., 32 tokens), the performance gains are marginal and not always statistically significant. However, as the sequence length increases to 64 and 128, PDR's improvements become both substantial and statistically significant (p-value $<$ 0.05) for most baselines. For instance, on Pythia-6.9B with length 128, PDR boosts the AUROC of Min-k\%++ from 65.9 to 72.2 (p-value $<$ 0.001). This demonstrates that the positional prior becomes a more robust and discriminative signal as more context becomes available in longer sequences, confirming that the observed gains are not due to random noise.

\section{FSD Results}
\label{apd:fsd results}
This section presents the detailed results of combining our method with the Finetuning-based Score Difference (FSD) framework on the WikiMIA dataset. Please refer to Tab.~\ref{tab:acl_fsd_auc} and Tab.~\ref{tab:acl_fsd_tpr} for the full experimental results.

\section{Mimir Results}
\label{apd:mimir results}
This section presents the complete AUROC and TPR results on the challenging MIMIR benchmark, which is known for its minimal distribution shift and increased difficulty compared to WikiMIA. The overall average results for five Pythia models are summarized in Table~\ref{tab:mimir_full}. The results in Tab.~\ref{tab:mimir_auroc} and Tab.~\ref{tab:mimir_tpr} comprehensively demonstrate the performance of our proposed PDR methods (LPDR, EPDR, PPDR) and all baselines on MIMIR, across different models and sub datasets.

\begin{table}[htbp]

\centering
\caption{\small{AUROC scores of various MIA methods over five Pythia models on the Mimir dataset. Pub, Wiki, and Hack denote Pubmed Central, Wikipedia (en) and  HackerNews. Avg* scores are computed by excluding Arxiv and HackerNews.{ $^\dag$Neighbor results are from \citet{mia_mink_plus}, induces significant extra computational cost than others ($25\times$ in this case), for which reason we don't run on the 12B model.}}}
\label{tab:mimir_full}
\resizebox{1.0\linewidth}{!}{%
    \begin{tabular}{lccccc>{\columncolor{gray!15}}c>{\columncolor{gray!15}}c c}
        \toprule
        \textbf{Method} & \textbf{Wiki} & \textbf{Pile-CC}& \textbf{Pub} & \textbf{Math} & \textbf{GitHub}& \textcolor{black!40}{\textbf{ArXiv}}    & \textcolor{black!40}{\textbf{Hack} }& \textbf{Avg*}\\
        \midrule
        Lowercase & 52.2  & 49.3  & 51.1  & 48.9  & 71.1  & 51.5  & 50.9  & 54.5  \\ 
        Zlib & 52.7  & 50.4  & 50.4  & 48.1  & 71.9  & 51.4  & 50.8  & 54.7  \\ 
        {$^\dag$Neighbor} & 51.9 & 50.1  & 49.2  & 47.4 & 69.3  &  51.5 &  51.5  & 53.6  \\
        Loss & 51.9  & 50.3  & 50.3  & 48.5  & 70.8  & 52.1  & 51.2  & 54.4  \\ 
        \rowcolor{gray!15} \textbf{w/ LPDR}& \textbf{52.8}  & \textbf{50.7}  & 50.3  & \textbf{48.6}  & \textbf{70.9}  & 51.8  & 51.3  & \textbf{54.6}  \\ 
        Min-$k$\% & 51.8  & 50.7  & 50.9  & 49.2  & 70.9  & 52.3  & 52.4  & 54.7  \\ 
        \rowcolor{gray!15} \textbf{w/ LPDR} & \textbf{54.2}  & \textbf{51.2}  & \textbf{51.0}  & \textbf{49.5}  & \textbf{71.0}  & 51.4  & 51.5  & \textbf{55.4}  \\ 
        Min-$k$\%++ & 54.0  & 50.5  & 51.9  & 50.3  & 70.4  & 52.6  & 52.9  & 55.4  \\ 
        \rowcolor{gray!15} \textbf{w/ LPDR} & \textbf{55.5}  & \textbf{50.8}  & \textbf{52.4}  & 50.3  & 70.1  & 52.7  & 52.1  & \textbf{55.8} \\ 
        \bottomrule
    \end{tabular}
    }
\end{table}

\section{Visualization of Weight Decay Functions} 
\label{apd:weight_decay_sec}


In this section, we visualize weight decay functions (Linear, Exponential, or Polynomial), with varying $\alpha$. We use the following ranges:
\begin{itemize}
    \item For \textbf{Linear} decay, the range for the coefficient $\alpha$ is:  $\{0.1, 0.3, 0.5, 0.7, 1.0\}$
    \item For \textbf{Exponential} decay, the range for the coefficient $\alpha$ is: \\ $\{0.002, 0.004, 0.006, 0.008, \\
    0.01, 0.02, 0.04, 0.06, 0.08, 0.1\}$.
    \item For \textbf{Polynomial} decay, the range for the coefficient $\alpha$ is: \\$\{0.1, 0.3, 0.5, 0.7, 1.0, 1.2, 1.5, 1.8, 2.0\}$.
\end{itemize}

Fig.~\ref{fig:weight_visualization} visualizes how the decay function becomes steeper as the hyperparameter ($\alpha$) 
increases. 

\begin{table*}[htbp]
\caption{AUROC results on the challenging MIMIR benchmark.{$^\dag$Neighbor results are from \citet{mia_mink_plus}, induces significant extra computational cost than others ($25\times$ in this case), for which reason we don't run on the 12B model.}}
\label{tab:mimir_auroc}
\begin{center} \scriptsize
\setlength{\tabcolsep}{0.7pt}
\begin{tabularx}{\textwidth}{l *{20}{>{\centering\arraybackslash}X}@{}}
    \toprule
    \multirow{2}{*}{}  & \multicolumn{5}{c}{\textbf{Wikipedia}} & \multicolumn{5}{c}{\textbf{Github}} & \multicolumn{5}{c}{\textbf{Pile CC}} & \multicolumn{5}{c}{\textbf{PubMed Central}} \\
    \cmidrule(lr){2-6}  \cmidrule(lr){7-11} \cmidrule(lr){12-16} \cmidrule(lr){17-21}
    \textbf{Method} & 160M & 1.4B & 2.8B & 6.9B & 12B
    & 160M & 1.4B & 2.8B & 6.9B & 12B
    & 160M & 1.4B & 2.8B & 6.9B & 12B
    & 160M & 1.4B & 2.8B & 6.9B & 12B
    \\
    \midrule
    Lowercase & 50.1  & 51.3  & 51.7  & 53.5  & 54.3  & 67.2  & 70.3  & 71.3  & 72.9  & 73.7  & 47.8  & 48.6  & 49.5  & 50.1  & 50.6  & 49.5  & 50.4  & 51.5  & 51.5  & 52.7  \\ 
    Zlib & 51.1  & 52.0  & 52.4  & 53.5  & 54.3  & 67.5  & 71.0  & 72.3  & 73.9  & 74.9  & 49.6  & 50.1  & 50.3  & 50.8  & 51.1  & 49.9  & 50.0  & 50.1  & 50.6  & 51.2  \\ 
    {$^\dag$Neighbor} & 50.7 & 51.7 & 52.2 & 53.2 & / & 65.3 & 69.4 & 70.5 & 72.1 & / & 49.6 & 50.0 & 50.1 & 50.8 & / & 47.9 & 49.1 & 49.7 & 50.1 & / \\
    \midrule
    Loss & 50.2  & 51.3  & 51.8  & 52.8  & 53.5  & 65.7  & 69.8  & 71.3  & 73.0  & 74.0  & 49.6  & 50.0  & 50.1  & 50.7  & 51.1  & 49.9  & 49.8  & 49.9  & 50.6  & 51.3  \\ 
    w/ LPDR (CAMIA slope) & 50.3  & 51.3  & 51.8  & 52.8  & 53.5  & 65.8  & 69.8  & 71.3  & 73.0  & 74.0  & 49.6  & 50.0  & 50.1  & 50.7  & 51.1  & 49.9  & 49.8  & 49.9  & 50.6  & 51.3  \\ 
    w/ LPDR & 51.2  & 52.0  & 52.6  & 53.7  & 54.4  & 66.0  & 70.0  & 71.5  & 73.0  & 73.9  & 49.9  & 50.4  & 50.5  & 51.0  & 51.4  & 50.1  & 49.9  & 50.0  & 50.6  & 51.1  \\ 
    w/ EPDR & 50.5  & 51.5  & 51.9  & 53.0  & 53.7  & 65.7  & 69.9  & 71.3  & 73.0  & 74.0  & 49.6  & 50.0  & 50.1  & 50.7  & 51.0  & 49.7  & 49.6  & 49.7  & 50.3  & 50.9  \\ 
    w/ PPDR & 50.4  & 51.4  & 51.9  & 52.9  & 53.6  & 66.0  & 70.0  & 71.5  & 73.2  & 74.2  & 49.6  & 50.1  & 50.2  & 50.8  & 51.1  & 49.9  & 49.8  & 49.9  & 50.6  & 51.3  \\ 
    \midrule
    Min-$k$\% & 48.8  & 51.0  & 51.7  & 53.1  & 54.2  & 65.7  & 70.0  & 71.4  & 73.3  & 74.2  & 50.1  & 50.5  & 50.5  & 51.2  & 51.5  & 50.3  & 50.3  & 50.5  & 51.2  & 52.3  \\ 
    w/ LPDR (CAMIA slope) & 48.9  & 51.0  & 51.7  & 53.2  & 54.2  & 65.7  & 70.0  & 71.5  & 73.3  & 74.2  & 50.1  & 50.5  & 50.5  & 51.2  & 51.5  & 50.3  & 50.3  & 50.5  & 51.2  & 52.3  \\ 
    w/ LPDR & 52.7  & 53.3  & 54.1  & 55.0  & 56.0  & 65.9  & 70.2  & 71.7  & 73.2  & 74.1  & 50.2  & 51.0  & 51.1  & 51.6  & 51.9  & 50.7  & 50.7  & 50.6  & 51.1  & 52.0  \\ 
    w/ EPDR & 50.4  & 51.7  & 52.5  & 53.9  & 54.9  & 65.7  & 70.0  & 71.5  & 73.3  & 74.3  & 50.4  & 50.7  & 50.7  & 51.4  & 51.6  & 50.1  & 50.1  & 50.2  & 50.8  & 51.7  \\ 
    w/ PPDR & 49.2  & 51.3  & 52.1  & 53.5  & 54.5  & 66.1  & 70.3  & 71.8  & 73.5  & 74.5  & 50.2  & 50.6  & 50.6  & 51.4  & 51.6  & 50.4  & 50.4  & 50.5  & 51.2  & 52.4  \\ 
    \midrule
    Min-$k$\%++ & 49.2  & 53.1  & 53.8  & 56.1  & 57.9  & 64.7  & 69.6  & 70.9  & 72.8  & 74.2  & 49.7  & 50.0  & 49.8  & 51.2  & 51.8  & 50.2  & 50.8  & 51.5  & 52.8  & 54.0  \\
    w/ LPDR (CAMIA slope) & 49.2  & 53.2  & 53.8  & 56.1  & 57.9  & 64.7  & 69.6  & 70.9  & 72.8  & 74.2  & 49.7  & 50.0  & 49.8  & 51.2  & 51.8  & 50.2  & 50.8  & 51.5  & 52.8  & 54.0  \\  
    w/ LPDR & 51.0  & 54.8  & 55.5  & 57.5  & 59.0  & 64.6  & 69.5  & 70.6  & 72.3  & 73.7  & 49.7  & 50.6  & 49.9  & 51.8  & 51.9  & 50.4  & 51.8  & 52.2  & 53.6  & 54.4  \\ 
    w/ EPDR & 49.9  & 53.7  & 54.4  & 56.7  & 58.5  & 64.7  & 69.7  & 70.8  & 72.7  & 74.1  & 49.9  & 50.3  & 49.9  & 51.5  & 51.8  & 50.0  & 50.9  & 51.5  & 52.8  & 53.7  \\ 
    w/ PPDR & 49.3  & 53.5  & 54.2  & 56.5  & 58.3  & 64.8  & 69.8  & 71.0  & 72.9  & 74.3  & 49.7  & 50.2  & 49.8  & 51.4  & 51.9  & 50.2  & 51.1  & 51.7  & 53.1  & 54.3  \\

        \vspace{-.6em} \\
    
    \toprule
    \multirow{2}{*}{}  & \multicolumn{5}{c}{\textbf{ArXiv}} & \multicolumn{5}{c}{\textbf{DM Mathematics}} & \multicolumn{5}{c}{\textbf{HackerNews}} & \multicolumn{5}{c}{\textbf{Average}}\\
    \cmidrule(lr){2-6}  \cmidrule(lr){7-11} \cmidrule(lr){12-16} \cmidrule(lr){17-21}
    \textbf{Method} & 160M & 1.4B & 2.8B & 6.9B & 12B
    & 160M & 1.4B & 2.8B & 6.9B & 12B
    & 160M & 1.4B & 2.8B & 6.9B & 12B
    & 160M & 1.4B & 2.8B & 6.9B & 12B
    \\
    \midrule

    Lowercase & 50.8  & 50.7  & 51.3  & 51.9  & 52.8  & 48.9  & 49.0  & 49.0  & 49.1  & 48.2  & 49.0  & 50.4  & 51.1  & 51.6  & 52.3  & 52.4  & 53.4  & 54.1  & 54.8  & 55.4  \\ 
    Zlib & 50.1  & 50.9  & 51.3  & 52.2  & 52.7  & 48.1  & 48.2  & 48.0  & 48.1  & 48.1  & 49.7  & 50.3  & 50.8  & 51.2  & 51.7  & 52.7  & 53.7  & 54.1  & 54.9  & 55.4  \\ 
    {$^\dag$Neighbor} & 50.7 & 51.4 & 51.8 & 52.2 & / & 49.0 & 47.0 & 46.8 & 46.6 & / & 50.9 & 51.7 & 51.5 & 51.9 & / & 52.0 & 52.9 & 53.2 & 53.8 & / \\
    \midrule
    Loss & 51.0  & 51.5  & 51.9  & 52.9  & 53.4  & 48.8  & 48.5  & 48.4  & 48.5  & 48.5  & 49.4  & 50.5  & 51.3  & 52.1  & 52.8  & 52.5  & 53.5  & 53.9  & 54.7  & 55.3  \\ 
    w/ LPDR (CAMIA slope) & 51.0  & 51.5  & 51.9  & 52.9  & 53.4  & 48.8  & 48.5  & 48.4  & 48.5  & 48.5  & 49.4  & 50.5  & 51.3  & 52.1  & 52.8  & 52.1  & 53.1  & 53.5  & 54.4  & 54.9  \\ 
    w/ LPDR & 50.4  & 51.1  & 51.5  & 52.7  & 53.2  & 48.7  & 48.6  & 48.4  & 48.5  & 48.6  & 49.7  & 50.6  & 51.3  & 52.2  & 52.6  & \textbf{52.7}  & \textbf{53.7}  & \textbf{54.1}  & \textbf{54.9}  & \textbf{55.4}  \\ 
    w/ EPDR & 50.7  & 51.2  & 51.6  & 52.5  & 53.0  & 48.5  & 48.3  & 48.2  & 48.3  & 48.3  & 49.5  & 50.3  & 51.0  & 51.6  & 52.1  & 52.4  & 53.4  & 53.8  & 54.6  & 55.1  \\ 
    w/ PPDR & 50.9  & 51.4  & 51.9  & 52.9  & 53.4  & 48.8  & 48.6  & 48.4  & 48.5  & 48.5  & 49.4  & 50.5  & 51.3  & 52.1  & 52.8  & \textbf{52.6}  & \textbf{53.6}  & \textbf{54.0}  &\textbf{54.8}  & \textbf{55.4} \\ 
    \midrule
    Min-$k$\% & 50.4  & 51.4  & 52.1  & 53.4  & 54.3  & 49.3  & 49.3  & 49.1  & 49.2  & 49.2  & 50.6  & 51.2  & 52.4  & 53.5  & 54.5  & 52.4  & 53.7  & 54.2  & 55.2  & 56.0  \\ 
    w/ LPDR (CAMIA slope) & 50.4  & 51.4  & 52.1  & 53.4  & 54.3  & 49.3  & 49.3  & 49.1  & 49.2  & 49.2  & 50.6  & 51.2  & 52.4  & 53.5  & 54.5  & 52.2  & 53.4  & 54.0  & 55.0  & 55.7  \\ 
    w/ LPDR & 49.2  & 50.8  & 51.2  & 52.5  & 53.2  & 49.5  & 49.7  & 49.3  & 49.5  & 49.5  & 50.3  & 50.7  & 51.3  & 52.4  & 52.8  & \textbf{53.0}  & \textbf{54.3}  & \textbf{54.6}  & \textbf{55.5}  & \textbf{56.1}  \\ 
    w/ EPDR & 49.9  & 51.0  & 51.5  & 52.8  & 53.5  & 49.3  & 49.4  & 49.1  & 49.2  & 49.2  & 50.5  & 51.0  & 51.9  & 52.9  & 53.3  & \textbf{52.6}  & \textbf{53.8}  & \textbf{54.3}  & 55.2  & 55.9  \\ 
    w/ PPDR & 50.2  & 51.4  & 52.1  & 53.4  & 54.3  & 49.3  & 49.4  & 49.2  & 49.3  & 49.3  & 50.5  & 51.1  & 52.3  & 53.5  & 54.4  & \textbf{52.6}  & \textbf{53.9}  & \textbf{54.4}  & \textbf{55.4}  & \textbf{56.1}  \\ 
    \midrule
    Min-$k$\%++ & 49.3  & 50.9  & 53.0  & 53.6  & 56.2  & 50.1  & 50.2  & 50.2  & 50.5  & 50.4  & 50.7  & 51.3  & 52.6  & 54.1  & 55.8  & 52.2  & 54.1  & 54.9  & 56.2  & 57.4  \\ 
    w/ LPDR (CAMIA slope) & 49.3  & 50.8  & 53.0  & 53.6  & 56.3  & 50.1  & 50.2  & 50.2  & 50.5  & 50.4  & 50.7  & 51.3  & 52.6  & 54.1  & 55.8  & 52.0  & 53.7  & 54.5  & 55.9  & 57.2  \\ 
    w/ LPDR & 50.0  & 51.1  & 52.4  & 53.9  & 56.1  & 50.1  & 50.3  & 50.2  & 50.4  & 50.5  & 50.7  & 50.9  & 51.7  & 53.0  & 54.3  & \textbf{52.6}  & \textbf{54.7}  & \textbf{55.1}  & \textbf{56.6}  & \textbf{57.6}  \\ 
    w/ EPDR & 49.6  & 50.8  & 52.6  & 53.4  & 55.7  & 49.8  & 50.0  & 49.9  & 50.2  & 50.2  & 50.6  & 50.9  & 52.0  & 53.3  & 54.7  & \textbf{52.3}  & \textbf{54.2}  & 54.9  & 56.2  & 57.3  \\ 
    w/ PPDR & 49.5  & 51.0  & 53.1  & 53.8  & 56.3  & 50.2  & 50.2  & 50.2  & 50.6  & 50.5  & 50.5  & 51.2  & 52.5  & 54.0  & 55.6  & \textbf{52.3}  & \textbf{54.3}  & \textbf{55.0}  & \textbf{56.4}  & \textbf{57.6} \\

\bottomrule
\end{tabularx}
\end{center}
\end{table*}
\clearpage
\begin{table*}[htbp]
\caption{TPR results on the challenging MIMIR benchmark.{$^\dag$Neighbor results are from \citet{mia_mink_plus}, induces significant extra computational cost than others ($25\times$ in this case), for which reason we don't run on the 12B model.}}
\label{tab:mimir_tpr}
\begin{center} \scriptsize
\setlength{\tabcolsep}{0.7pt}
\begin{tabularx}{\textwidth}{l *{20}{>{\centering\arraybackslash}X}@{}}
    \toprule
    \multirow{2}{*}{}  & \multicolumn{5}{c}{\textbf{Wikipedia}} & \multicolumn{5}{c}{\textbf{Github}} & \multicolumn{5}{c}{\textbf{Pile CC}} & \multicolumn{5}{c}{\textbf{PubMed Central}} \\
    \cmidrule(lr){2-6}  \cmidrule(lr){7-11} \cmidrule(lr){12-16} \cmidrule(lr){17-21}
    \textbf{Method} & 160M & 1.4B & 2.8B & 6.9B & 12B
    & 160M & 1.4B & 2.8B & 6.9B & 12B
    & 160M & 1.4B & 2.8B & 6.9B & 12B
    & 160M & 1.4B & 2.8B & 6.9B & 12B
    \\
    \midrule
        Lowercase & 4.6 & 4.5 & 4.9 & 5.2 & 5.6 & 24.4 & 32.2 & 34.3 & 38.1 & 38.6 & 3.4 & 5.3 & 5.3 & 6.2 & 6.4 & 3.5 & 5 & 5.3 & 6 & 5.2 \\ 
        Zlib & 4.2 & 5.7 & 5.9 & 6.3 & 6.8 & 25.1 & 32.8 & 36.2 & 40.1 & 40.8 & 4 & 5.1 & 5.4 & 6.2 & 6.6 & 3.8 & 3.6 & 3.5 & 4.3 & 4.4 \\ 
        {$^\dag$Neighbor} & 4.0 & 4.5 & 4.9 & 5.8 & / & 24.7 & 31.6 & 29.8 & 34.1 & / & 3.9 & 3.6 & 4.0 & 5.3 & / & 3.9 & 3.7 & 4.5 & 4.5 & / \\
        \midrule
        Loss & 4.2 & 4.7 & 4.7 & 5.1 & 5 & 22.6 & 32.1 & 33.6 & 38.5 & 40.7 & 3.1 & 5 & 4.8 & 4.9 & 5.1 & 4 & 4.4 & 4.3 & 4.9 & 5 \\ 
        w/ LPDR (CAMIA slope) & 4.5 & 4.7 & 4.7 & 4.9 & 5 & 22.6 & 32.1 & 33.6 & 38.5 & 40.7 & 3.0 & 4.9 & 4.8 & 4.9 & 5.3 & 4.0 & 4.5 & 4.3 & 4.8 & 5.0 \\ 
        w/ LPDR & 4.6 & 4.6 & 5.4 & 5.7 & 6 & 21 & 30.8 & 32.4 & 34.7 & 36.4 & 2.4 & 3.4 & 4 & 4.8 & 5.8 & 4.1 & 4.6 & 4.5 & 4.9 & 5.5 \\ 
        w/ EPDR & 4.5 & 4.8 & 5.1 & 5.5 & 5.7 & 21.5 & 31.6 & 32.6 & 36.6 & 38.7 & 3.2 & 4.5 & 4.6 & 4.7 & 5.2 & 3.7 & 4.6 & 4.6 & 4.8 & 4.9 \\ 
        w/ PPDR & 4.2 & 4.7 & 4.9 & 4.9 & 5.3 & 22.7 & 32.1 & 35 & 38.8 & 39.8 & 3 & 4.3 & 4.7 & 5.3 & 6.1 & 4.1 & 4.2 & 4.1 & 4.4 & 5.2 \\ 
        \midrule
        Min-$k$\% & 4.8 & 5.6 & 5 & 6.1 & 5.8 & 22.6 & 31.5 & 34 & 39 & 40.7 & 3.5 & 4.5 & 4.8 & 5 & 4.8 & 4.7 & 4.6 & 4.5 & 5.1 & 4.9 \\ 
        w/ LPDR (CAMIA slope) & 4.8 & 5.6 & 5.0 & 6.1 & 5.8 & 22.5 & 31.5 & 34.0 & 39.0 & 40.7 & 3.6 & 4.4 & 4.8 & 5.0 & 4.8 & 4.7 & 4.6 & 4.5 & 5.1 & 5.1 \\ 
        w/ LPDR & 5.5 & 6.4 & 6.2 & 6.2 & 7.2 & 21.4 & 30.9 & 32.7 & 36 & 36.7 & 2.3 & 2.9 & 3.1 & 2.4 & 3.6 & 6 & 5.6 & 5.6 & 5.7 & 4.8 \\ 
        w/ EPDR & 5 & 5.9 & 5.9 & 5.8 & 6.8 & 20.4 & 31 & 32.9 & 37.1 & 38.4 & 3.3 & 4.4 & 4.4 & 4.1 & 5.2 & 4 & 4.6 & 4.9 & 5.5 & 5.8 \\ 
        w/ PPDR & 5.3 & 5.3 & 5.4 & 5.6 & 5.6 & 22.9 & 31.9 & 34.6 & 39.1 & 40.2 & 3.8 & 4.5 & 4.2 & 4.7 & 5.2 & 5.3 & 5.2 & 4.7 & 5.2 & 6.4 \\ 
        \midrule
        Min-$k$\%++ & 5.2 & 5.3 & 5.9 & 7 & 7.8 & 25.2 & 33 & 34.2 & 38.2 & 40.1 & 5 & 3.7 & 3.7 & 4.8 & 4.6 & 4.8 & 6.1 & 4.8 & 5.6 & 6.4 \\
        w/ LPDR (CAMIA slope) & 5.2 & 5.3 & 5.9 & 7.0 & 7.8 & 25.2 & 33.0 & 34.2 & 38.1 & 40.1 & 5.0 & 3.7 & 3.6 & 4.8 & 4.6 & 4.8 & 5.9 & 4.8 & 5.6 & 6.3 \\  
        w/ LPDR & 4.6 & 6 & 6.3 & 7.5 & 7.9 & 22.6 & 30.6 & 32.7 & 35 & 39.7 & 4.7 & 2.9 & 3.4 & 4 & 5 & 4.2 & 6.4 & 5.5 & 7.8 & 6.2 \\ 
        w/ EPDR & 5.1 & 5.9 & 6.6 & 7.8 & 8.2 & 24 & 33 & 33.5 & 37.4 & 40.7 & 4.6 & 3.4 & 3.6 & 4.3 & 4.9 & 4.9 & 5.4 & 5.4 & 6.8 & 5.8 \\ 
        w/ PPDR & 5.2 & 5 & 5.5 & 7.7 & 7.5 & 26.2 & 34.2 & 34.9 & 38.8 & 40 & 5.2 & 3.5 & 3.3 & 4.3 & 4.6 & 5.1 & 6.4 & 5.1 & 5.8 & 6.7 \\ 

        \vspace{-.6em} \\
    
    \toprule
    \multirow{2}{*}{}  & \multicolumn{5}{c}{\textbf{ArXiv}} & \multicolumn{5}{c}{\textbf{DM Mathematics}} & \multicolumn{5}{c}{\textbf{HackerNews}} & \multicolumn{5}{c}{\textbf{Average}}\\
    \cmidrule(lr){2-6}  \cmidrule(lr){7-11} \cmidrule(lr){12-16} \cmidrule(lr){17-21}
    \textbf{Method} & 160M & 1.4B & 2.8B & 6.9B & 12B
    & 160M & 1.4B & 2.8B & 6.9B & 12B
    & 160M & 1.4B & 2.8B & 6.9B & 12B
    & 160M & 1.4B & 2.8B & 6.9B & 12B
    \\
    \midrule

        Lowercase & 5.1 & 4.7 & 5.4 & 5.6 & 5.2 & 5.6 & 6.2 & 5.5 & 6.8 & 5.8 & 5.2 & 5.2 & 6.3 & 6.6 & 6.4 & 7.8  & 9.7  & 10.1  & 11.3  & 11.1  \\ 
        Zlib & 2.9 & 4.3 & 4.1 & 4.6 & 4.7 & 4.1 & 5 & 4.6 & 4.3 & 4.3 & 5 & 5.5 & 5.8 & 5.6 & 5.8 & 7.4  & 9.4  & 10.0  & 11.0  & 11.3  \\ 
        {$^\dag$Neighbor} & 4.7 & 4.8 & 4.4 & 4.1 & / & \textbf{5.6} & 4.4 & 4.5 & 4.5 & / & \textbf{6.5} & 5.2 & 5.3 & 5.7 & / & 7.6 & 8.3 & 8.2 & 9.1 & / \\
        \midrule
        Loss & 4 & 4.8 & 4.6 & 5.4 & 5.6 & 3.8 & 4.3 & 4.1 & 4.1 & 4 & 5 & 4.8 & 5.5 & 5.9 & 6.8 & 7.0  & 9.2  & 9.4  & 10.5  & 10.9  \\ 
        w/ LPDR (CAMIA slope) & 4.0 & 4.8 & 4.6 & 5.4 & 5.6 & 3.8 & 4.3 & 4.1 & 4.1 & 4.0 & 4.9 & 4.8 & 5.4 & 5.9 & 6.7 & 6.7 & 8.6 & 8.8 & 9.8 & 10.3 \\ 
        w/ LPDR & 3.4 & 3.8 & 2.8 & 4.1 & 3.8 & 3.6 & 3.7 & 3.7 & 3.5 & 3.6 & 4.8 & 5.6 & 5.5 & 6 & 5.2 & 6.5  & 8.5  & 8.8  & 9.6  & 10.2  \\ 
        w/ EPDR & 3.5 & 3.9 & 4 & 4.3 & 4.6 & 4.8 & 4.5 & 4.7 & 4.6 & 4.5 & 5.8 & 6.1 & 6 & 6 & 6.2 & 6.9  & 9.0  & 9.3  & 10.1  & 10.6  \\ 
        w/ PPDR & 4.4 & 4.2 & 4.8 & 5.5 & 5.3 & 3.9 & 4.2 & 3.8 & 4.1 & 3.8 & 5.5 & 5.6 & 5.4 & 5.9 & 6.1 & \textbf{7.1}  & 9.0  & \textbf{9.6}  & 10.5  & 10.9  \\ 
        \midrule
        Min-$k$\% & 4.4 & 4.3 & 4.5 & 5.4 & 5.3 & 3.9 & 4.1 & 4.6 & 4.3 & 4.6 & 4.2 & 4.6 & 5.7 & 6.3 & 6.1 & 7.3  & 9.1  & 9.6  & 10.8  & 11.0  \\ 
        w/ LPDR (CAMIA slope) & 4.6 & 4.3 & 4.4 & 5.2 & 5.2 & 4.0 & 4.1 & 4.6 & 4.2 & 4.6 & 4.2 & 4.7 & 5.7 & 6.3 & 6.0 & 6.9 & 8.5 & 9.0 & 10.1 & 10.3 \\ 
        w/ LPDR & 4 & 4.1 & 3.7 & 4.6 & 4.6 & 4 & 4.1 & 4.6 & 3.9 & 3.6 & 4.7 & 6.3 & 4.1 & 6 & 5.7 & 7.2  & 9.0  & 9.3  & 9.8  & 10.1  \\ 
        w/ EPDR & 3.5 & 3.5 & 4 & 4.2 & 5 & 4.3 & 4.2 & 4.5 & 4.5 & 4.7 & 4.8 & 5.5 & 5 & 5.5 & 5.4 & 6.8  & 8.9  & 9.4  & 10.2  & 11.0  \\ 
        w/ PPDR & 4.6 & 3.8 & 4.4 & 5 & 6.1 & 3.8 & 3.4 & 3.9 & 4.1 & 3.8 & 4.5 & 4.7 & 5.5 & 6.8 & 5.7 & \textbf{7.6}  & 9.0  & 9.5  & 10.6  & \textbf{11.2}  \\ 
        \midrule
        Min-$k$\%++ & 5.4 & 4.7 & 6.2 & 6.8 & 7 & 4.4 & 4.8 & 5.4 & 4.5 & 5.4 & 4.4 & 3.5 & 4.6 & 5.7 & 5.7 & 8.3  & 9.6  & 10.0  & 11.2  & 11.9  \\ 
        w/ LPDR (CAMIA slope) & 5.5 & 4.7 & 6.2 & 6.6 & 6.8 & 4.4 & 4.8 & 5.4 & 4.5 & 5.4 & 4.4 & 3.3 & 4.7 & 5.7 & 5.9 & 7.8 & 8.7 & 9.3 & 10.3 & 11.0 \\ 
        w/ LPDR & 4.6 & 5.2 & 6.8 & 7.6 & 8.6 & 4.4 & 4.4 & 4.4 & 4.5 & 4.7 & 5.3 & 3.9 & 4.8 & 6.8 & 7.2 & 7.5  & 9.3  & 9.9  & 11.1  & \textbf{12.0}  \\ 
        w/ EPDR & 5.1 & 5 & 7.1 & 7.3 & 6.4 & 4.6 & 4.8 & 4.8 & 4.5 & 5 & 4.3 & 4.7 & 5.4 & 6.2 & 6.4 & 8.1  & 9.6  & \textbf{10.2}  & \textbf{11.4}  & 11.8  \\ 
        w/ PPDR & 5.5 & 4.6 & 6.4 & 7.3 & 6.7 & 4.3 & 4.8 & 5.2 & 4.6 & 5.1 & 4.3 & 3.7 & 5.1 & 6.6 & 5.8 & 
        \textbf{8.6}  & \textbf{9.8}  & \textbf{10.1}  & \textbf{11.4}  & 11.8 \\

\bottomrule

\end{tabularx}
\end{center}
\end{table*}
\clearpage

\section{Truncation Analysis}
\label{apd:truncation_analysis}

\begin{figure}[h]
    \centering
    \includegraphics[width=0.95\linewidth]{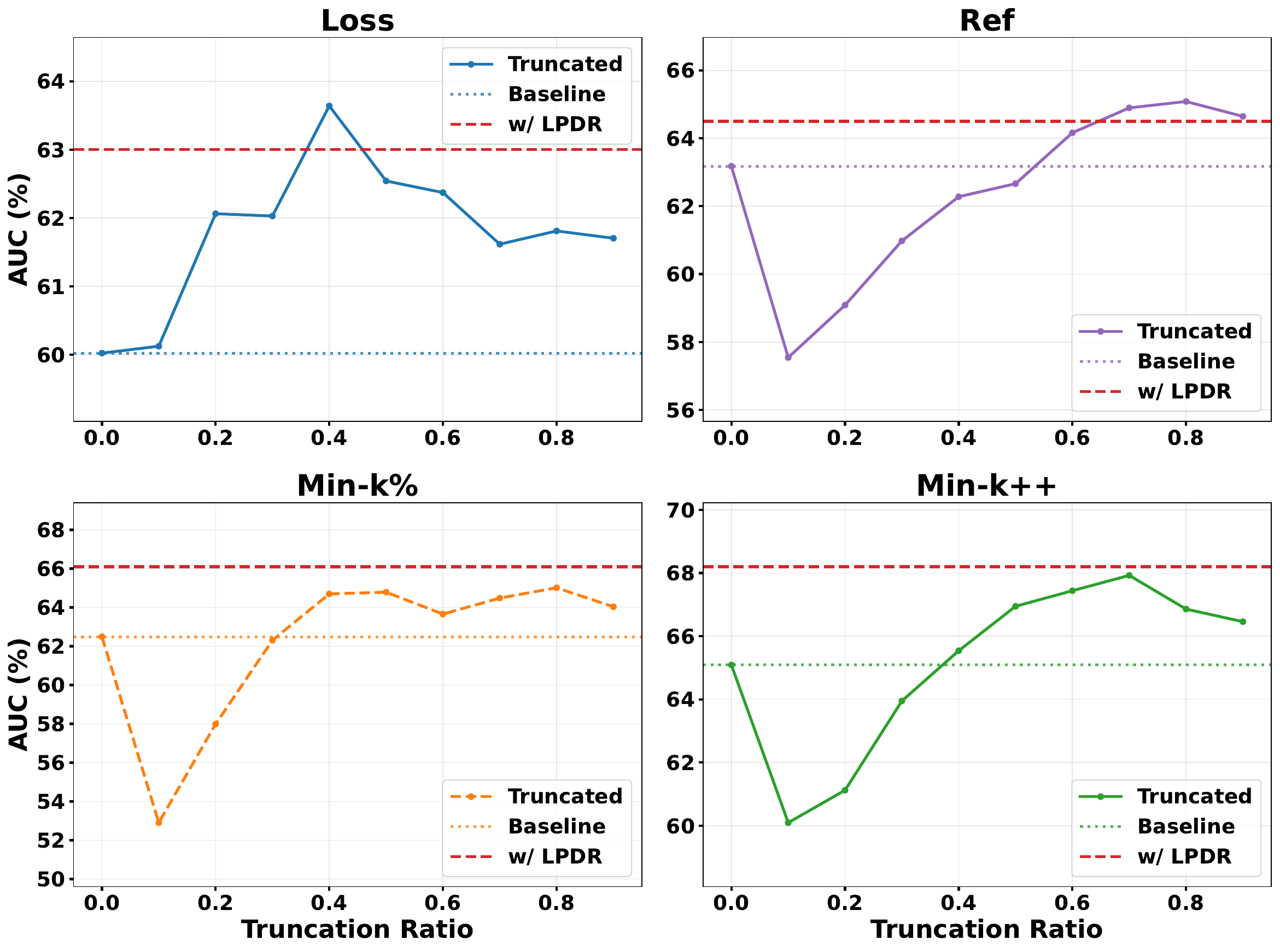}
    \vspace{-8pt}
    \caption{{Performance comparison between the Truncation baseline and our LPDR method ($\alpha=1.0$) on the WikiMIA dataset (length 64, Pythia-12B). The x-axis represents the percentage of the sequence retained for the truncation method. 
    Solid lines indicate results of truncated base methods, whereas dashed lines indicate results with LPDR.}}
    \label{fig:truncation_analysis}
    \vspace{-20pt}
\end{figure}

We compare PDR to a simple \textbf{Truncation} baseline, which discards a suffix of the sequence before scoring. We varied the truncation percentage (the portion of the sequence retained) to find the optimal performance for each base method. As shown in Figure~\ref{fig:truncation_analysis}, the optimal truncation percentage is highly inconsistent across different methods, making it difficult to find a single best setting. In contrast, our LPDR method (with a fixed $\alpha=1.0$, shown as dashed lines) robustly outperforms even the best possible truncation result for all base methods. This demonstrates that PDR's "soft" reweighting is more effective and reliable than the "hard" cutoff of truncation.

\section{Sample Reweighted Analysis}
\label{apd:sample_reweighted_analysis}
\begin{figure}[hb]
    \centering
    \begin{minipage}{0.48\textwidth}
        \centering
        \includegraphics[width=\linewidth]{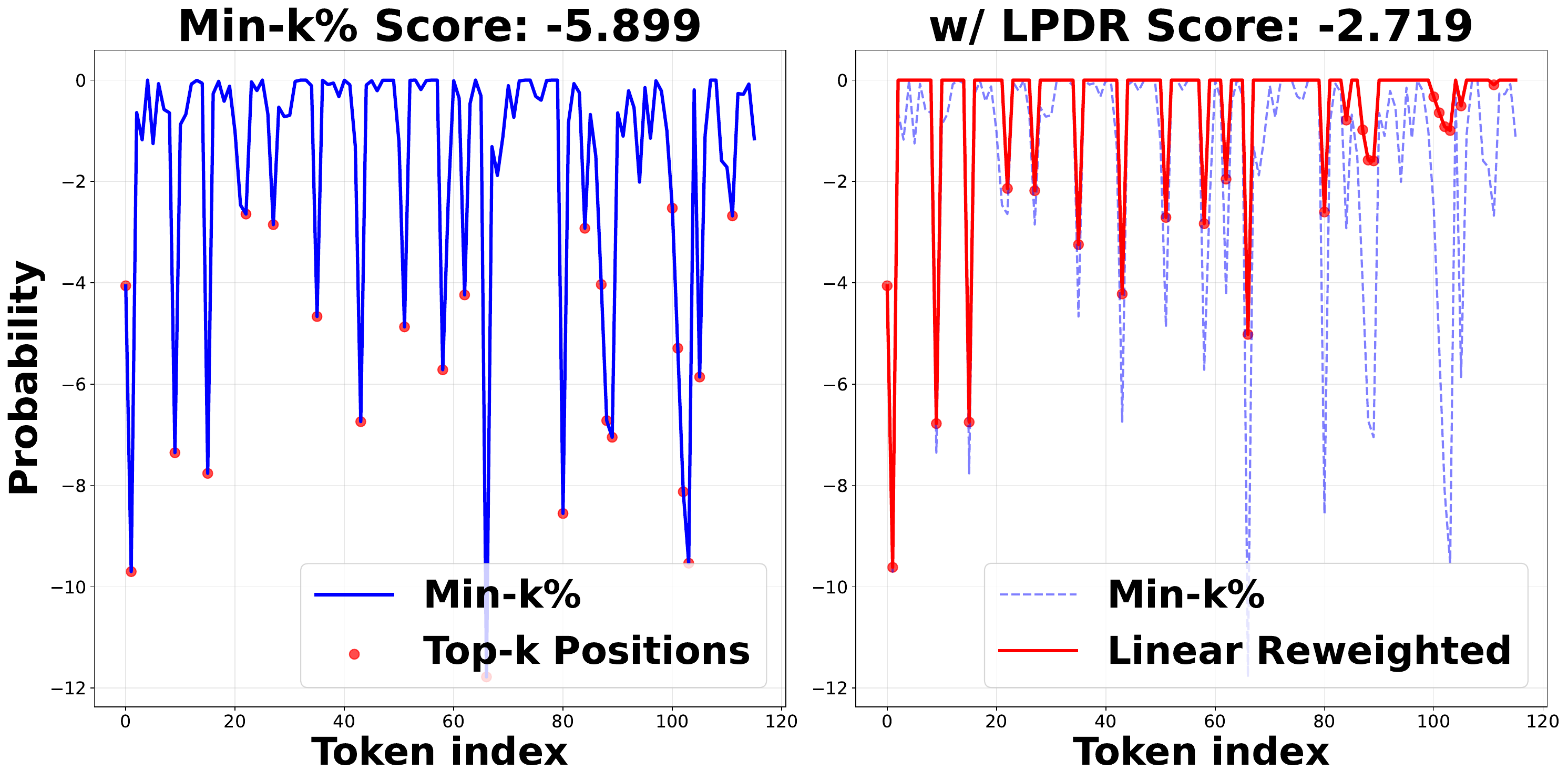}
        \centerline{(a) Member sample}
    \end{minipage}\hfill 
    \begin{minipage}{0.48\textwidth}
        \centering
        \includegraphics[width=\linewidth]{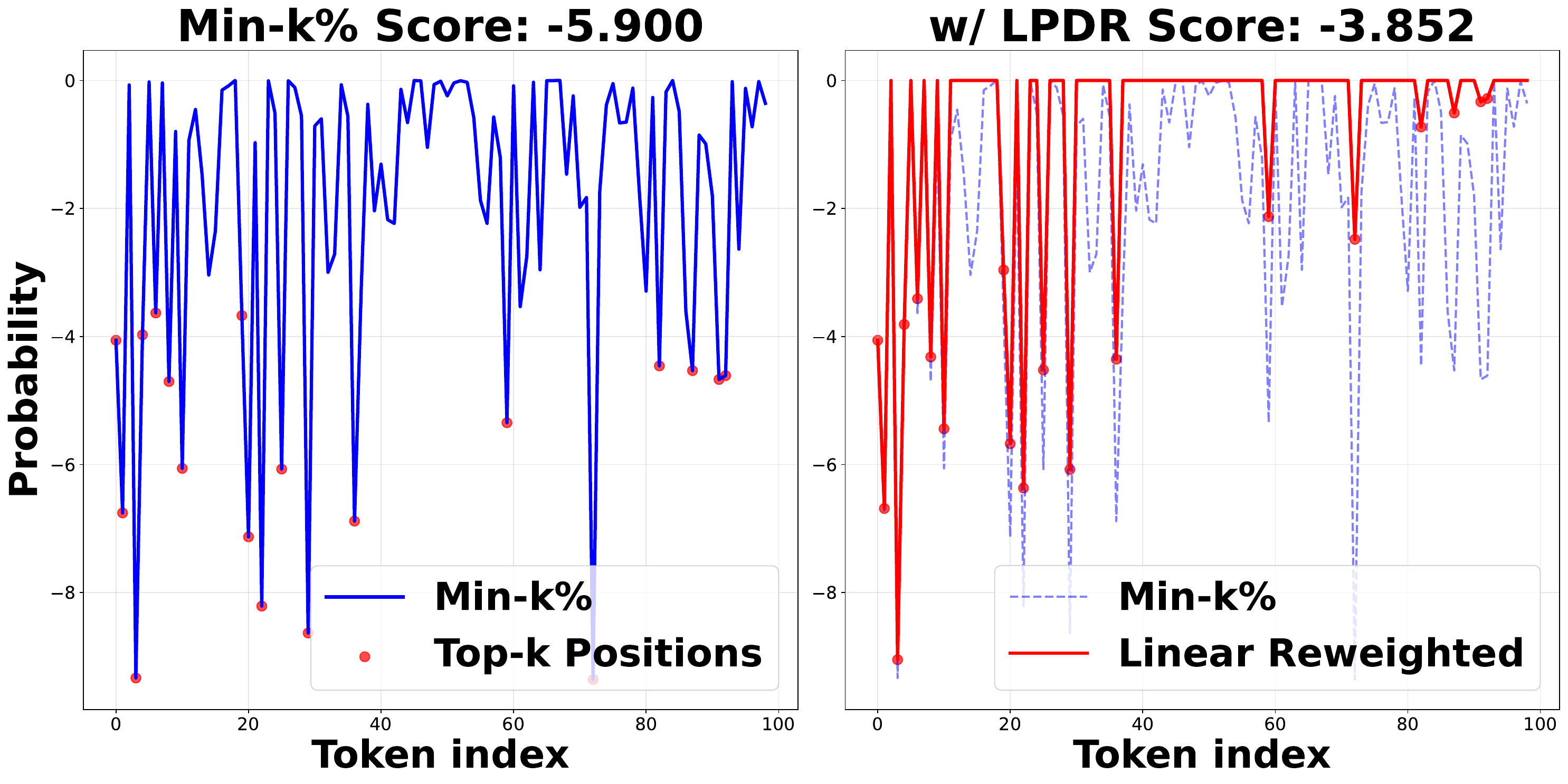}
        \centerline{(b) Non-member sample}
    \end{minipage}
    \caption{Visualization of token-level score changes for (a)  member sample and (b) non-member sample after applying LPDR to the Min-$k$\% method.} 
    \label{fig:reweighting_effect_comparison_linear}
\end{figure}

\begin{figure}[hb]
    \centering
    \begin{minipage}{0.48\textwidth}
        \centering
        \includegraphics[width=\linewidth]{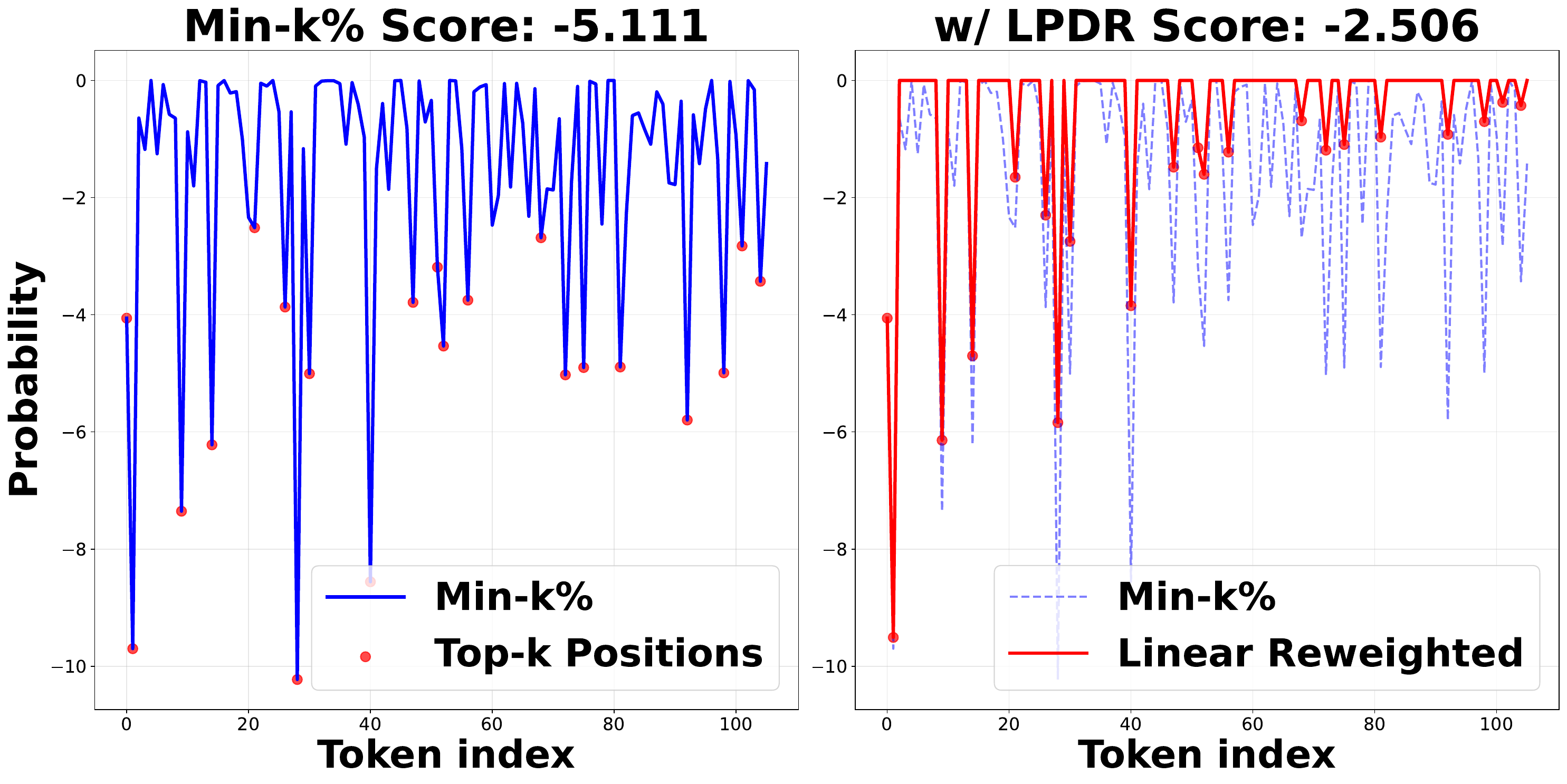}
        \centerline{(a) Member sample}
    \end{minipage}\hfill 
    \begin{minipage}{0.48\textwidth}
        \centering
        \includegraphics[width=\linewidth]{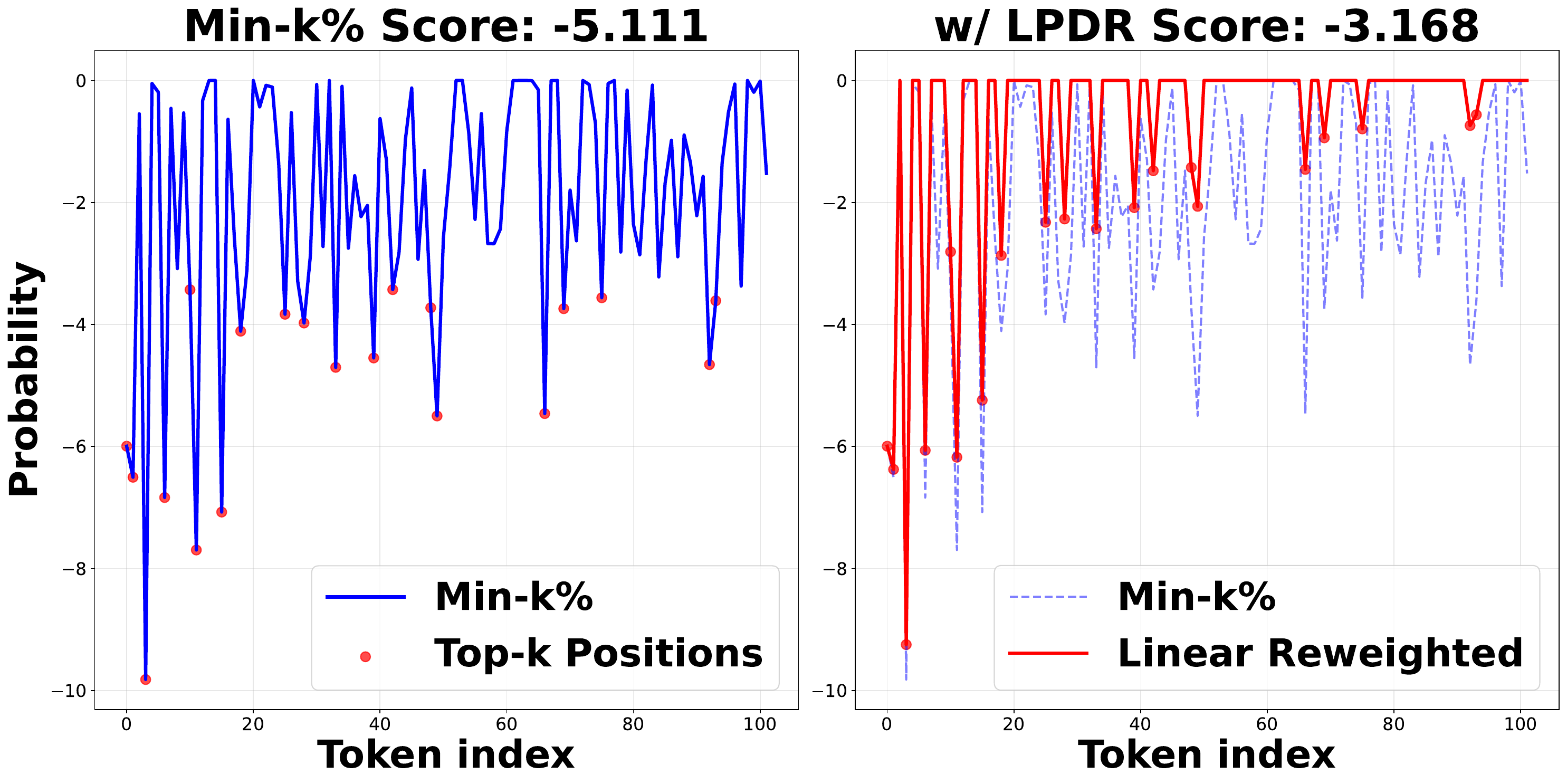}
        \centerline{(b) Non-member sample}
    \end{minipage}
    \caption{Visualization of token-level score changes for (a)  member sample and (b) non-member sample after applying EPDR to the Min-$k$\% method.} 
    \label{fig:reweighting_effect_comparison_exponential}
\end{figure}

\begin{figure}[hb]
    \centering
    \begin{minipage}{0.48\textwidth}
        \centering
        \includegraphics[width=\linewidth]{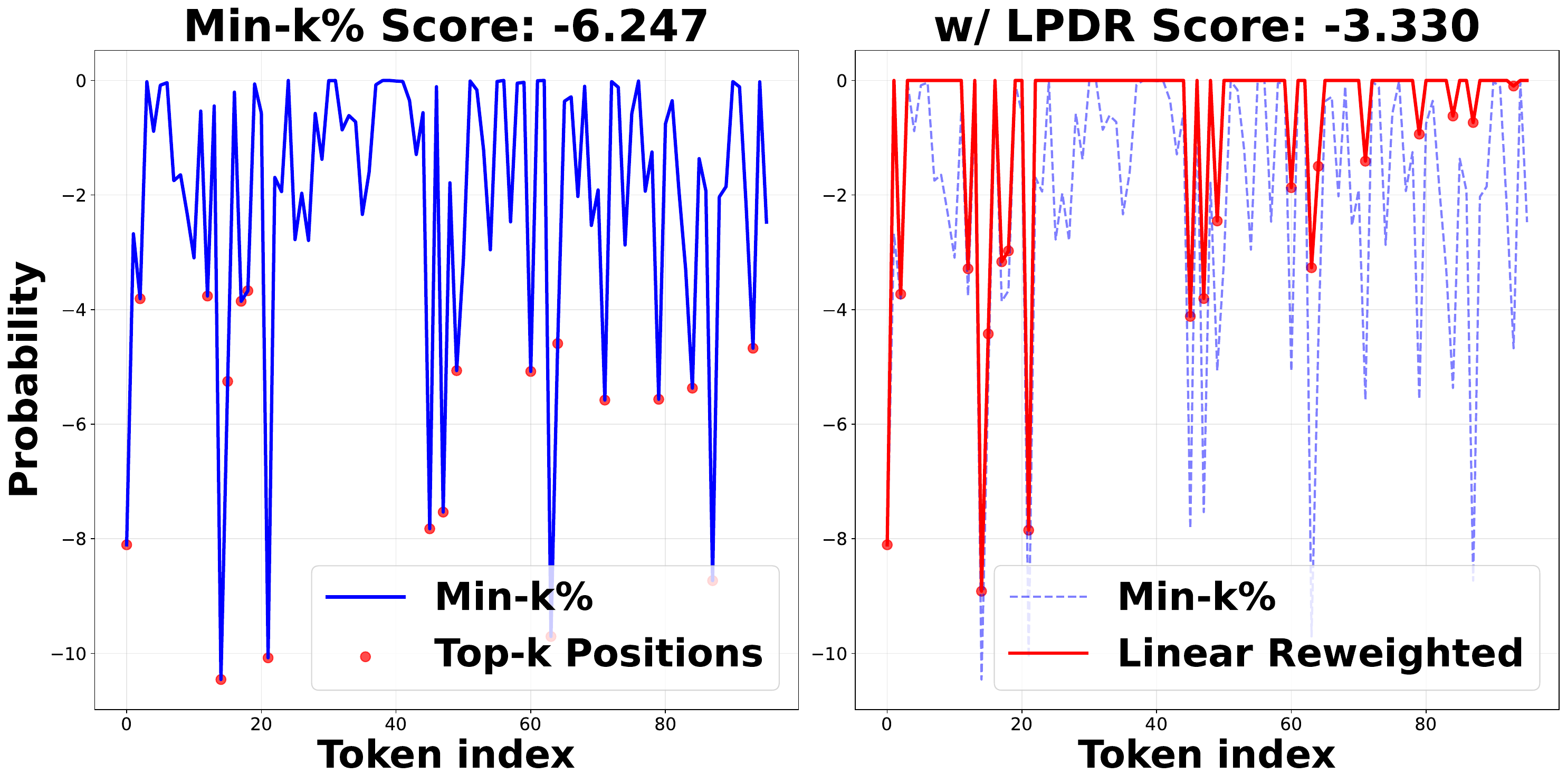}
        \centerline{(a) Member sample}
    \end{minipage}\hfill 
    \begin{minipage}{0.48\textwidth}
        \centering
        \includegraphics[width=\linewidth]{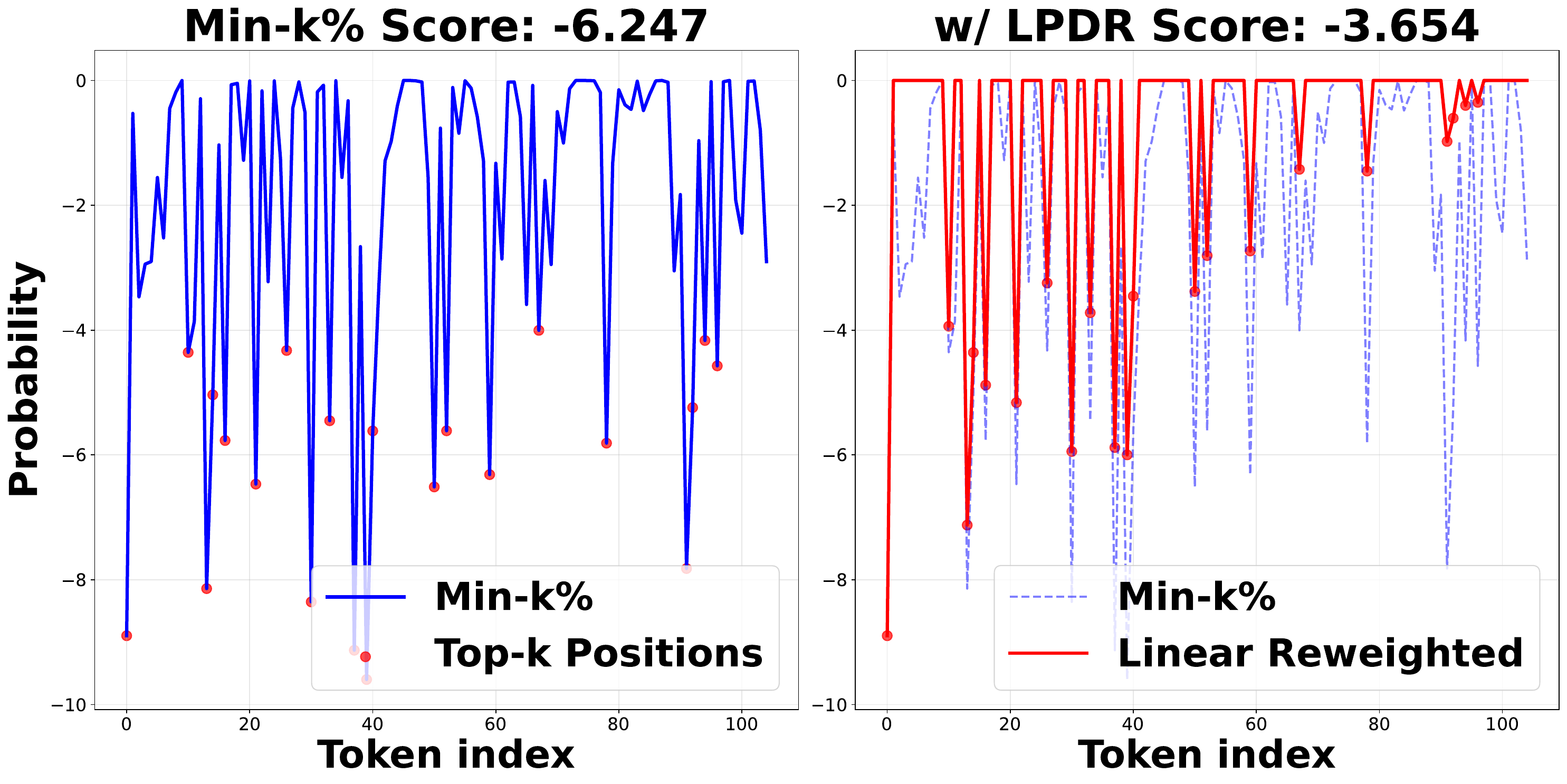}
        \centerline{(b) Non-member sample}
    \end{minipage}
    \caption{Visualization of token-level score changes for (a)  member sample and (b) non-member sample after applying PPDR to the Min-$k$\% method. } 
    \label{fig:reweighting_effect_comparison_ploy}
\end{figure}

This section presents additional visualizations of its effect on individual samples. We specifically select pairs of member and non-member samples that are challenging for the baseline Min-$k$\% method, meaning their original scores are very close and difficult to distinguish. 
Figures \ref{fig:reweighting_effect_comparison_linear}, \ref{fig:reweighting_effect_comparison_exponential}, and \ref{fig:reweighting_effect_comparison_ploy} illustrate how applying our LPDR, EPDR, and PPDR methods, respectively, alters the token-level scores for these ambiguous pairs. In each case, the reweighting process amplifies the scores of the member samples more significantly than the non-member samples by emphasizing the low-probability tokens that appear early in the sequence. This creates a more distinct separation between them, demonstrating how PDR enhances detection accuracy at the individual sample level.

\section{Analysis of Selected Token Distribution and Case Study}
\label{apd:analysis_of_selected_token}

We analyze token-level probability distributions to explain PDR's effectiveness. Non-member samples often feature high-surprise factual tokens (e.g., dates) early in the sequence, whereas member samples, being memorized, show low surprise on these tokens. Standard methods dilute this early signal by averaging across the sequence. PDR, by assigning higher weights to the prefix, acts as a "matched filter": it amplifies the informative early tokens while suppressing the noise from later function words.

\begin{figure*}[htbp]
    \centering
    \includegraphics[width=1.0\linewidth]{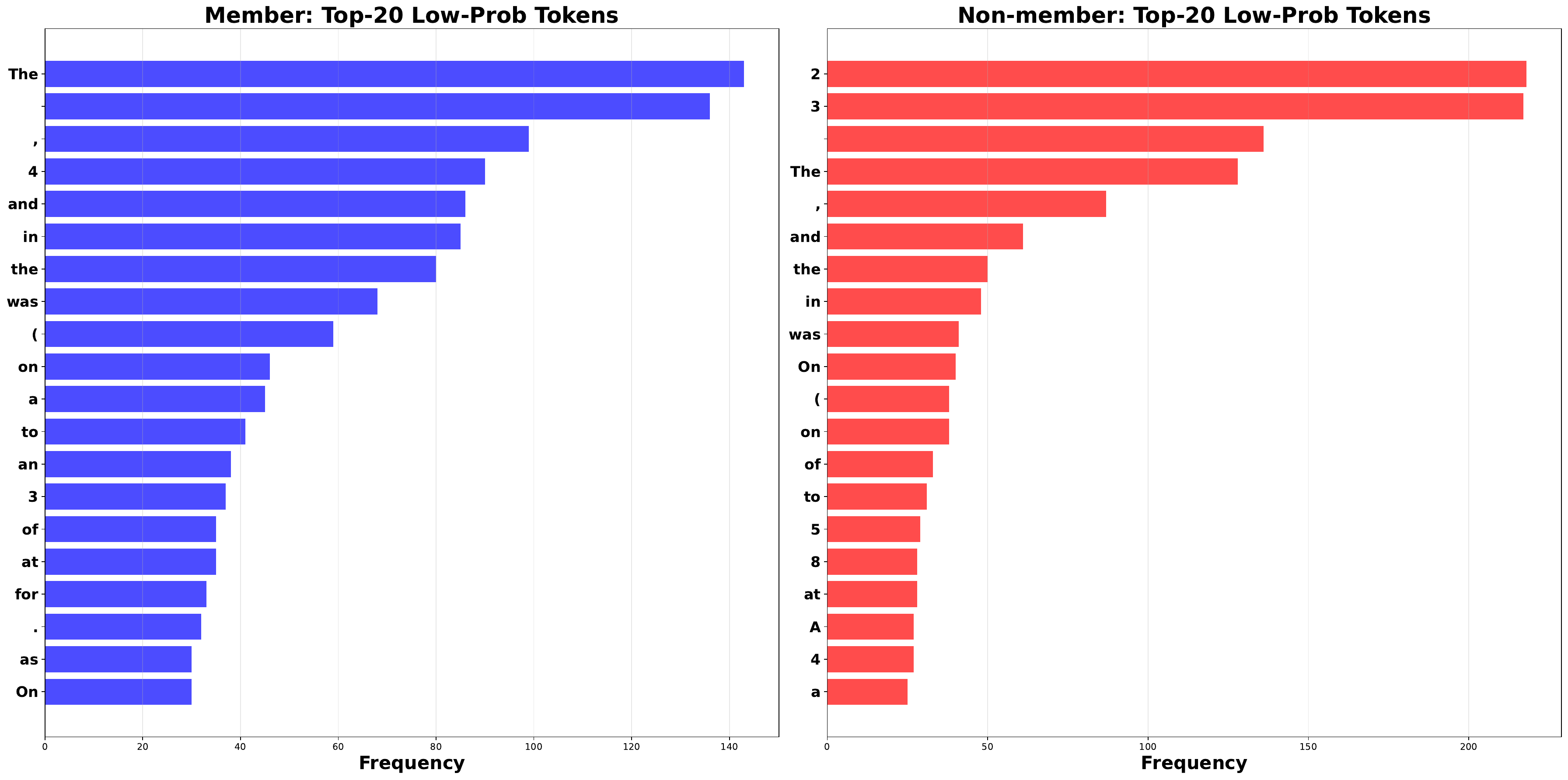}
    \vspace{-8pt}
    \caption{{Top $k$ token frequency comparison between member and non-member samples on LLaMA-13B model with 64-token input length on WikiMIA dataset.}}
    \label{fig:topk_tokens}
\end{figure*}

\textbf{Statistical Evidence (Figure~\ref{fig:topk_tokens}):} We aggregated the positional frequency of the top-$k$\% lowest-probability tokens across the dataset. The results reveal distinct patterns: for non-members, low-probability tokens are often unmemorized factual details (e.g., dates). In contrast, for memorized members, factual tokens have high probability; the lowest-probability "outliers" are instead common function words (e.g., "the", "of") due to their inherent contextual uncertainty. PDR leverages this by amplifying the early, informative signals while suppressing later noise.

\begin{figure*}[htbp]
    \centering
    \begin{minipage}{0.5\textwidth}
        \centering
        \includegraphics[width=\linewidth]{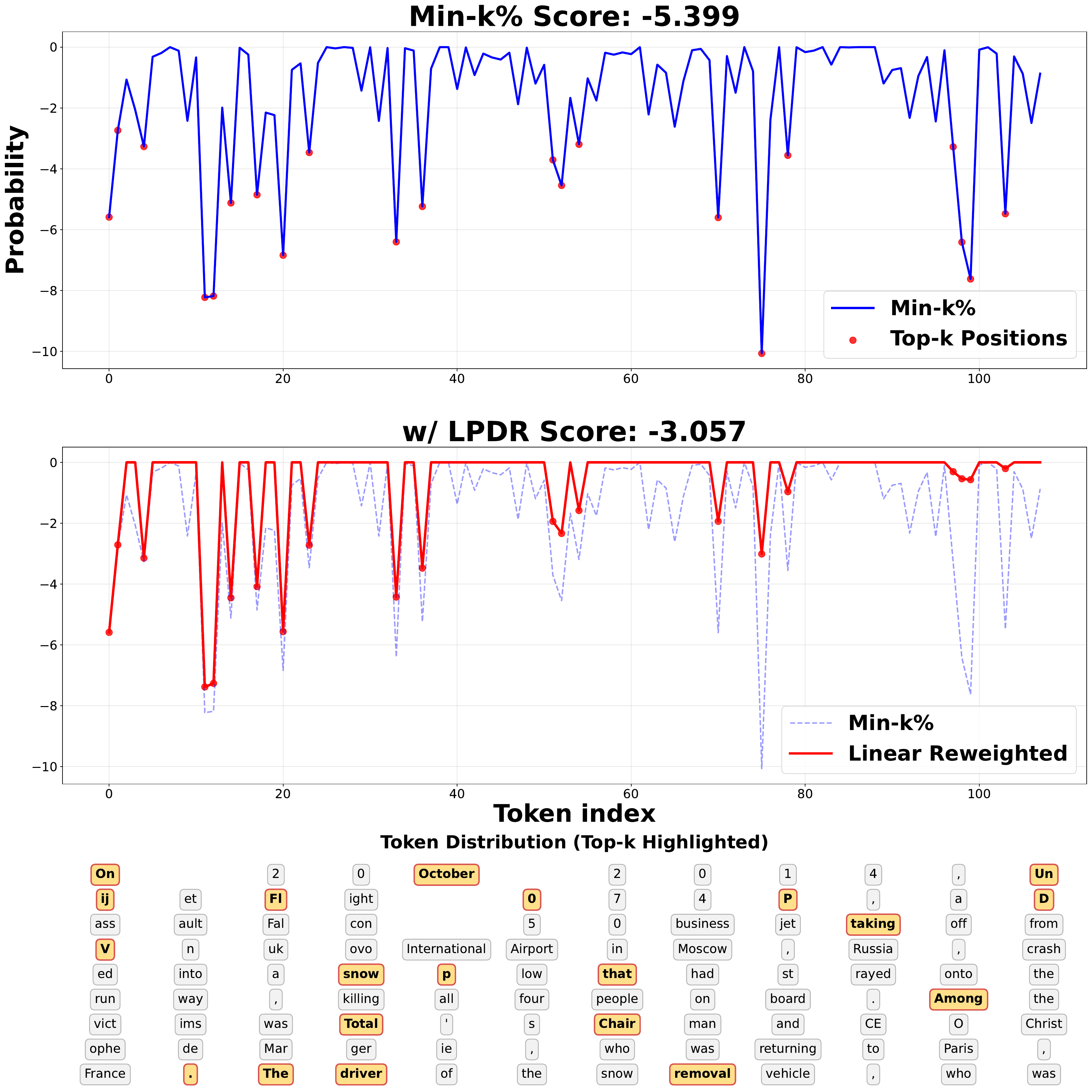}
        \centerline{(a) Member sample}
    \end{minipage}\hfill 
    \begin{minipage}{0.5\textwidth}
        \centering
        \includegraphics[width=\linewidth]{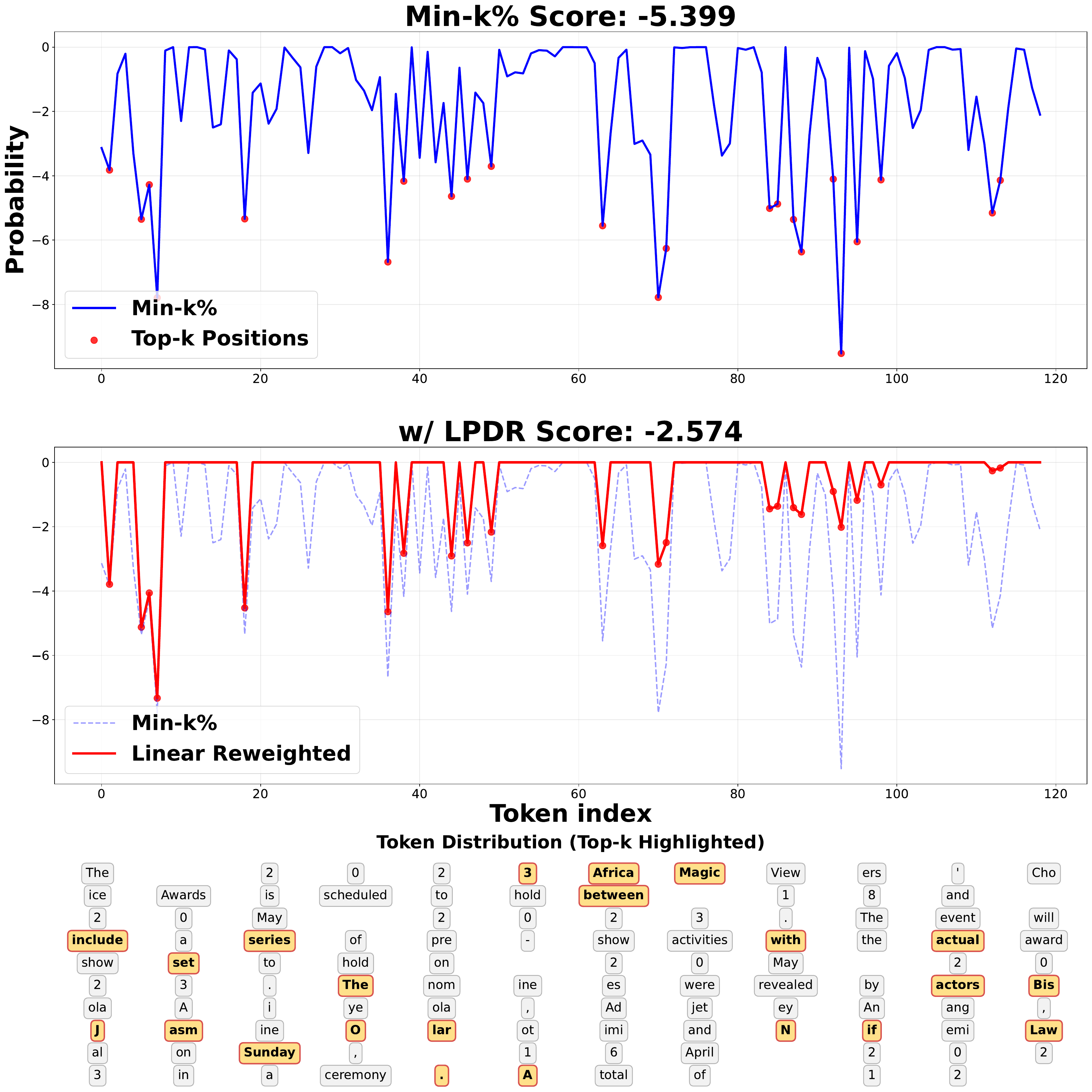}
        \centerline{(b) Non-member sample}
    \end{minipage}
    \caption{{Visualization of token-level score changes and highlight top $k$ tokens for (a)  member sample and (b) non-member sample after applying PPDR to the Min-$k$\% method. } }
    \label{fig:topk_token_visualization}
     \vspace{-10pt}
\end{figure*}

\textbf{Case Study (Figure~\ref{fig:topk_token_visualization}):} Visualizing a specific pair and highlight top k token shows that in the Member sample (a), outlier tokens appear early and are preserved by PDR. In the Non-Member sample (b), outliers are scattered or late, and are effectively suppressed by PDR's decay. This dual mechanism enhances separability.

\begin{figure*}[htbp]
    \centering
    \begin{minipage}{0.5\textwidth}
        \centering
        \includegraphics[width=\linewidth]{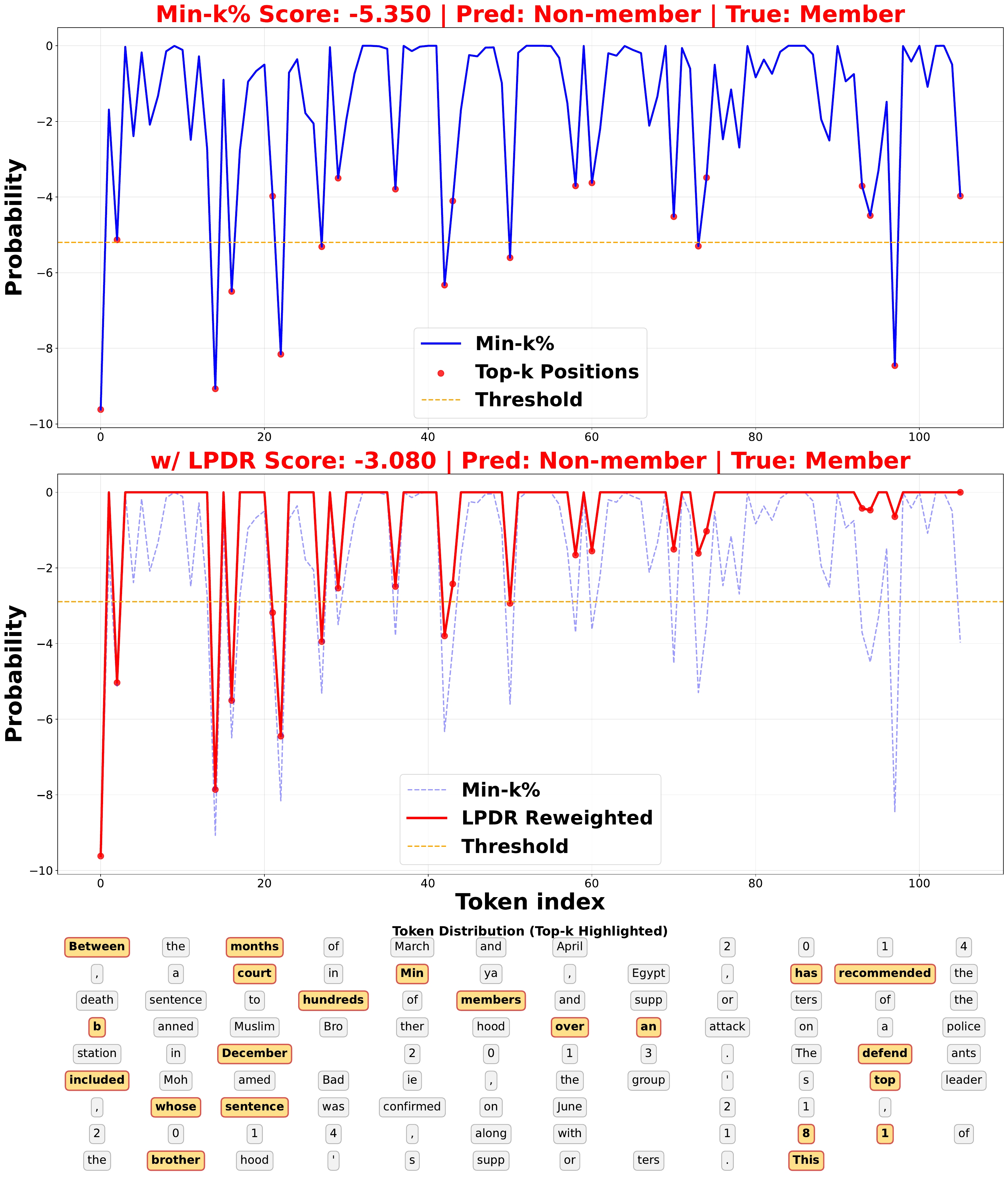}
        \centerline{(a) Member sample}
    \end{minipage}\hfill 
    \begin{minipage}{0.5\textwidth}
        \centering
        \includegraphics[width=\linewidth]{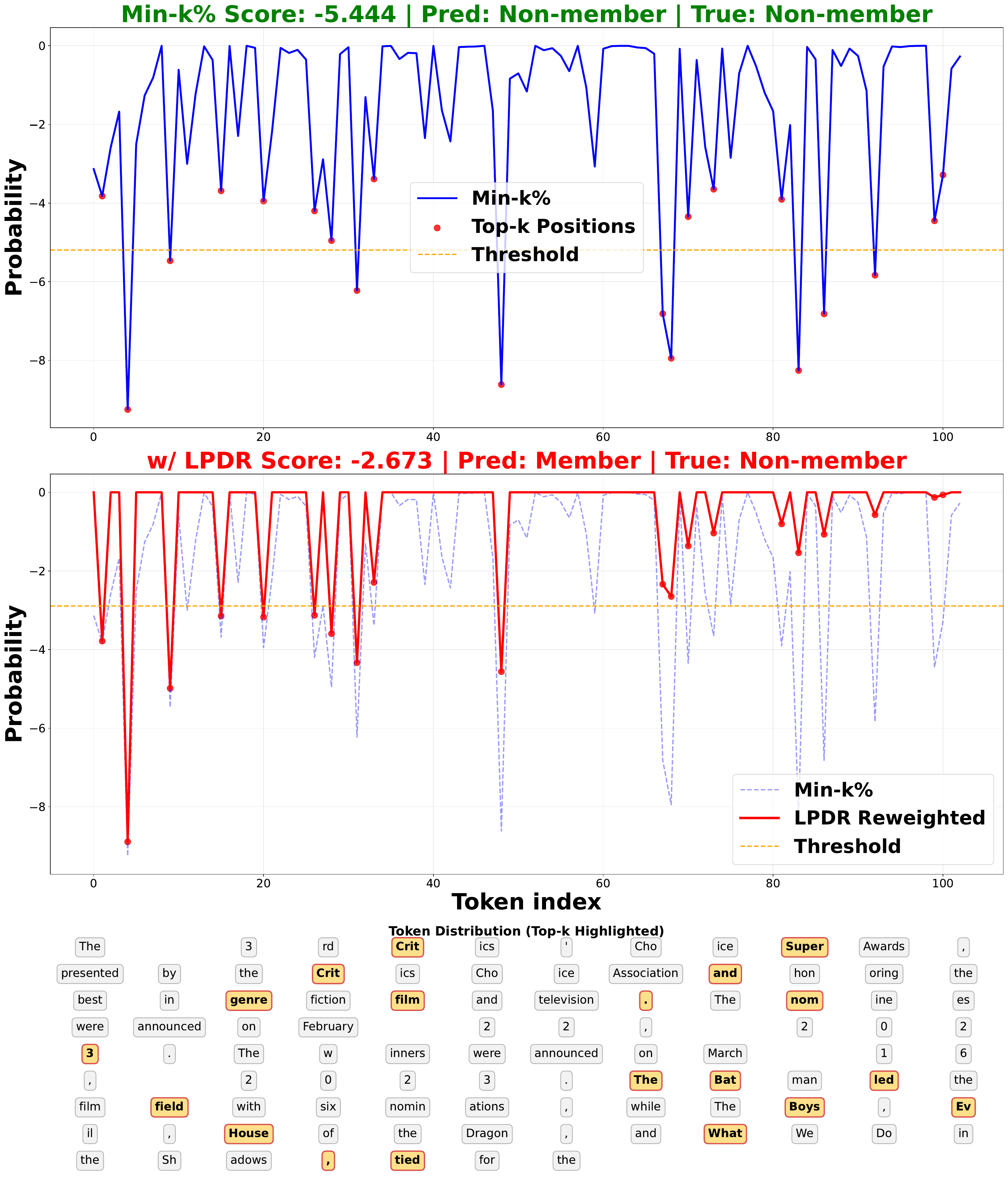}
        \centerline{(b) Non-member sample}
    \end{minipage}
    \caption{{Visualization error case about token-level score changes and highlight top $k$ tokens for (a)  member sample and (b) non-member sample.} }
    \label{fig:error_cases}
    \vspace{-10pt}
\end{figure*}

\textbf{Error Study (Figure~\ref{fig:error_cases}):} While PDR shows consistent improvements in most cases, examining failure cases provides valuable insights into its limitations. Figure~\ref{fig:error_cases} shows challenging examples: (a) a member sample that remains misclassified after PDR, and (b) a non-member incorrectly pushed towards a higher score. A key observation is that the highlighted Top-$k$ tokens (yellow background) are distributed uniformly across the sequence rather than concentrated at the start. This anomaly suggests \textbf{weak memorization}—the model encountered the text but formed no strong memory trace, possibly due to low training frequency, generic content (common function words lacking distinctive features), or \textbf{sentence fragmentation} where the dataset's fixed-length segmentation splits sentences mid-stream, causing the "new sentence start" in the latter half to carry unexpectedly high informativeness and scatter the Top-$k$ tokens. When such positional patterns are absent, PDR's monotonic decay assumption becomes less effective or even counterproductive. These cases highlight potential improvements: adaptive weighting that detects weak memorization or sentence boundaries, and sentence-aware segmentation to preserve natural information flow.

\section{Score distribution.}
\label{apd:score_distribution}
\begin{figure*}[htbp]
    \centering
    \includegraphics[width=1.0\linewidth]{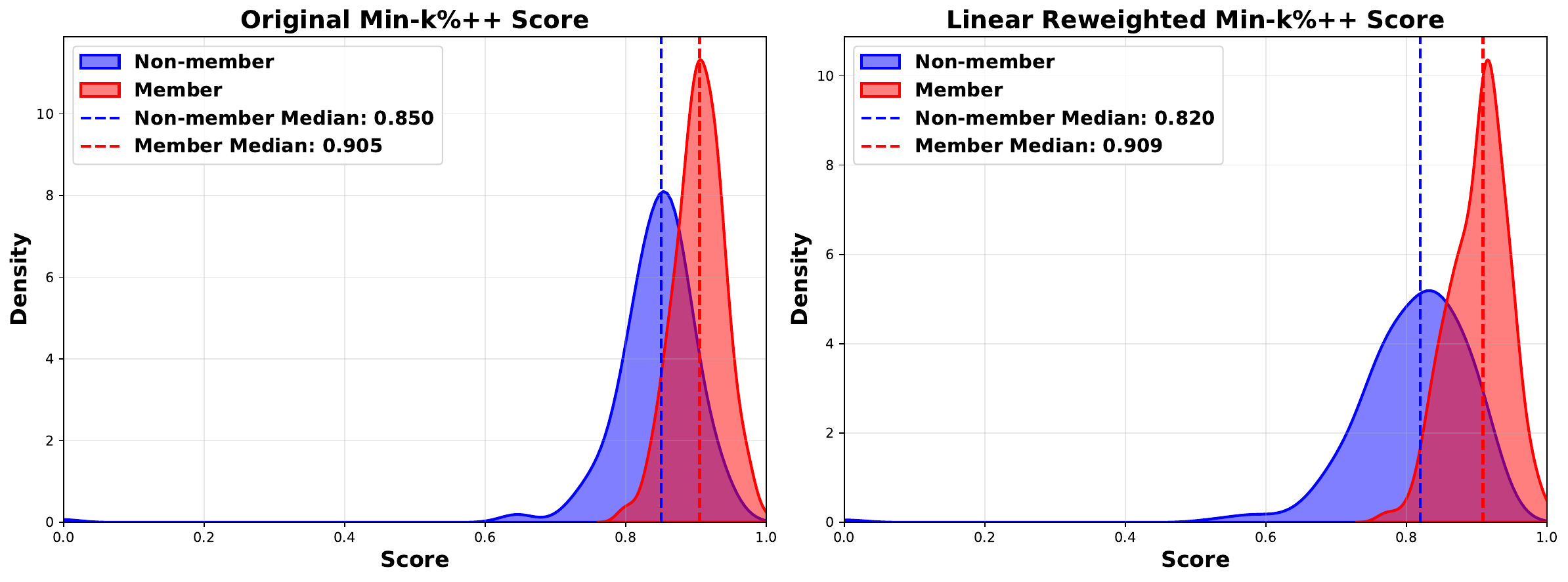}
    \vspace{-8pt}
    \caption{Member and non-member score distribution comparison between Min-$k$\% and LPDR-Min-$k$\% on LLaMA-13B model with 64-token input length on WikiMIA dataset. Our PDR method enhances the separation between member and non-member distributions.}
    \label{fig:mink_vs_linear_distribution}
    
\end{figure*}

To visually demonstrate the effectiveness of our method, we analyze the score distributions of member and non-member samples before and after applying PDR. Figure.~\ref{fig:mink_vs_linear_distribution} illustrates this comparison for the Min-$k$\%++ method on the LLaMA-13B model, using the WikiMIA dataset with a sequence length of 64. For a clearer visualization, the scores are normalized to a range of [0,1].
As the figure shows, the original Min-$k$\%++ method already provides some separation between the two distributions. However, after applying our Linear PDR (LPDR), the distributions are pushed further apart. The member sample distribution shifts noticeably towards higher scores, while the non-member distribution remains relatively stable. This increased separation makes it easier to distinguish between member and non-member samples, directly contributing to the improved AUROC performance we observe in our experiments.

\end{document}